\documentclass[10pt,twocolumn,letterpaper]{article}
\usepackage{arxiv_style/sgpreprint}

\newcommand{\sgModelURL}{https://huggingface.co/pei2333/SuperGlance-9B}
\newcommand{\sgStageOneURL}{https://huggingface.co/pei2333/SuperGlance-9B-Stage1}
\newcommand{\sgStageTwoURL}{https://huggingface.co/pei2333/SuperGlance-9B-Stage2}
\newcommand{\sgDataURL}{https://huggingface.co/datasets/pei2333/SuperGlance-QA}
\newcommand{\sgCodeURL}{https://github.com/GuobaPei/SuperGlance}
\newcommand{\sgProjectURL}{https://guobapei.github.io/SuperGlance/}

\usepackage{amsmath,amssymb,booktabs,xspace}
\usepackage[hyphens]{url}
\usepackage[shortlabels,inline]{enumitem}
\setlist[itemize]{noitemsep,leftmargin=*,topsep=0em}
\setlist[enumerate]{noitemsep,leftmargin=*,topsep=0em}

\newcommand{\suppword}{appendix}   %
\newcommand{\sgtable}[2]{\begin{table}[#2]}   %
\newcommand{\sgcvpr}[1]{}\newcommand{\sgarxiv}[1]{#1}   %

\usepackage{cuted}      %
\usepackage{currfile}
\usepackage{caption}
\usepackage{multirow}
\usepackage{array}
\usepackage{microtype}   %
\usepackage{placeins}    %

\makeatletter
\newcommand{\defval}[2]{\expandafter\def\csname jd@val@#1\endcsname{#2}}
\newcommand{\val}[1]{\ifcsname jd@val@#1\endcsname\csname jd@val@#1\endcsname\else\PackageError{numbers}{Undefined value key #1}{Add it to tools/numbers_registry.py}\textbf{??}\fi}
\makeatother
\defval{design.nopt}{16} %
\defval{design.ntraj}{256} %
\defval{design.nleaves}{4,096} %
\defval{design.horizon}{4} %
\defval{design.hz}{2} %
\defval{design.nposes}{8} %
\defval{design.vecdim}{24} %
\defval{design.shead}{4} %
\defval{design.cbscenes}{92,339} %
\defval{design.tau}{0.1} %
\defval{design.beta}{0.1} %
\defval{design.altmin}{0.05} %
\defval{design.maxpix}{262,144} %
\defval{design.minpix}{65,536} %
\defval{design.imgw}{672} %
\defval{design.imgh}{384} %
\defval{design.imgtok}{252} %
\defval{design.fov}{63.7} %
\defval{design.past}{1.5} %
\defval{design.lthreeoff}{0.3--0.5} %
\defval{design.chance}{6.25} %
\defval{design.bfstep}{0.125} %
\defval{design.bflogit}{25} %
\defval{design.ties}{12--16} %
\defval{train.rank}{32} %
\defval{train.alpha}{64} %
\defval{train.pbig}{87,293,952} %
\defval{train.pnine}{80,216,064} %
\defval{train.lr}{$10^{-4}$} %
\defval{train.warm}{3} %
\defval{train.wd}{0.01} %
\defval{train.batch}{64} %
\defval{train.draws}{32} %
\defval{train.kl}{0.1} %
\defval{s1.steps}{7,813} %
\defval{s1.draws}{500,000} %
\defval{s1.w.sat}{30} %
\defval{s1.w.gqa}{30} %
\defval{s1.w.tally}{25} %
\defval{s1.w.aitwod}{6} %
\defval{s1.w.chart}{8} %
\defval{s1.w.textvqa}{1} %
\defval{s1.taskexp}{1/2} %
\defval{s1.ansexp}{3/4} %
\defval{s1.satfirst}{139,632} %
\defval{s1.satcnt1}{16,548} %
\defval{s1.satcnt}{19,101} %
\defval{s1.satcnt1pct}{87} %
\defval{s1.train}{725,647} %
\defval{s1.dev}{45,826} %
\defval{s1.all}{771,473} %
\defval{s1.n.sat}{139,632} %
\defval{s1.d.sat}{7,382} %
\defval{s1.n.gqa}{435,825} %
\defval{s1.d.gqa}{29,562} %
\defval{s1.n.tally}{119,859} %
\defval{s1.d.tally}{6,933} %
\defval{s1.n.aitwod}{5,777} %
\defval{s1.d.aitwod}{678} %
\defval{s1.n.chart}{22,136} %
\defval{s1.d.chart}{1,147} %
\defval{s1.n.textvqa}{2,418} %
\defval{s1.d.textvqa}{124} %
\defval{s1.rmdup}{1,688} %
\defval{s1.rmext}{8,171} %
\defval{s1.rmmulti}{88} %
\defval{s1.chartgap}{15} %
\defval{s1.devfrac}{5} %
\defval{s1.flips}{12,158} %
\defval{s1.pool.d}{737,805} %
\defval{s1.box}{17,966} %
\defval{s1.ss}{10,339} %
\defval{s1.pool.e}{766,110} %
\defval{s1.text}{349,552} %
\defval{s1.textshare}{25} %
\defval{s1.ntextsrc}{13} %
\defval{s1.pool.f}{1,115,662} %
\defval{s1.above}{2,488} %
\defval{s1.hedge}{40,000} %
\defval{s1.pool.g}{1,158,150} %
\defval{s1.phash}{10} %
\defval{s1.ngram}{10} %
\defval{s1.ngramfrac}{50} %
\defval{s1.boxiou}{0.3--0.8} %
\defval{readout.agree}{100} %
\defval{readout.mass}{1.000} %
\defval{readout.masstr}{0.957--0.995} %
\defval{tab.satdev.base}{75.75} %
\defval{tab.satdev.base.r}{75.8} %
\defval{tab.satdev.orig}{89.8} %
\defval{tab.satdev.orig.r}{89.8} %
\defval{tab.satdev.a}{87.5} %
\defval{tab.satdev.a.r}{87.5} %
\defval{tab.satdev.c}{87.9} %
\defval{tab.satdev.c.r}{87.9} %
\defval{tab.satdev.d}{84.9} %
\defval{tab.satdev.d.r}{84.9} %
\defval{tab.satdev.e}{84.7} %
\defval{tab.satdev.e.r}{84.7} %
\defval{tab.satdev.f}{84.45} %
\defval{tab.satdev.f.r}{84.5} %
\defval{tab.satdev.g}{83.6} %
\defval{tab.satdev.g.r}{83.6} %
\defval{tab.satdev.stwof}{83.9} %
\defval{tab.satdev.stwof.r}{83.9} %
\defval{tab.satdev.nbase}{75.8} %
\defval{tab.satdev.nbase.r}{75.8} %
\defval{tab.satdev.ng}{86.25} %
\defval{tab.satdev.ng.r}{86.3} %
\defval{tab.satdev.nstwog}{85.05} %
\defval{tab.satdev.nstwog.r}{85.1} %
\defval{tab.satdev.nsthree}{86.3} %
\defval{tab.satdev.nsthree.r}{86.3} %
\defval{tab.cvbench.base}{85.22} %
\defval{tab.cvbench.base.r}{85.2} %
\defval{tab.cvbench.orig}{85.49} %
\defval{tab.cvbench.orig.r}{85.5} %
\defval{tab.cvbench.a}{86.63} %
\defval{tab.cvbench.a.r}{86.6} %
\defval{tab.cvbench.c}{84.94} %
\defval{tab.cvbench.c.r}{84.9} %
\defval{tab.cvbench.d}{86.3} %
\defval{tab.cvbench.d.r}{86.3} %
\defval{tab.cvbench.e}{87.67} %
\defval{tab.cvbench.e.r}{87.7} %
\defval{tab.cvbench.f}{86.93} %
\defval{tab.cvbench.f.r}{86.9} %
\defval{tab.cvbench.g}{86.91} %
\defval{tab.cvbench.g.r}{86.9} %
\defval{tab.cvbench.stwof}{85.88} %
\defval{tab.cvbench.stwof.r}{85.9} %
\defval{tab.cvbench.nbase}{86.36} %
\defval{tab.cvbench.nbase.r}{86.4} %
\defval{tab.cvbench.ng}{87.21} %
\defval{tab.cvbench.ng.r}{87.2} %
\defval{tab.cvbench.nstwog}{86.84} %
\defval{tab.cvbench.nstwog.r}{86.8} %
\defval{tab.cvbench.nsthree}{87.01} %
\defval{tab.cvbench.nsthree.r}{87.0} %
\defval{tab.blink.base}{80.55} %
\defval{tab.blink.base.r}{80.6} %
\defval{tab.blink.orig}{73.87} %
\defval{tab.blink.orig.r}{73.9} %
\defval{tab.blink.a}{77.21} %
\defval{tab.blink.a.r}{77.2} %
\defval{tab.blink.c}{75.83} %
\defval{tab.blink.c.r}{75.8} %
\defval{tab.blink.d}{77.6} %
\defval{tab.blink.d.r}{77.6} %
\defval{tab.blink.e}{79.57} %
\defval{tab.blink.e.r}{79.6} %
\defval{tab.blink.f}{81.73} %
\defval{tab.blink.f.r}{81.7} %
\defval{tab.blink.g}{81.73} %
\defval{tab.blink.g.r}{81.7} %
\defval{tab.blink.stwof}{80.75} %
\defval{tab.blink.stwof.r}{80.8} %
\defval{tab.blink.nbase}{79.37} %
\defval{tab.blink.nbase.r}{79.4} %
\defval{tab.blink.ng}{83.3} %
\defval{tab.blink.ng.r}{83.3} %
\defval{tab.blink.nstwog}{81.53} %
\defval{tab.blink.nstwog.r}{81.5} %
\defval{tab.blink.nsthree}{81.73} %
\defval{tab.blink.nsthree.r}{81.7} %
\defval{tab.blinkol.base}{72.1} %
\defval{tab.blinkol.base.r}{72.1} %
\defval{tab.blinkol.orig}{63.1} %
\defval{tab.blinkol.orig.r}{63.1} %
\defval{tab.blinkol.a}{60.7} %
\defval{tab.blinkol.a.r}{60.7} %
\defval{tab.blinkol.c}{61.5} %
\defval{tab.blinkol.c.r}{61.5} %
\defval{tab.blinkol.d}{63.9} %
\defval{tab.blinkol.d.r}{63.9} %
\defval{tab.blinkol.e}{75.4} %
\defval{tab.blinkol.e.r}{75.4} %
\defval{tab.blinkol.f}{75.4} %
\defval{tab.blinkol.f.r}{75.4} %
\defval{tab.blinkol.g}{76.2} %
\defval{tab.blinkol.g.r}{76.2} %
\defval{tab.blinkol.stwof}{74.59} %
\defval{tab.blinkol.stwof.r}{74.6} %
\defval{tab.blinkol.nbase}{68.85} %
\defval{tab.blinkol.nbase.r}{68.9} %
\defval{tab.blinkol.ng}{80.33} %
\defval{tab.blinkol.ng.r}{80.3} %
\defval{tab.blinkol.nstwog}{74.59} %
\defval{tab.blinkol.nstwog.r}{74.6} %
\defval{tab.blinkol.nsthree}{73.77} %
\defval{tab.blinkol.nsthree.r}{73.8} %
\defval{tab.blinksr.base}{88.8} %
\defval{tab.blinksr.base.r}{88.8} %
\defval{tab.blinksr.orig}{67.1} %
\defval{tab.blinksr.orig.r}{67.1} %
\defval{tab.blinksr.a}{81.8} %
\defval{tab.blinksr.a.r}{81.8} %
\defval{tab.blinksr.c}{79.7} %
\defval{tab.blinksr.c.r}{79.7} %
\defval{tab.blinksr.d}{86.7} %
\defval{tab.blinksr.d.r}{86.7} %
\defval{tab.blinksr.e}{85.3} %
\defval{tab.blinksr.e.r}{85.3} %
\defval{tab.blinksr.f}{88.8} %
\defval{tab.blinksr.f.r}{88.8} %
\defval{tab.blinksr.g}{87.4} %
\defval{tab.blinksr.g.r}{87.4} %
\defval{tab.blinksr.stwof}{86.71} %
\defval{tab.blinksr.stwof.r}{86.7} %
\defval{tab.blinksr.nbase}{88.81} %
\defval{tab.blinksr.nbase.r}{88.8} %
\defval{tab.blinksr.ng}{91.61} %
\defval{tab.blinksr.ng.r}{91.6} %
\defval{tab.blinksr.nstwog}{88.81} %
\defval{tab.blinksr.nstwog.r}{88.8} %
\defval{tab.blinksr.nsthree}{89.51} %
\defval{tab.blinksr.nsthree.r}{89.5} %
\defval{tab.tdsr.base}{54.04} %
\defval{tab.tdsr.base.r}{54.0} %
\defval{tab.tdsr.orig}{52.22} %
\defval{tab.tdsr.orig.r}{52.2} %
\defval{tab.tdsr.a}{53.89} %
\defval{tab.tdsr.a.r}{53.9} %
\defval{tab.tdsr.c}{54.94} %
\defval{tab.tdsr.c.r}{54.9} %
\defval{tab.tdsr.d}{54.62} %
\defval{tab.tdsr.d.r}{54.6} %
\defval{tab.tdsr.e}{54.82} %
\defval{tab.tdsr.e.r}{54.8} %
\defval{tab.tdsr.f}{55.89} %
\defval{tab.tdsr.f.r}{55.9} %
\defval{tab.tdsr.g}{57.03} %
\defval{tab.tdsr.g.r}{57.0} %
\defval{tab.tdsr.stwof}{55.96} %
\defval{tab.tdsr.stwof.r}{56.0} %
\defval{tab.tdsr.nbase}{45.39} %
\defval{tab.tdsr.nbase.r}{45.4} %
\defval{tab.tdsr.ng}{58.39} %
\defval{tab.tdsr.ng.r}{58.4} %
\defval{tab.tdsr.nstwog}{58.83} %
\defval{tab.tdsr.nstwog.r}{58.8} %
\defval{tab.tdsr.nsthree}{57.86} %
\defval{tab.tdsr.nsthree.r}{57.9} %
\defval{tab.above.base}{69.9} %
\defval{tab.above.base.r}{69.9} %
\defval{tab.above.orig}{48.0} %
\defval{tab.above.orig.r}{48.0} %
\defval{tab.above.a}{51.6} %
\defval{tab.above.a.r}{51.6} %
\defval{tab.above.c}{53.0} %
\defval{tab.above.c.r}{53.0} %
\defval{tab.above.d}{57.1} %
\defval{tab.above.d.r}{57.1} %
\defval{tab.above.e}{58.5} %
\defval{tab.above.e.r}{58.5} %
\defval{tab.above.f}{60.9} %
\defval{tab.above.f.r}{60.9} %
\defval{tab.above.g}{65.5} %
\defval{tab.above.g.r}{65.5} %
\defval{tab.above.stwof}{66.5} %
\defval{tab.above.stwof.r}{66.5} %
\defval{tab.above.nbase}{71.2} %
\defval{tab.above.nbase.r}{71.2} %
\defval{tab.above.ng}{75.2} %
\defval{tab.above.ng.r}{75.2} %
\defval{tab.above.nstwog}{76.4} %
\defval{tab.above.nstwog.r}{76.4} %
\defval{tab.above.nsthree}{69.73} %
\defval{tab.above.nsthree.r}{69.7} %
\defval{tab.aboveyes.base}{424} %
\defval{tab.aboveyes.base.r}{424} %
\defval{tab.aboveyes.orig}{1262} %
\defval{tab.aboveyes.orig.r}{1262} %
\defval{tab.aboveyes.a}{1160} %
\defval{tab.aboveyes.a.r}{1160} %
\defval{tab.aboveyes.c}{1130} %
\defval{tab.aboveyes.c.r}{1130} %
\defval{tab.aboveyes.d}{986} %
\defval{tab.aboveyes.d.r}{986} %
\defval{tab.aboveyes.e}{933} %
\defval{tab.aboveyes.e.r}{933} %
\defval{tab.aboveyes.f}{837} %
\defval{tab.aboveyes.f.r}{837} %
\defval{tab.aboveyes.g}{626} %
\defval{tab.aboveyes.g.r}{626} %
\defval{tab.aboveyes.stwof}{723} %
\defval{tab.aboveyes.stwof.r}{723} %
\defval{tab.aboveyes.nbase}{540} %
\defval{tab.aboveyes.nbase.r}{540} %
\defval{tab.aboveyes.ng}{715} %
\defval{tab.aboveyes.ng.r}{715} %
\defval{tab.aboveyes.nstwog}{689} %
\defval{tab.aboveyes.nstwog.r}{689} %
\defval{tab.aboveyes.nsthree}{859} %
\defval{tab.aboveyes.nsthree.r}{859} %
\defval{tab.dlmst.base}{66.8} %
\defval{tab.dlmst.base.r}{66.8} %
\defval{tab.dlmst.orig}{63.7} %
\defval{tab.dlmst.orig.r}{63.7} %
\defval{tab.dlmst.a}{65.4} %
\defval{tab.dlmst.a.r}{65.4} %
\defval{tab.dlmst.c}{62.9} %
\defval{tab.dlmst.c.r}{62.9} %
\defval{tab.dlmst.d}{65.6} %
\defval{tab.dlmst.d.r}{65.6} %
\defval{tab.dlmst.e}{68.6} %
\defval{tab.dlmst.e.r}{68.6} %
\defval{tab.dlmst.f}{67.6} %
\defval{tab.dlmst.f.r}{67.6} %
\defval{tab.dlmst.g}{66.3} %
\defval{tab.dlmst.g.r}{66.3} %
\defval{tab.dlmst.stwof}{82.3} %
\defval{tab.dlmst.stwof.r}{82.3} %
\defval{tab.dlmst.nbase}{70.87} %
\defval{tab.dlmst.nbase.r}{70.9} %
\defval{tab.dlmst.ng}{74.24} %
\defval{tab.dlmst.ng.r}{74.2} %
\defval{tab.dlmst.nstwog}{82.56} %
\defval{tab.dlmst.nstwog.r}{82.6} %
\defval{tab.dlmst.nsthree}{83.9} %
\defval{tab.dlmst.nsthree.r}{83.9} %
\defval{tab.dlmall.base}{58.8} %
\defval{tab.dlmall.base.r}{58.8} %
\defval{tab.dlmall.orig}{58.2} %
\defval{tab.dlmall.orig.r}{58.2} %
\defval{tab.dlmall.a}{60.6} %
\defval{tab.dlmall.a.r}{60.6} %
\defval{tab.dlmall.c}{59.2} %
\defval{tab.dlmall.c.r}{59.2} %
\defval{tab.dlmall.d}{61.4} %
\defval{tab.dlmall.d.r}{61.4} %
\defval{tab.dlmall.e}{65.3} %
\defval{tab.dlmall.e.r}{65.3} %
\defval{tab.dlmall.f}{64.1} %
\defval{tab.dlmall.f.r}{64.1} %
\defval{tab.dlmall.g}{62.5} %
\defval{tab.dlmall.g.r}{62.5} %
\defval{tab.dlmall.stwof}{81.3} %
\defval{tab.dlmall.stwof.r}{81.3} %
\defval{tab.dlmall.nbase}{60.81} %
\defval{tab.dlmall.nbase.r}{60.8} %
\defval{tab.dlmall.ng}{70.04} %
\defval{tab.dlmall.ng.r}{70.0} %
\defval{tab.dlmall.nstwog}{78.79} %
\defval{tab.dlmall.nstwog.r}{78.8} %
\defval{tab.dlmall.nsthree}{80.37} %
\defval{tab.dlmall.nsthree.r}{80.4} %
\defval{tab.mmlu.base}{72.9} %
\defval{tab.mmlu.base.r}{72.9} %
\defval{tab.mmlu.orig}{74.61} %
\defval{tab.mmlu.orig.r}{74.6} %
\defval{tab.mmlu.a}{74.75} %
\defval{tab.mmlu.a.r}{74.8} %
\defval{tab.mmlu.c}{73.83} %
\defval{tab.mmlu.c.r}{73.8} %
\defval{tab.mmlu.d}{74.15} %
\defval{tab.mmlu.d.r}{74.2} %
\defval{tab.mmlu.e}{73.81} %
\defval{tab.mmlu.e.r}{73.8} %
\defval{tab.mmlu.f}{74.92} %
\defval{tab.mmlu.f.r}{74.9} %
\defval{tab.mmlu.g}{75.01} %
\defval{tab.mmlu.g.r}{75.0} %
\defval{tab.mmlu.stwof}{75.38} %
\defval{tab.mmlu.stwof.r}{75.4} %
\defval{tab.mmlu.nbase}{75.91} %
\defval{tab.mmlu.nbase.r}{75.9} %
\defval{tab.mmlu.ng}{77.5} %
\defval{tab.mmlu.ng.r}{77.5} %
\defval{tab.mmlu.nstwog}{77.58} %
\defval{tab.mmlu.nstwog.r}{77.6} %
\defval{tab.mmlu.nsthree}{77.45} %
\defval{tab.mmlu.nsthree.r}{77.5} %
\defval{tab.mmlupro.base}{48.67} %
\defval{tab.mmlupro.base.r}{48.7} %
\defval{tab.mmlupro.orig}{49.77} %
\defval{tab.mmlupro.orig.r}{49.8} %
\defval{tab.mmlupro.a}{50.71} %
\defval{tab.mmlupro.a.r}{50.7} %
\defval{tab.mmlupro.c}{50.88} %
\defval{tab.mmlupro.c.r}{50.9} %
\defval{tab.mmlupro.d}{51.37} %
\defval{tab.mmlupro.d.r}{51.4} %
\defval{tab.mmlupro.e}{50.88} %
\defval{tab.mmlupro.e.r}{50.9} %
\defval{tab.mmlupro.f}{52.31} %
\defval{tab.mmlupro.f.r}{52.3} %
\defval{tab.mmlupro.g}{51.7} %
\defval{tab.mmlupro.g.r}{51.7} %
\defval{tab.mmlupro.stwof}{52.17} %
\defval{tab.mmlupro.stwof.r}{52.2} %
\defval{tab.mmlupro.nbase}{50.41} %
\defval{tab.mmlupro.nbase.r}{50.4} %
\defval{tab.mmlupro.ng}{52.73} %
\defval{tab.mmlupro.ng.r}{52.7} %
\defval{tab.mmlupro.nstwog}{53.26} %
\defval{tab.mmlupro.nstwog.r}{53.3} %
\defval{tab.mmlupro.nsthree}{51.77} %
\defval{tab.mmlupro.nsthree.r}{51.8} %
\defval{tab.tqa.base}{72.71} %
\defval{tab.tqa.base.r}{72.7} %
\defval{tab.tqa.orig}{62.06} %
\defval{tab.tqa.orig.r}{62.1} %
\defval{tab.tqa.a}{64.26} %
\defval{tab.tqa.a.r}{64.3} %
\defval{tab.tqa.c}{60.47} %
\defval{tab.tqa.c.r}{60.5} %
\defval{tab.tqa.d}{66.59} %
\defval{tab.tqa.d.r}{66.6} %
\defval{tab.tqa.e}{64.99} %
\defval{tab.tqa.e.r}{65.0} %
\defval{tab.tqa.f}{65.97} %
\defval{tab.tqa.f.r}{66.0} %
\defval{tab.tqa.g}{68.67} %
\defval{tab.tqa.g.r}{68.7} %
\defval{tab.tqa.stwof}{64.14} %
\defval{tab.tqa.stwof.r}{64.1} %
\defval{tab.tqa.nbase}{74.42} %
\defval{tab.tqa.nbase.r}{74.4} %
\defval{tab.tqa.ng}{71.11} %
\defval{tab.tqa.ng.r}{71.1} %
\defval{tab.tqa.nstwog}{72.34} %
\defval{tab.tqa.nstwog.r}{72.3} %
\defval{tab.tqa.nsthree}{71.24} %
\defval{tab.tqa.nsthree.r}{71.2} %
\defval{tab.sattest.base}{65.87} %
\defval{tab.sattest.base.r}{65.9} %
\defval{tab.sattest.orig}{74.52} %
\defval{tab.sattest.orig.r}{74.5} %
\defval{tab.sattest.a}{72.6} %
\defval{tab.sattest.a.r}{72.6} %
\defval{tab.sattest.c}{69.23} %
\defval{tab.sattest.c.r}{69.2} %
\defval{tab.sattest.d}{66.83} %
\defval{tab.sattest.d.r}{66.8} %
\defval{tab.sattest.e}{70.19} %
\defval{tab.sattest.e.r}{70.2} %
\defval{tab.sattest.f}{73.08} %
\defval{tab.sattest.f.r}{73.1} %
\defval{tab.sattest.g}{73.56} %
\defval{tab.sattest.g.r}{73.6} %
\defval{tab.sattest.stwof}{71.63} %
\defval{tab.sattest.stwof.r}{71.6} %
\defval{tab.sattest.nbase}{69.23} %
\defval{tab.sattest.nbase.r}{69.2} %
\defval{tab.sattest.ng}{72.6} %
\defval{tab.sattest.ng.r}{72.6} %
\defval{tab.sattest.nstwog}{75.48} %
\defval{tab.sattest.nstwog.r}{75.5} %
\defval{tab.sattest.nsthree}{74.04} %
\defval{tab.sattest.nsthree.r}{74.0} %
\defval{tab.blinkcnt.base}{69.2} %
\defval{tab.blinkcnt.base.r}{69.2} %
\defval{tab.blinkcnt.orig}{77.5} %
\defval{tab.blinkcnt.orig.r}{77.5} %
\defval{tab.blinkcnt.a}{73.3} %
\defval{tab.blinkcnt.a.r}{73.3} %
\defval{tab.blinkcnt.c}{74.2} %
\defval{tab.blinkcnt.c.r}{74.2} %
\defval{tab.blinkcnt.d}{70.8} %
\defval{tab.blinkcnt.d.r}{70.8} %
\defval{tab.blinkcnt.e}{70.0} %
\defval{tab.blinkcnt.e.r}{70.0} %
\defval{tab.blinkcnt.f}{71.7} %
\defval{tab.blinkcnt.f.r}{71.7} %
\defval{tab.blinkcnt.g}{74.2} %
\defval{tab.blinkcnt.g.r}{74.2} %
\defval{tab.blinkcnt.stwof}{74.17} %
\defval{tab.blinkcnt.stwof.r}{74.2} %
\defval{tab.blinkcnt.nbase}{74.17} %
\defval{tab.blinkcnt.nbase.r}{74.2} %
\defval{tab.blinkcnt.ng}{70.0} %
\defval{tab.blinkcnt.ng.r}{70.0} %
\defval{tab.blinkcnt.nstwog}{71.67} %
\defval{tab.blinkcnt.nstwog.r}{71.7} %
\defval{tab.blinkcnt.nsthree}{75.0} %
\defval{tab.blinkcnt.nsthree.r}{75.0} %
\defval{tab.blinkdep.base}{90.3} %
\defval{tab.blinkdep.base.r}{90.3} %
\defval{tab.blinkdep.orig}{88.7} %
\defval{tab.blinkdep.orig.r}{88.7} %
\defval{tab.blinkdep.a}{91.9} %
\defval{tab.blinkdep.a.r}{91.9} %
\defval{tab.blinkdep.c}{87.1} %
\defval{tab.blinkdep.c.r}{87.1} %
\defval{tab.blinkdep.d}{87.1} %
\defval{tab.blinkdep.d.r}{87.1} %
\defval{tab.blinkdep.e}{86.3} %
\defval{tab.blinkdep.e.r}{86.3} %
\defval{tab.blinkdep.f}{89.5} %
\defval{tab.blinkdep.f.r}{89.5} %
\defval{tab.blinkdep.g}{87.9} %
\defval{tab.blinkdep.g.r}{87.9} %
\defval{tab.blinkdep.stwof}{86.29} %
\defval{tab.blinkdep.stwof.r}{86.3} %
\defval{tab.blinkdep.nbase}{83.87} %
\defval{tab.blinkdep.nbase.r}{83.9} %
\defval{tab.blinkdep.ng}{89.52} %
\defval{tab.blinkdep.ng.r}{89.5} %
\defval{tab.blinkdep.nstwog}{89.52} %
\defval{tab.blinkdep.nstwog.r}{89.5} %
\defval{tab.blinkdep.nsthree}{87.1} %
\defval{tab.blinkdep.nsthree.r}{87.1} %
\defval{tab.srayes.base}{71} %
\defval{tab.srayes.base.r}{71} %
\defval{tab.srayes.orig}{118} %
\defval{tab.srayes.orig.r}{118} %
\defval{tab.srayes.a}{93} %
\defval{tab.srayes.a.r}{93} %
\defval{tab.srayes.c}{98} %
\defval{tab.srayes.c.r}{98} %
\defval{tab.srayes.d}{80} %
\defval{tab.srayes.d.r}{80} %
\defval{tab.srayes.e}{76} %
\defval{tab.srayes.e.r}{76} %
\defval{tab.srayes.f}{75} %
\defval{tab.srayes.f.r}{75} %
\defval{tab.srayes.g}{75} %
\defval{tab.srayes.g.r}{75} %
\defval{tab.srayes.stwof}{76} %
\defval{tab.srayes.stwof.r}{76} %
\defval{tab.srayes.nbase}{81} %
\defval{tab.srayes.nbase.r}{81} %
\defval{tab.srayes.ng}{79} %
\defval{tab.srayes.ng.r}{79} %
\defval{tab.srayes.nstwog}{77} %
\defval{tab.srayes.nstwog.r}{77} %
\defval{tab.srayes.nsthree}{80} %
\defval{tab.srayes.nsthree.r}{80} %
\defval{tab.arcc.base}{91.89} %
\defval{tab.arcc.base.r}{91.9} %
\defval{tab.arcc.orig}{93.34} %
\defval{tab.arcc.orig.r}{93.3} %
\defval{tab.arcc.a}{93.43} %
\defval{tab.arcc.a.r}{93.4} %
\defval{tab.arcc.c}{93.86} %
\defval{tab.arcc.c.r}{93.9} %
\defval{tab.arcc.d}{93.6} %
\defval{tab.arcc.d.r}{93.6} %
\defval{tab.arcc.e}{93.0} %
\defval{tab.arcc.e.r}{93.0} %
\defval{tab.arcc.f}{93.69} %
\defval{tab.arcc.f.r}{93.7} %
\defval{tab.arcc.g}{94.2} %
\defval{tab.arcc.g.r}{94.2} %
\defval{tab.arcc.stwof}{94.03} %
\defval{tab.arcc.stwof.r}{94.0} %
\defval{tab.arcc.nbase}{93.77} %
\defval{tab.arcc.nbase.r}{93.8} %
\defval{tab.arcc.ng}{94.2} %
\defval{tab.arcc.ng.r}{94.2} %
\defval{tab.arcc.nstwog}{94.28} %
\defval{tab.arcc.nstwog.r}{94.3} %
\defval{tab.arcc.nsthree}{94.2} %
\defval{tab.arcc.nsthree.r}{94.2} %
\defval{tab.arce.base}{97.81} %
\defval{tab.arce.base.r}{97.8} %
\defval{tab.arce.orig}{97.39} %
\defval{tab.arce.orig.r}{97.4} %
\defval{tab.arce.a}{97.81} %
\defval{tab.arce.a.r}{97.8} %
\defval{tab.arce.c}{98.06} %
\defval{tab.arce.c.r}{98.1} %
\defval{tab.arce.d}{97.94} %
\defval{tab.arce.d.r}{97.9} %
\defval{tab.arce.e}{97.73} %
\defval{tab.arce.e.r}{97.7} %
\defval{tab.arce.f}{98.02} %
\defval{tab.arce.f.r}{98.0} %
\defval{tab.arce.g}{97.9} %
\defval{tab.arce.g.r}{97.9} %
\defval{tab.arce.stwof}{97.98} %
\defval{tab.arce.stwof.r}{98.0} %
\defval{tab.arce.nbase}{98.15} %
\defval{tab.arce.nbase.r}{98.2} %
\defval{tab.arce.ng}{98.23} %
\defval{tab.arce.ng.r}{98.2} %
\defval{tab.arce.nstwog}{98.15} %
\defval{tab.arce.nstwog.r}{98.2} %
\defval{tab.arce.nsthree}{98.02} %
\defval{tab.arce.nsthree.r}{98.0} %
\defval{tab.hella.base}{87.56} %
\defval{tab.hella.base.r}{87.6} %
\defval{tab.hella.orig}{89.45} %
\defval{tab.hella.orig.r}{89.5} %
\defval{tab.hella.a}{88.64} %
\defval{tab.hella.a.r}{88.6} %
\defval{tab.hella.c}{88.62} %
\defval{tab.hella.c.r}{88.6} %
\defval{tab.hella.d}{88.02} %
\defval{tab.hella.d.r}{88.0} %
\defval{tab.hella.e}{88.08} %
\defval{tab.hella.e.r}{88.1} %
\defval{tab.hella.f}{88.85} %
\defval{tab.hella.f.r}{88.9} %
\defval{tab.hella.g}{89.09} %
\defval{tab.hella.g.r}{89.1} %
\defval{tab.hella.stwof}{88.72} %
\defval{tab.hella.stwof.r}{88.7} %
\defval{tab.hella.nbase}{88.67} %
\defval{tab.hella.nbase.r}{88.7} %
\defval{tab.hella.ng}{90.69} %
\defval{tab.hella.ng.r}{90.7} %
\defval{tab.hella.nstwog}{90.29} %
\defval{tab.hella.nstwog.r}{90.3} %
\defval{tab.hella.nsthree}{90.11} %
\defval{tab.hella.nsthree.r}{90.1} %
\defval{tab.wino.base}{70.17} %
\defval{tab.wino.base.r}{70.2} %
\defval{tab.wino.orig}{73.01} %
\defval{tab.wino.orig.r}{73.0} %
\defval{tab.wino.a}{73.01} %
\defval{tab.wino.a.r}{73.0} %
\defval{tab.wino.c}{73.8} %
\defval{tab.wino.c.r}{73.8} %
\defval{tab.wino.d}{72.38} %
\defval{tab.wino.d.r}{72.4} %
\defval{tab.wino.e}{72.38} %
\defval{tab.wino.e.r}{72.4} %
\defval{tab.wino.f}{74.11} %
\defval{tab.wino.f.r}{74.1} %
\defval{tab.wino.g}{73.4} %
\defval{tab.wino.g.r}{73.4} %
\defval{tab.wino.stwof}{72.77} %
\defval{tab.wino.stwof.r}{72.8} %
\defval{tab.wino.nbase}{72.53} %
\defval{tab.wino.nbase.r}{72.5} %
\defval{tab.wino.ng}{75.37} %
\defval{tab.wino.ng.r}{75.4} %
\defval{tab.wino.nstwog}{74.27} %
\defval{tab.wino.nstwog.r}{74.3} %
\defval{tab.wino.nsthree}{75.06} %
\defval{tab.wino.nsthree.r}{75.1} %
\defval{tab.boolq.base}{87.22} %
\defval{tab.boolq.base.r}{87.2} %
\defval{tab.boolq.orig}{88.5} %
\defval{tab.boolq.orig.r}{88.5} %
\defval{tab.boolq.a}{89.02} %
\defval{tab.boolq.a.r}{89.0} %
\defval{tab.boolq.c}{88.44} %
\defval{tab.boolq.c.r}{88.4} %
\defval{tab.boolq.d}{88.69} %
\defval{tab.boolq.d.r}{88.7} %
\defval{tab.boolq.e}{88.62} %
\defval{tab.boolq.e.r}{88.6} %
\defval{tab.boolq.f}{88.87} %
\defval{tab.boolq.f.r}{88.9} %
\defval{tab.boolq.g}{88.78} %
\defval{tab.boolq.g.r}{88.8} %
\defval{tab.boolq.stwof}{88.93} %
\defval{tab.boolq.stwof.r}{88.9} %
\defval{tab.boolq.nbase}{88.62} %
\defval{tab.boolq.nbase.r}{88.6} %
\defval{tab.boolq.ng}{89.36} %
\defval{tab.boolq.ng.r}{89.4} %
\defval{tab.boolq.nstwog}{88.38} %
\defval{tab.boolq.nstwog.r}{88.4} %
\defval{tab.boolq.nsthree}{88.93} %
\defval{tab.boolq.nsthree.r}{88.9} %
\defval{tab.gold.above}{600} %
\defval{tab.rows.above}{1,384} %
\defval{tab.gold.sr}{77} %
\defval{tab.rows.sr}{143} %
\defval{obj.satdev.base}{75.75} %
\defval{obj.satdev.paired}{89.20} %
\defval{obj.satdev.ce}{88.10} %
\defval{obj.satdev.brier}{88.05} %
\defval{obj.satval.base}{50.24} %
\defval{obj.satval.paired}{51.46} %
\defval{obj.satval.ce}{52.20} %
\defval{obj.satval.brier}{55.93} %
\defval{obj.cv.base}{85.22} %
\defval{obj.cv.paired}{85.44} %
\defval{obj.cv.ce}{84.64} %
\defval{obj.cv.brier}{84.94} %
\defval{obj.blink.base}{80.55} %
\defval{obj.blink.paired}{76.42} %
\defval{obj.blink.ce}{72.10} %
\defval{obj.blink.brier}{73.28} %
\defval{obj.tdsr.base}{54.04} %
\defval{obj.tdsr.paired}{52.30} %
\defval{obj.tdsr.ce}{51.66} %
\defval{obj.tdsr.brier}{51.81} %
\defval{obj.n}{50,000} %
\defval{obj.steps}{782} %
\defval{s2.scenes}{696} %
\defval{s2.frames}{4,072} %
\defval{s2.qa}{377,956} %
\defval{s2.split.tr}{558} %
\defval{s2.split.dev}{69} %
\defval{s2.split.te}{69} %
\defval{s2.obj}{108,192} %
\defval{s2.ego}{19,564} %
\defval{s2.box}{13,783} %
\defval{s2.rel}{14,270} %
\defval{s2.t.sign}{16,932} %
\defval{s2.t.path}{16,932} %
\defval{s2.t.consider}{16,932} %
\defval{s2.t.appear}{16,956} %
\defval{s2.t.moving}{15,133} %
\defval{s2.t.future}{15,024} %
\defval{s2.t.motion}{8,593} %
\defval{s2.t.elem}{1,690} %
\defval{s2.t.target}{3,928} %
\defval{s2.t.steer}{3,953} %
\defval{s2.t.speed}{3,953} %
\defval{s2.t.signal}{3,777} %
\defval{s2.t.inview}{3,953} %
\defval{s2.rows}{124,378} %
\defval{s2.pool}{1,240,040} %
\defval{s2.devrows}{48,526} %
\defval{s2.test}{6,571} %
\defval{s2.cap}{400} %
\defval{s2.devper}{150} %
\defval{s2.skip}{450} %
\defval{s2.ph.qa}{175} %
\defval{s2.ph.box}{150} %
\defval{s2.mix.drive}{70} %
\defval{s2.mix.obj}{30} %
\defval{s2.mix.ego}{14} %
\defval{s2.mix.box}{12} %
\defval{s2.mix.rel}{14} %
\defval{s2.mix.sone}{30} %
\defval{s2.steps}{3,000} %
\defval{s2.merge.big}{99.5} %
\defval{s2.merge.views}{600} %
\defval{s3.steps}{6,000} %
\defval{s3.mix.nav}{60} %
\defval{s3.mix.lone}{30} %
\defval{s3.mix.ltwo}{30} %
\defval{s3.mix.drivelm}{20} %
\defval{s3.mix.sone}{20} %
\defval{s3.rl.lr}{$2{\times}10^{-5}$} %
\defval{s3.rl.steps}{2,000} %
\defval{s3.rl.anchor}{0.3} %
\defval{s3.rl.warm}{5} %
\defval{s3.navtrain}{103,288} %
\defval{s3.heldout}{2,000} %
\defval{s3.navtest}{12,146} %
\defval{s3.pool.train}{1,506,951} %
\defval{s3.pool.dev}{6,465} %
\defval{s3.rows.one}{92,339} %
\defval{s3.rows.two}{174,572} %
\defval{s3.passes}{0.86} %
\defval{s3.passes.one}{1.25} %
\defval{s3.passes.two}{0.66} %
\defval{s3.egotree}{70.8} %
\defval{s3.dead}{6.9} %
\defval{s3.good}{4.6} %
\defval{s3.halfdead}{37} %
\defval{nav.vone.ho}{45.03} %
\defval{nav.vone.nt}{43.92} %
\defval{nav.vone.nc}{79.8} %
\defval{nav.vone.dac}{71.4} %
\defval{nav.vone.ep}{41.5} %
\defval{nav.vone.ttc}{68.9} %
\defval{nav.vone.ddc}{90.9} %
\defval{nav.vtwo.ho}{80.54} %
\defval{nav.vtwo.nt}{77.57} %
\defval{nav.vtwo.nc}{95.2} %
\defval{nav.vtwo.dac}{89.1} %
\defval{nav.vtwo.ep}{73.6} %
\defval{nav.vtwo.ttc}{88.2} %
\defval{nav.vtwo.ddc}{93.7} %
\defval{nav.vthree.ho}{82.56} %
\defval{nav.vthree.nt}{81.36} %
\defval{nav.vthree.nc}{95.8} %
\defval{nav.vthree.dac}{92.5} %
\defval{nav.vthree.ep}{77.3} %
\defval{nav.vthree.ttc}{89.2} %
\defval{nav.vthree.ddc}{95.9} %
\defval{nav.vfour.ho}{79.41} %
\defval{nav.vfour.nt}{78.14} %
\defval{nav.vfour.nc}{94.3} %
\defval{nav.vfour.dac}{91.2} %
\defval{nav.vfour.ep}{74.9} %
\defval{nav.vfour.ttc}{86.0} %
\defval{nav.vfour.ddc}{94.9} %
\defval{nav.rl.ho}{83.98} %
\defval{nav.rl.nt}{83.24} %
\defval{nav.rl.nc}{96.6} %
\defval{nav.rl.dac}{93.9} %
\defval{nav.rl.ep}{78.2} %
\defval{nav.rl.ttc}{90.9} %
\defval{nav.rl.ddc}{96.6} %
\defval{nav.vtwo.hotwo}{79.38} %
\defval{nav.vthree.hoa}{82.57} %
\defval{nav.vthree.nta}{81.44} %
\defval{nav.vone.acc}{29.0 / 8.3 / 1.3} %
\defval{nav.vtwo.acc}{78.0 / 53.5 / 11.3} %
\defval{nav.vthree.acc}{76.7 / 55.1} %
\defval{nav.vfour.acc}{58.0 / 61.7} %
\defval{nav.rl.ckone}{82.95} %
\defval{nav.rl.cktwo}{83.93} %
\defval{nav.rl.gain}{1.88} %
\defval{nav.nine.ho}{85.02} %
\defval{nav.nine.nt}{83.91} %
\defval{nav.nine.honc}{97.05} %
\defval{nav.nine.hodac}{94.35} %
\defval{nav.nine.hoep}{80.91} %
\defval{nav.nine.hottc}{91.45} %
\defval{nav.nine.hoddc}{96.22} %
\defval{nav.nine.dho}{2.47} %
\defval{nav.nine.dnt}{2.55} %
\defval{nav.nine.dho.ci}{[+1.39, +3.54]} %
\defval{nav.nine.dnt.ci}{[+2.10, +3.00]} %
\defval{nav.nine.redecode}{85.11} %
\defval{nav.env.d}{0.09} %
\defval{nav.env.ci}{[$-$0.15, +0.32]} %
\defval{nav.tp.p05}{80.97} %
\defval{nav.tp.p10}{81.93} %
\defval{nav.tp.p15}{80.77} %
\defval{nav.tp.p20}{82.87} %
\defval{nav.tp.p25}{82.62} %
\defval{nav.tp.p30}{84.08} %
\defval{nav.tp.p35}{84.72} %
\defval{nav.tp.p40}{85.51} %
\defval{nav.tp.p45}{84.67} %
\defval{nav.tp.p50}{85.17} %
\defval{nav.tp.p55}{84.47} %
\defval{nav.tp.p60}{85.74} %
\defval{nav.tp.p65}{85.59} %
\defval{nav.tp.p70}{86.17} %
\defval{nav.tp.p75}{86.06} %
\defval{nav.tp.p80}{86.55} %
\defval{nav.tp.p85}{86.62} %
\defval{nav.tp.p90}{86.67} %
\defval{nav.tp.bestpass}{10.0} %
\defval{nav.tp.dnine}{1.56} %
\defval{nav.tp.dnine.ci}{[+0.55, +2.62]} %
\defval{nav.tp.dnine.cilog}{[+0.32, +2.84]} %
\defval{nav.tp.deight}{1.44} %
\defval{nav.tp.deight.ci}{[+0.36, +2.55]} %
\defval{nav.tp.deight.cilog}{[+0.10, +2.90]} %
\defval{nav.tp.honc}{97.4} %
\defval{nav.tp.hodac}{94.7} %
\defval{nav.tp.hoep}{81.23} %
\defval{nav.tp.hottc}{92.0} %
\defval{nav.tp.hoddc}{97.3} %
\defval{nav.tp.lrone}{$2{\times}10^{-4}$} %
\defval{nav.tp.lrtwo}{$10^{-4}$} %
\defval{nav.tp.share}{80} %
\defval{nav.tp.steps}{52,100} %
\defval{nav.tp.visits}{9.99} %
\defval{nav.tp.divpass}{3.5} %
\defval{nav.tp.gradnorm}{542} %
\defval{nav.tp.restart}{3.0} %
\defval{nav.tp.passes}{10} %
\defval{nav.tp.dsix}{0.63} %
\defval{nav.tp.dsix.ci}{[$-$0.40, +1.66]} %
\defval{nav.tp.dfour}{0.40} %
\defval{nav.tp.dfour.ci}{[$-$0.66, +1.43]} %
\defval{nav.tp.ce}{1.529} %
\defval{nav.tp.cepass}{5.7} %
\defval{nav.tp.offsix}{4.70} %
\defval{nav.tpf.ho}{86.72} %
\defval{nav.tpf.nt}{85.41} %
\defval{nav.tpf.nc}{97.22} %
\defval{nav.tpf.dac}{95.24} %
\defval{nav.tpf.ep}{80.99} %
\defval{nav.tpf.ttc}{91.51} %
\defval{nav.tpf.ddc}{97.72} %
\defval{nav.tpr.ho}{87.31} %
\defval{nav.tpr.nt}{86.41} %
\defval{nav.tpr.nc}{97.29} %
\defval{nav.tpr.dac}{96.03} %
\defval{nav.tpr.ep}{81.81} %
\defval{nav.tpr.ttc}{91.96} %
\defval{nav.tpr.ddc}{97.82} %
\defval{nav.tp.dten}{1.61} %
\defval{nav.tp.dten.ci}{[+0.55, +2.63]} %
\defval{nav.tpr.dnt}{1.00} %
\defval{ret.tp.mmlu}{45.74} %
\defval{ret.tp.mmlu.r}{45.7} %
\defval{dev.big.one}{51.5} %
\defval{dev.big.six}{62.7} %
\defval{dev.nine.one}{57.0} %
\defval{dev.nine.six}{66.2} %
\defval{ceil.logged}{95.05} %
\defval{ceil.osixteen}{84.57} %
\defval{ceil.otwofivesix}{98.97} %
\defval{ceil.oleaves}{99.71} %
\defval{ceil.snapleaf}{79.93} %
\defval{ceil.snaptwo}{74.11} %
\defval{ceil.target}{95.08} %
\defval{ceil.targetddc}{99.2} %
\defval{ceil.agree}{81} %
\defval{ceil.nobeta}{97.55} %
\defval{ceil.nobetaagree}{37.5} %
\defval{fit.sft.tr}{86.08} %
\defval{fit.sft.ho}{82.70} %
\defval{fit.rl.tr}{87.61} %
\defval{fit.rl.ho}{83.89} %
\defval{fit.or.tr}{99.31} %
\defval{fit.or.ho}{98.94} %
\defval{fit.sft.d}{3.38} %
\defval{fit.sft.ci}{[+1.62, +5.18]} %
\defval{fit.rl.d}{3.71} %
\defval{fit.rl.ci}{[+2.07, +5.36]} %
\defval{fit.n}{2,000} %
\defval{fail.b.nc.n}{63} %
\defval{fail.b.nc.l}{3.10} %
\defval{fail.b.dac.n}{105} %
\defval{fail.b.dac.l}{5.25} %
\defval{fail.b.ddc.n}{101} %
\defval{fail.b.ddc.l}{0.57} %
\defval{fail.b.ttc.n}{90} %
\defval{fail.b.ttc.l}{1.96} %
\defval{fail.b.slow.n}{1,641} %
\defval{fail.b.slow.l}{4.10} %
\defval{fail.c.nc.n}{53} %
\defval{fail.c.nc.l}{2.63} %
\defval{fail.c.dac.n}{104} %
\defval{fail.c.dac.l}{5.20} %
\defval{fail.c.ddc.n}{75} %
\defval{fail.c.ddc.l}{0.40} %
\defval{fail.c.ttc.n}{94} %
\defval{fail.c.ttc.l}{2.04} %
\defval{fail.c.slow.n}{1,674} %
\defval{fail.c.slow.l}{4.21} %
\defval{fail.inview}{4.80} %
\defval{fail.outview}{0.40} %
\defval{fail.straight}{2.80} %
\defval{fail.turn}{2.15} %
\defval{fail.fixed}{39} %
\defval{fail.new}{32} %
\defval{rs.prop}{84.05} %
\defval{rs.texteight}{82.68} %
\defval{rs.texteight.d}{$-$1.38} %
\defval{rs.vtwotext}{82.46} %
\defval{rs.vtwotext.d}{$-$1.60} %
\defval{rs.vtwodrawn}{82.37} %
\defval{rs.vtwodrawn.d}{$-$1.68} %
\defval{rs.aligned}{81.42} %
\defval{rs.aligned.d}{$-$2.63} %
\defval{rs.ktwo}{83.07} %
\defval{rs.ktwo.d}{$-$0.98} %
\defval{rs.combo}{84.25} %
\defval{rs.combo.d}{+0.19} %
\defval{peer.autovla.sft}{80.54} %
\defval{peer.autovla.fin}{89.11} %
\defval{peer.recog.sft}{86.5} %
\defval{peer.recog.fin}{90.8} %
\defval{peer.recog.ep}{83.6} %
\defval{peer.wzero.sft}{88.4} %
\defval{peer.wzero.fin}{90.2} %
\defval{peer.wzero.ep}{86.1} %
\defval{peer.ada.sft}{86.2} %
\defval{peer.ada.fin}{90.3} %
\defval{peer.curious.sft}{87.6} %
\defval{peer.curious.fin}{90.3} %
\defval{peer.curious.ep}{85.4} %
\defval{peer.teach.sft}{88.2} %
\defval{peer.teach.fin}{90.4} %
\defval{peer.teach.ep}{85.4} %
\defval{peer.fine.sft}{86.7} %
\defval{peer.fine.fin}{90.7} %
\defval{peer.fine.ep}{86.7} %
\defval{peer.wam.sft}{86.7} %
\defval{peer.wam.fin}{90.3} %
\defval{peer.wam.ep}{84.7} %
\defval{peer.qd.sft}{88.2} %
\defval{peer.qd.fin}{90.7} %
\defval{peer.fine.epfix}{89.7} %
\defval{ref.egomlp}{65.6} %
\defval{ref.transfuser}{84.0} %
\defval{ref.human}{94.8} %
\defval{ref.drivor}{93.1} %
\defval{ref.ztrs.epo}{85.3} %
\defval{ref.ztrs.rf}{75.0} %
\defval{epdms.vone}{\pend} %
\defval{epdms.vtwo}{\pend} %
\defval{epdms.vthree}{\pend} %
\defval{epdms.vfour}{\pend} %
\defval{epdms.rl}{\pend} %
\defval{epdms.nine}{\pend} %
\defval{epdms.tpf}{\pend} %
\defval{epdms.tpr}{\pend} %
\defval{epdms.nine.ho}{\pend} %
\defval{epdms.tpfour.ho}{\pend} %
\defval{ci.level}{95} %
\defval{q2.dst}{+8.32} %
\defval{q2.dst.ci}{[+7.40, +9.18]} %
\defval{q2.dall}{+8.75} %
\defval{q2.dall.ci}{[+7.85, +9.69]} %
\defval{rs.texteight.ci}{[$-$2.41, $-$0.35]} %
\defval{rs.vtwotext.ci}{[$-$2.61, $-$0.56]} %
\defval{rs.vtwodrawn.ci}{[$-$2.65, $-$0.72]} %
\defval{rs.aligned.ci}{[$-$3.85, $-$1.48]} %
\defval{rs.ktwo.ci}{[$-$2.05, +0.08]} %
\defval{rs.combo.ci}{[$-$0.20, +0.59]} %
\defval{s2.k.sign}{2} %
\defval{s2.k.path}{2} %
\defval{s2.k.consider}{2} %
\defval{s2.k.appear}{4} %
\defval{s2.k.moving}{2} %
\defval{s2.k.future}{5} %
\defval{s2.k.motion}{3} %
\defval{s2.k.elem}{5} %
\defval{s2.k.target}{4} %
\defval{s2.k.steer}{5} %
\defval{s2.k.speed}{5} %
\defval{s2.k.signal}{5} %
\defval{s2.k.inview}{2} %
\defval{s2.k.box}{2} %
\defval{s2.k.rel}{2} %

\newcommand{\pend}{{\color{BurntOrange}\scriptsize pending}}

\newcommand{\method}{SuperGlance\xspace}
\newcommand{\nineb}{Qwen3.5-9B\xspace}
\newcommand{\tq}[1]{`#1'}   %

\usepackage{colortbl}
\definecolor{jdmain}{HTML}{4B44C6}
\definecolor{jdours}{HTML}{FDECEA}
\AtBeginDocument{\hypersetup{citecolor=black,linkcolor=black,urlcolor=black}}   %
\newcommand{\ours}{\bfseries}   %
\definecolor{jdgroup}{HTML}{F1EEEA}   %
\newcommand{\grouprow}{\rowcolor{jdgroup}}   %
\definecolor{jdrank1}{HTML}{B8DDFF}   %
\definecolor{jdrank2}{HTML}{E6F4FF}   %
\definecolor{jdheat}{HTML}{A1D1F8}     %
\definecolor{jdheatlo}{HTML}{FFFFFF}   %
\newcommand{\heat}[2]{\cellcolor{jdheat!\the\numexpr#1*#1/100\relax!jdheatlo}#2}   %
\newcommand{\heatin}[2]{{\setlength{\fboxsep}{0pt}\colorbox{jdheat!\the\numexpr#1*#1/100\relax!jdheatlo}{\hspace{0.8pt}\strut#2\hspace{0.8pt}}}}   %
\let\heatv\heat \let\heatvin\heatin   %
\newcommand{\na}{\textcolor{black!35}{--}}   %
\definecolor{jdyes}{HTML}{2E7D4F}\definecolor{jdno}{HTML}{D63B32}   %
\usepackage{pdfrender}   %
\newcommand{\markw}[1]{\textpdfrender{TextRenderingMode=FillStroke,LineWidth=0.25pt,LineJoinStyle=Round}{#1}}
\newcommand{\cmark}{\textcolor{jdyes}{\markw{\checkmark}}}
\newcommand{\xmark}{\textcolor{jdno}{\markw{$\times$}}}
\usepackage{siunitx}
\newcommand{\vone}[1]{\num[mode=text,reset-text-family=false,reset-text-series=false,reset-text-shape=false,round-mode=places,round-precision=1]{\val{#1}}}   %
\newcommand{\vtwo}[1]{\num[mode=text,reset-text-family=false,reset-text-series=false,reset-text-shape=false,round-mode=places,round-precision=2]{\val{#1}}}   %
\makeatletter\AtBeginDocument{\let\SG@backref\BR@backref\def\BR@backref#1{\unskip\nobreak\SG@backref{#1}}}\makeatother   %
\providecommand{\suppword}{supplement}   %

\providecommand{\sgtable}[2]{\begin{table}[#1]}   %
\providecommand{\sgcvpr}[1]{#1}\providecommand{\sgarxiv}[1]{}   %

\defval{r.k.total}{5,121}   %
\defval{r.k.km}{4,096}      %
\defval{r.k.fit}{92,339}    %
\defval{r.k.fast}{1,024}    %
\defval{r.k.fastrows}{71,090} %
\defval{r.k.fastv}{13}      %
\defval{r.k.horizon}{4}     %
\defval{r.k.hz}{2}          %
\defval{r.k.poses}{8}       %
\defval{r.k.gap}{0.40}      %
\defval{r.k.bestho}{99.84}  %
\defval{r.k.nho}{10,549}    %
\defval{r.k.nearnav}{85.69} %
\defval{r.k.nearwod}{8.01}  %
\defval{r.k.bestwod}{9.55}  %
\defval{r.in.nhist}{three}  %
\defval{r.in.gap}{0.5}      %
\defval{r.in.past}{1.5}     %
\defval{r.in.histw}{544}    %
\defval{r.in.histh}{320}    %
\defval{r.in.maxpix}{589,824} %
\defval{r.par.total}{9.75}  %
\defval{r.par.head}{21.6}   %
\defval{r.par.lora}{80}     %

\defval{r.d.s1}{1.16}       %
\defval{r.d.s2}{124,378}  %
\defval{r.d.sft}{2,191,689} %
\defval{r.d.sftdraw}{2.68}  %
\defval{r.d.dag}{46,583}    %
\defval{r.d.dagrow}{2.1}    %
\defval{r.d.dagw}{4.5}      %
\defval{r.d.dagsc}{492}     %
\defval{r.d.wodft}{61,460}  %
\defval{r.t.ftsteps}{2,500} %
\defval{r.t.ftlr}{$5\times10^{-5}$} %
\defval{r.t.ftwod}{40}      %
\defval{r.t.tft}{0.5}       %
\defval{r.d.nav}{258,580}   %
\defval{r.d.pai}{1,210,514} %
\defval{r.d.wod}{466,205}   %
\defval{r.d.qa}{256,390}    %
\defval{r.d.rl}{391,322}    %
\defval{r.d.rlnav}{92,339}  %
\defval{r.d.rlpai}{298,504} %
\defval{r.d.rlwod}{479}     %

\defval{r.t.rank}{32}
\defval{r.t.alpha}{64}
\defval{r.t.sftsteps}{20,900}
\defval{r.t.batch}{128}
\defval{r.t.lr}{$10^{-4}$}
\defval{r.t.warm}{5}
\defval{r.t.wd}{0.01}
\defval{r.t.kl}{0.1}        %
\defval{r.t.gpuh}{145.0}    %
\defval{r.t.gpuhft}{17}\defval{r.t.gpuhrl}{8}   %
\defval{r.t.rlsteps}{2,000}
\defval{r.t.rlbatch}{64}
\defval{r.t.rllr}{$5\times10^{-5}$}
\defval{r.t.beta}{0.3}
\defval{r.t.mix}{0.5\,/\,0.45\,/\,0.05} %
\defval{r.t.tnav}{0.1}      %
\defval{r.t.adew}{0.1}      %
\defval{r.t.tpai}{0.5}      %
\defval{r.t.twod}{1}        %
\defval{r.t.pen}{10}        %

\defval{r.l.ours}{84}       %
\defval{r.l.qd}{518}        %
\defval{r.l.qdr}{1,213}     %
\defval{r.l.autovla}{1,529} %
\defval{r.l.text}{5,558}    %
\defval{r.l.recog}{199}     %
\defval{r.l.oursfirst}{83}  %
\defval{r.l.qdfirst}{187}   %
\defval{r.l.qdexpert}{331}  %
\defval{r.l.head}{0.2}
\defval{r.l.qdread}{187}   %
\defval{r.l.step}{33--35}   %
\defval{r.l.qdpass}{11}
\defval{r.l.qdrpass}{28}
\defval{r.l.avpass}{45}
\defval{r.l.textpass}{155}  %
\defval{r.l.recogpass}{6}
\defval{r.l.xqd}{6.2}       %
\defval{r.l.xqdw}{six}      %

\defval{r.cl.base.km}{1.486}\defval{r.cl.base.pass}{73.4}\defval{r.cl.base.ade}{0.400}   %
\defval{r.cl.nh.nav}{90.83}\defval{r.cl.nh.km}{0.993}\defval{r.cl.nh.pass}{65.3}\defval{r.cl.nh.d}{\textminus{}0.104 (\textminus{}0.134, \textminus{}0.076)}   %
\defval{r.cl.reg.nav}{87.82}\defval{r.cl.reg.km}{1.360}\defval{r.cl.reg.pass}{70.3}\defval{r.cl.reg.d}{\textminus{}0.026 (\textminus{}0.055, +0.004)}\defval{r.cl.reg.ade}{0.320}   %
\defval{r.cl.ne.km}{0.929}\defval{r.cl.ne.pass}{64.3}\defval{r.cl.ne.d}{\textminus{}0.090 (\textminus{}0.114, \textminus{}0.064)}   %
\defval{r.cl.nr.nav}{91.35}\defval{r.cl.nr.km}{0.958}\defval{r.cl.nr.pass}{71.3}\defval{r.cl.nr.d}{\textminus{}0.027 (\textminus{}0.058, +0.002)}   %
\defval{r.cl.im.km}{1.199}\defval{r.cl.im.pass}{71.0}\defval{r.cl.im.d}{\textminus{}0.030 (\textminus{}0.052, \textminus{}0.010)}   %
\defval{r.cl.nw.nav}{91.48}\defval{r.cl.nw.km}{1.470}\defval{r.cl.nw.pass}{73.6}\defval{r.cl.nw.d}{+0.004 (\textminus{}0.018, +0.025)}   %
\defval{r.cl.ns.nav}{91.48}\defval{r.cl.ns.km}{1.425}\defval{r.cl.ns.pass}{74.9}\defval{r.cl.ns.d}{+0.014 (\textminus{}0.010, +0.038)}   %
\defval{r.cl.np.nav}{90.95}\defval{r.cl.np.km}{0.826}\defval{r.cl.np.pass}{75.2}\defval{r.cl.np.d}{\textminus{}0.115 (\textminus{}0.148, \textminus{}0.081)}\defval{r.cl.np.prog}{64.1}\defval{r.cl.np.rear}{33.8}\defval{r.cl.base.prog}{88.1}\defval{r.cl.base.rear}{4.7}   %
\defval{r.cl.nq.nav}{91.30}\defval{r.cl.nq.km}{1.361}\defval{r.cl.nq.pass}{73.4}\defval{r.cl.nq.d}{+0.001 (\textminus{}0.024, +0.026)}   %
\defval{r.cl.nh.ade}{0.397}\defval{r.cl.nr.ade}{0.393}\defval{r.cl.im.ade}{0.395}\defval{r.cl.nw.ade}{0.389}\defval{r.cl.np.ade}{1.048}\defval{r.cl.ns.ade}{0.393}\defval{r.cl.nq.ade}{0.389}   %
\defval{r.q.sft.dl}{79.07}\defval{r.q.sft.sr}{57.65}\defval{r.q.f1.dlv}{75.87}\defval{r.q.f1.srv}{53.29}\defval{r.q.f6.dlv}{43.33}\defval{r.q.f6.srv}{44.85}   %
\defval{r.cl.af.base}{37}\defval{r.cl.af.nh}{73}\defval{r.cl.af.nr}{88}   %
\defval{r.cl.st.base}{10}\defval{r.cl.st.nh}{44}\defval{r.cl.mv.base}{18}\defval{r.cl.mv.nr}{59}   %
\defval{r.cl.rw.nav}{9.27}\defval{r.cl.rw.d}{0.030}   %
\defval{r.cl.inv.nav}{8.8}\defval{r.cl.inv.d}{0.075}\defval{r.cl.inv.ci}{0.047 to 0.101}   %
\defval{r.q.f1.dl}{3.20}\defval{r.q.f1.sr}{4.36}   %
\defval{r.q.cv.dec}{86.47}\defval{r.q.bl.dec}{76.62}\defval{r.q.sr.dec}{57.09}\defval{r.q.dl.dec}{60.95}\defval{r.q.ml.dec}{85.21}\defval{r.q.mp.dec}{73.93}\defval{r.q.ar.dec}{96.16}\defval{r.q.hs.dec}{89.49}\defval{r.q.wg.dec}{85.64}\defval{r.q.bq.dec}{88.29}\defval{r.q.tq.dec}{76.25}   %
\defval{r.q.dec.srgap}{11.70}\defval{r.q.dec.dl}{24.56}\defval{r.q.dec.mp}{17.08}\defval{r.q.dec.wg}{9.24}\defval{r.q.dec.ml}{6.60}\defval{r.q.dec.once}{1.4}   %
\defval{r.cl.txt.nav}{87.23}\defval{r.cl.txt.dnav}{4.07}   %
\defval{r.q.f6.dl}{35.74}\defval{r.q.f6.sr}{12.80}\defval{r.q.f6.nav}{91.30}\defval{r.q.f6.pai}{0.01}\defval{r.q.f6.as}{0.006}   %
\defval{r.cl.reg.xqd}{2.6}\defval{r.cl.reg.dnav}{3.48}   %
\defval{r.t.meansft}{1.21}\defval{r.t.meanrl}{0.23}   %
\defval{r.cl.xqd}{1.7}   %
\defval{r.q.cv.rl}{87.40}\defval{r.q.bl.rl}{79.37}\defval{r.q.sr.rl}{58.29}\defval{r.q.dl.rl}{85.51}\defval{r.q.ml.rl}{78.61}\defval{r.q.mp.rl}{56.85}\defval{r.q.ar.rl}{95.31}\defval{r.q.hs.rl}{90.15}\defval{r.q.wg.rl}{76.40}\defval{r.q.bq.rl}{89.42}\defval{r.q.tq.rl}{65.61}
\defval{r.l.fast}{73}   %
\defval{r.l.xtext}{66}      %
\defval{r.l.e2e.ours}{145.5}
\defval{r.l.e2e.qd}{1,014.1}
\defval{r.l.e2e.qdr}{1,710.1}
\defval{r.l.e2e.recog}{296.5}
\defval{r.l.e2e.av}{1,900.3}
\defval{r.l.e2e.text}{5,622.4}
\defval{r.l.pre.ours}{61.7}
\defval{r.l.pre.qd}{492.7}
\defval{r.l.pre.recog}{97.6}

\defval{r.nav.n}{12,146}
\defval{r.nav.sft.nc}{98.73}   %
\defval{r.nav.sft.dac}{98.25}   %
\defval{r.nav.sft.ep}{86.18}
\defval{r.nav.sft.ttc}{95.56}
\defval{r.nav.sft.c}{100.00}
\defval{r.nav.sft}{91.24}          %
\defval{r.nav.sft.top}{95.37}  %
\defval{r.nav.qd.sft}{87.8}    %
\defval{r.nav.qd.sftr}{88.2}
\defval{r.nav.qd.rl}{90.7}
\defval{r.nav.qd.rlbest}{91.4} %
\defval{r.nav.dsft}{3.0}       %

\defval{r.wod.sft}{7.978}      %
\defval{r.wod.sft.ci}{7.800--8.156}
\defval{r.wod.qd.sftrun}{7.962} %
\defval{r.wod.dsft}{0.02}       %
\defval{r.wod.rl}{8.466}        %
\defval{r.wod.qd.rlrun}{8.446}  %
\defval{r.wod.drl}{0.020}       %
\defval{r.wod.sft.at}{1.094}     %
\defval{r.wod.sft.af}{2.356}
\defval{r.wod.qd.sftrun.at}{0.989}  %
\defval{r.wod.qd.sftrun.af}{2.311}
\defval{r.wod.rl.at}{1.101}      %
\defval{r.wod.rl.af}{2.291}
\defval{r.wod.qd.rlrun.at}{0.617} %
\defval{r.wod.qd.rlrun.af}{1.273}
\defval{r.wod.test}{7.987}      %
\defval{r.wod.test.adet}{1.190}
\defval{r.wod.test.adef}{2.546}
\defval{r.wod.qd.test}{7.91}      %
\defval{r.wod.dtest}{0.08}       %

\defval{r.k.rl}{5,273}            %
\defval{r.nav.rl.nc}{99.28}   %
\defval{r.nav.rl.dac}{98.74}
\defval{r.nav.rl.ep}{84.03}
\defval{r.nav.rl.ttc}{97.62}
\defval{r.nav.rl.c}{100.00}
\defval{r.nav.rl.ddc}{98.18}
\defval{r.nav.rl}{91.51}          %
\defval{r.nav.rl.top}{94.80}      %
\defval{r.nav.rl.dqd}{0.81}       %
\defval{r.pai.rl.t}{0.373}        %
\defval{r.pai.rl.f}{1.076}
\defval{r.pai.rl.mt}{0.271}       %
\defval{r.pai.rl.mf}{0.777}
\defval{r.pai.rlval.t}{0.418}     %
\defval{r.pai.rlval.f}{1.141}

\defval{r.pai.topk}{64}
\defval{r.pai.tau}{0.2}
\defval{r.pai.kd3}{0.0009}   %
\defval{r.pai.kd5}{0.0017}   %
\defval{r.pai.k1s}{0.028}    %
\defval{r.pai.k1r}{0.031}    %
\defval{r.pai.sft.t}{0.394}   %
\defval{r.pai.sft.f}{1.109}
\defval{r.pai.sft.mt}{0.284}
\defval{r.pai.sft.mf}{0.821}
\defval{r.pai.qdsft.t}{0.414}   %
\defval{r.pai.qdsft.f}{1.205}
\defval{r.pai.qdsft.mt}{0.385}  %
\defval{r.pai.qdsft.mf}{1.095}
\defval{r.pai.qdrl.t}{0.454}    %
\defval{r.pai.qdrl.f}{1.231}
\defval{r.pai.qdrl.mt}{0.415}
\defval{r.pai.qdrl.mf}{1.102}
\defval{r.pai.dt}{0.020}        %
\defval{r.pai.dmt}{0.101}       %

\defval{r.as.sft.col}{11.2}
\defval{r.as.sft.colf}{6.0}
\defval{r.as.sft.off}{8.8}
\defval{r.as.sft.prog}{87.9}
\defval{r.as.sft.pass}{72.1}
\defval{r.as.sft.km}{0.886}
\defval{r.as.sft.kmf}{1.199}
\defval{r.as.rl.col}{10.6}
\defval{r.as.rl.colf}{5.9}
\defval{r.as.rl.off}{7.6}
\defval{r.as.rl.prog}{88.6}
\defval{r.as.rl.pass}{73.5}
\defval{r.as.rl.km}{0.975}
\defval{r.as.rl.kmf}{1.318}
\defval{r.as.qdsft.col}{32.8}
\defval{r.as.qdsft.colf}{14.4}
\defval{r.as.qdsft.off}{15.4}
\defval{r.as.qdsft.prog}{71.1}
\defval{r.as.qdsft.pass}{58.7}
\defval{r.as.qdsft.km}{0.284}
\defval{r.as.qdsft.kmf}{0.461}
\defval{r.as.qdrl.col}{43.6}
\defval{r.as.qdrl.colf}{15.6}
\defval{r.as.qdrl.off}{13.1}
\defval{r.as.qdrl.prog}{57.3}
\defval{r.as.qdrl.pass}{53.6}
\defval{r.as.qdrl.km}{0.194}
\defval{r.as.qdrl.kmf}{0.383}
\defval{r.as.rlsft.d}{$+0.017$}
\defval{r.as.rlsft.ci}{$-0.004$ to $+0.038$}
\defval{r.dly.kmown}{1.248}\defval{r.dly.passown}{72.2}
\defval{r.dly.km300}{1.224}\defval{r.dly.pass300}{70.9}
\defval{r.dly.kmqd}{1.061}\defval{r.dly.passqd}{66.0}
\defval{r.dly.km}{0.259}\defval{r.dly.pass}{16.2}
\defval{r.dly.km2}{0.212}\defval{r.dly.pass2}{8.4}
\defval{r.dly.km5}{0.192}\defval{r.dly.pass5}{41.5}
\defval{r.dly.qdpkm0}{0.470}\defval{r.dly.qdppass0}{60.0}
\defval{r.dly.qdpkm300}{0.502}\defval{r.dly.qdppass300}{59.9}
\defval{r.dly.col}{10.9}\defval{r.dly.colf}{6.8}\defval{r.dly.off}{7.5}\defval{r.dly.prog}{88.5}\defval{r.dly.kma}{0.970}
\defval{r.dly.qdcol}{26.7}\defval{r.dly.qdcolf}{21.0}\defval{r.dly.qdoff}{36.3}\defval{r.dly.qdprog}{51.8}\defval{r.dly.qdkma}{0.145}\defval{r.dly.qdkm}{0.158}\defval{r.dly.qdpass}{9.3}
\defval{r.dly.qdpcol}{32.6}\defval{r.dly.qdpcolf}{18.1}\defval{r.dly.qdpoff}{19.3}\defval{r.dly.qdpprog}{67.9}\defval{r.dly.qdpkma}{0.229}\defval{r.dly.qdpkm}{0.317}\defval{r.dly.qdppass}{36.6}
\defval{r.dly.qdpkm1200}{0.165}\defval{r.dly.qdppass1200}{9.8}
\defval{r.dly.qdpkm2}{0.150}\defval{r.dly.qdppass2}{3.2}
\defval{r.dly.qdpkm5}{0.158}\defval{r.dly.qdppass5}{33.5}
\defval{r.as.x}{3.9}   %
\defval{r.as.x.ci}{3.3 to 4.7}
\defval{r.dly.lagx}{1.2}   %
\defval{r.dly.lagx.ci}{1.1 to 1.4}
\defval{r.dly.lagx12}{5.1}   %
\defval{r.dly.lagx12.ci}{4.4 to 6.0}
\defval{r.dly.qdlagx}{1.5}   %
\defval{r.dly.qdlagx.ci}{1.3 to 1.7}
\defval{r.as.x0}{2.6}   %
\defval{r.as.x0.ci}{2.2 to 3.1}
\defval{r.as.x0rl}{3.1}   %
\defval{r.as.x0rl.ci}{2.6 to 3.8}
\defval{r.dly.qdx300}{1.07}   %
\defval{r.dly.qdx300.ci}{0.99 to 1.17}
\defval{r.dly.qdx518}{0.68}   %
\defval{r.dly.qdx518.ci}{0.60 to 0.76}
\defval{r.dly.x300}{0.93}   %
\defval{r.dly.x300.ci}{0.84 to 1.02}
\defval{r.dly.w300}{8}   %
\defval{r.dly.x518}{0.80}   %
\defval{r.dly.x518.ci}{0.72 to 0.90}
\defval{r.dly.x1200}{0.20}   %
\defval{r.dly.x1200.ci}{0.17 to 0.23}
\defval{r.dly.x2000}{0.16}   %
\defval{r.dly.x2000.ci}{0.14 to 0.19}
\defval{r.dly.x12qd}{0.82}   %
\defval{r.dly.x12qd.ci}{0.74 to 0.90}
\defval{r.dly.x5000}{0.15}   %
\defval{r.dly.x5000.ci}{0.12 to 0.17}
\defval{r.dly.qdx1200}{0.35}   %
\defval{r.dly.qdx1200.ci}{0.31 to 0.39}
\defval{r.dly.qdx2000}{0.32}   %
\defval{r.dly.qdx2000.ci}{0.29 to 0.36}
\defval{r.dly.qdx5000}{0.34}   %
\defval{r.dly.qdx5000.ci}{0.30 to 0.38}
\defval{r.dly.xqd0}{2.81}   %
\defval{r.dly.xqd0.ci}{2.38 to 3.37}
\defval{r.dly.xqd0a}{2.8}   %
\defval{r.dly.xqd1200}{1.57}   %
\defval{r.dly.xqd1200.ci}{1.45 to 1.71}
\defval{r.dly.xqd2000}{1.42}   %
\defval{r.dly.xqd2000.ci}{1.33 to 1.52}
\defval{r.dly.xqd5000}{1.21}   %
\defval{r.dly.xqd5000.ci}{1.12 to 1.31}
\defval{r.dly.rear5}{303}\defval{r.dly.rearpass5}{302}\defval{r.dly.npass5}{380}   %
\defval{r.dly.qdrear5}{278}\defval{r.dly.qdrearpass5}{278}\defval{r.dly.qdnpass5}{307}   %
\defval{r.dly.prog5}{75}   %
\defval{r.dly.prog2}{99}\defval{r.dly.qdprog2}{98}   %
\defval{r.ci.rl84}{1.06 to 1.51}   %
\defval{r.ci.rl1200}{0.240 to 0.282}   %
\defval{r.ci.qdp518}{0.289 to 0.351}   %
\defval{r.ci.qdr1200}{0.148 to 0.170}   %
\defval{r.ci.s1}{1.01 to 1.44}   %
\defval{r.ci.rl0}{1.11 to 1.60}   %
\defval{r.ci.qdr0}{0.410 to 0.521}   %
\defval{r.ci.qdrl0}{0.342 to 0.436}   %
\defval{r.pd.o0}{0.07}   %
\defval{r.pd.o100}{0.07}   %
\defval{r.pd.o300}{0.07}   %
\defval{r.pd.o600}{0.15}   %
\defval{r.pd.o1200}{0.47}   %
\defval{r.pd.o2000}{0.46}   %
\defval{r.pd.q0}{0.06}   %
\defval{r.pd.q300}{0.07}   %
\defval{r.pd.q600}{0.65}   %
\defval{r.pd.qr1200}{0.84}   %
\defval{r.as.nox.kmf}{0.908}   %
\defval{r.as.nox.d}{0.106}     %
\defval{r.as.nox.ci}{0.077 to 0.134}
\defval{r.as.nox.off}{12.6}   %
\defval{r.as.sftnox.kmf}{0.902}\defval{r.as.sftnox.pass}{67.1}\defval{r.as.sftnox.d}{0.066}\defval{r.as.sftnox.ci}{0.037 to 0.095}\defval{r.as.sftb.kmf}{1.248}\defval{r.as.sftb.pass}{73.4}
\defval{r.as.qd.rep}{0.37}      %
\defval{r.a1.km}{1.026}

\defval{r.ab.io.nav}{81.31}     %
\defval{r.ab.io.km}{0.914}
\defval{r.ab.io.pass}{60.9}
\defval{r.ab.ih.nav}{82.03}     %
\defval{r.ab.ih.km}{1.199}
\defval{r.ab.ih.pass}{67.6}
\defval{r.ab.ro.nav}{90.83}     %
\defval{r.ab.ro.km}{0.993}
\defval{r.ab.ro.pass}{62.6}
\defval{r.ab.rh.nav}{91.30}     %
\defval{r.fz.base.nav}{79.44}\defval{r.fz.s2.nav}{82.17}   %
\defval{r.fz.s2.d}{2.73}\defval{r.fz.s2.ci}{1.83 to 3.59}\defval{r.fz.e2e.d}{9.13}\defval{r.fz.e2e.ci}{8.10 to 10.27}   %
\defval{r.ab.rh.km}{1.204}      %
\defval{r.ab.rh.pass}{67.8}
\defval{r.ab.rh.nc}{98.80}\defval{r.ab.rh.ttc}{95.65}\defval{r.ab.rh.ep}{86.35}   %
\defval{r.ab.hist.d}{0.08}      %
\defval{r.ab.hist.lo}{0.05}
\defval{r.ab.hist.hi}{0.11}
\defval{r.ab.hist.only}{106}
\defval{r.ab.hist.onlyno}{46}
\defval{r.ab.hist.col}{46}
\defval{r.ab.hist.colno}{82}
\defval{r.ab.rlsft.nc}{98.48}   %
\defval{r.ab.rlsft.ttc}{95.03}
\defval{r.ab.rlsft.ep}{85.93}
\defval{r.ab.rls.nc}{97.87}     %
\defval{r.ab.rls.ttc}{93.45}
\defval{r.ab.rls.ep}{89.12}
\defval{r.ab.rls.nav}{91.30}
\defval{r.ab.rls.km}{1.103}
\defval{r.ab.rln.nc}{98.93}     %
\defval{r.ab.rln.ttc}{96.50}
\defval{r.ab.rln.ep}{85.55}
\defval{r.ab.rln.nav}{91.60}
\defval{r.ab.rln.km}{1.113}
\defval{r.ab.rlr.nc}{99.11}     %
\defval{r.ab.rlr.ttc}{97.16}
\defval{r.ab.rlr.ep}{83.68}
\defval{r.ab.rlr.nav}{91.16}
\defval{r.ab.rlr.km}{1.074}
\defval{r.ab.rlb.nc}{99.21}     %
\defval{r.ab.rlb.ttc}{97.33}
\defval{r.ab.rlb.ep}{82.53}
\defval{r.ab.rlb.nav}{90.74}
\defval{r.ab.rlb.km}{0.978}
\defval{r.ab.ih.nc}{98.05}      %
\defval{r.ab.ih.ttc}{93.12}
\defval{r.ab.ih.ep}{76.91}
\defval{r.ab.rlm.nc}{99.35}     %
\defval{r.ab.rlm.ttc}{97.92}
\defval{r.ab.rlm.ep}{79.44}
\defval{r.ab.rlm.nav}{89.79}
\defval{r.ab.rlm.km}{1.655}
\defval{r.ab.beta.d}{0.020}     %
\defval{r.ab.beta.only}{57}     %
\defval{r.ab.beta.onlyb}{34}    %
\defval{r.ab.v2.pai}{0.378}     %
\defval{r.ab.v1.pai}{0.397}     %
\defval{r.ab.v2.km}{1.018}      %
\defval{r.ab.v2.nav}{90.81}     %
\defval{r.ab.v2.n}{152}         %
\defval{r.ab.v1rl.nav}{91.65}
\defval{r.ab.v1rl.pai}{0.403}   %
\defval{r.ab.v1rl.kmf}{1.301}
\defval{r.ab.v1rl.pass}{68.3}
\defval{r.ab.v2v1.d}{$+0.003$}
\defval{r.ab.v2v1.ci}{$-0.012$ to $+0.018$}
\defval{r.ab.v2rl.nav}{91.60}
\defval{r.ab.v2rl.pai}{0.383}
\defval{r.ab.v2rl.kmf}{1.369}
\defval{r.ab.v2rl.pass}{68.7}

\defval{r.q.cv.base}{86.36}
\defval{r.q.cv.qd}{85.29}
\defval{r.q.cv.s1}{87.60}
\defval{r.q.cv.s2}{87.66}
\defval{r.q.cv.sft}{86.70}
\defval{r.q.bl.base}{79.37}
\defval{r.q.bl.qd}{77.60}
\defval{r.q.bl.s1}{84.68}
\defval{r.q.bl.s2}{81.34}
\defval{r.q.bl.sft}{80.75}
\defval{r.q.sr.base}{45.39}
\defval{r.q.sr.qd}{46.81}
\defval{r.q.sr.s1}{60.27}
\defval{r.q.sr.s2}{60.50}
\defval{r.q.sr.sft}{57.44}
\defval{r.q.dl.base}{60.81}
\defval{r.q.dl.qd}{65.06}
\defval{r.q.dl.s1}{73.55}
\defval{r.q.dl.s2}{84.33}
\defval{r.q.dl.sft}{79.32}
\defval{r.q.ml.base}{75.91}
\defval{r.q.ml.qd}{70.69}
\defval{r.q.ml.s1}{79.79}
\defval{r.q.ml.s2}{79.63}
\defval{r.q.ml.sft}{76.79}
\defval{r.q.mp.base}{50.41}
\defval{r.q.mp.qd}{44.35}
\defval{r.q.mp.s1}{59.03}
\defval{r.q.mp.s2}{57.98}
\defval{r.q.mp.sft}{52.43}
\defval{r.q.ar.base}{93.77}
\defval{r.q.ar.qd}{91.38}
\defval{r.q.ar.s1}{95.39}
\defval{r.q.ar.s2}{95.14}
\defval{r.q.ar.sft}{93.34}
\defval{r.q.hs.base}{88.67}
\defval{r.q.hs.qd}{81.83}
\defval{r.q.hs.s1}{91.00}
\defval{r.q.hs.s2}{91.04}
\defval{r.q.hs.sft}{88.81}
\defval{r.q.wg.base}{72.53}
\defval{r.q.wg.qd}{67.40}
\defval{r.q.wg.s1}{76.40}
\defval{r.q.wg.s2}{77.43}
\defval{r.q.wg.sft}{73.48}
\defval{r.q.bq.base}{88.62}
\defval{r.q.bq.qd}{87.65}
\defval{r.q.bq.s1}{89.72}
\defval{r.q.bq.s2}{89.48}
\defval{r.q.bq.sft}{88.99}
\defval{r.q.tq.base}{74.42}
\defval{r.q.tq.qd}{72.95}
\defval{r.q.tq.s1}{67.07}
\defval{r.q.tq.s2}{66.22}
\defval{r.q.tq.sft}{71.48}
\defval{r.q.tq.negb}{75.8}\defval{r.q.tq.neg1}{61.2}   %
\defval{r.q.n}{eleven}
\defval{r.q.nbase}{nine}    %
\defval{r.q.nqd}{ten}       %
\defval{r.q.maxdrop}{2.21}  %

\defval{r.l2d.n}{38}\defval{r.l2d.sisr}{40.5}\defval{r.l2d.cfps}{21.1}\defval{r.l2d.scs}{65.0}\defval{r.l2d.pcs}{56.5}
\defval{r.l2d.cfpsvla}{0.0}   %
\defval{r.l2d.pcsmax}{6.7}   %
\defval{r.l2d.rows}{301k}\defval{r.l2d.scenes}{100k}   %

\defval{r.tk.n}{12,146}
\defval{r.tk.rl.end1}{30}   %
\defval{r.tk.sft.k1}{90.03}\defval{r.tk.sft.k6}{95.37}
\defval{r.tk.rl.k1}{91.28}\defval{r.tk.rl.k6}{94.80}
\defval{r.tk.sft.top1}{89}\defval{r.tk.rl.top1}{89}           %
\defval{r.tk.sft.one}{84}\defval{r.tk.rl.one}{88}             %
\defval{r.tk.sft.in}{96}\defval{r.tk.rl.in}{94}               %
\defval{r.tk.sft.spread}{1.2}\defval{r.tk.rl.spread}{1.3}     %
\defval{r.tk.rand.spread}{33}                                 %
\defval{r.tk.sft.big}{6}\defval{r.tk.rl.big}{3}               %
\defval{r.tk.sft.bigshare}{78}\defval{r.tk.rl.bigshare}{59}   %
\defval{r.tk.sft.same}{90}\defval{r.tk.rl.same}{93}           %
\defval{r.tk.rl.dist}{1.8}                                    %
\defval{r.tk.straight}{56}   %
\defval{r.tk.d1}{1.25}\defval{r.tk.d6}{0.57}

\usepackage[breaklinks,colorlinks]{hyperref}
\usepackage[capitalize,nameinlink]{cleveref}
\crefname{section}{Sec.}{Secs.}\Crefname{section}{Section}{Sections}   %
\crefname{table}{Tab.}{Tabs.}\Crefname{table}{Table}{Tables}

\renewcommand{\dblfloatpagefraction}{0.7}\renewcommand{\dbltopfraction}{0.7}
\sgcaptionfix{figure}{2}{\looseness=-1}                         %
\sgcaptionfix{table}{14}{\sgminlast{8em}}                       %
\sgcaptionfix{table}{18}{\looseness=-1}                         %
\sgparapad{WOD-E2E and PAI-AV.}{5pt}                            %
\sgparapad{AlpaSim.}{-1.5pt}                                    %
\sgitemfix{2}{\sgminlast{8em}}                                  %
\sgsecfix{Latency}{\sgminlast{4em}}                             %
\newcommand{\maketitlesupplementary}{\setcounter{page}{\sgsupppage}%
  \twocolumn[{\centering\Large\sffamily\bfseries\color{sgink}Appendix\par\vskip 1em}]}

\newcommand{\sgtitle}{SuperGlance: Driving at a Glance with a Vision-Language Model}
\newcommand{\sgstatementhead}{LLM Usage Statement}
\newcommand{\sgstatement}{We used Claude Code (Anthropic) extensively in this work: to search related work, propose and discuss ideas, write training, evaluation and figure code, launch and monitor experiments, analyse results, and draft and edit the text of this paper; Gemini (Google) also suggested wording edits.
The authors directed the research, checked the results and the text, and take full responsibility for them.}

\title{\sgtitle}
\newcommand{\sgcorr}{\scalebox{0.85}{\faEnvelope[regular]}}
\author[1,2,3,\textasteriskcentered]{Jieyuan Pei}
\author[4,\textasteriskcentered]{Xin Yan}
\author[5]{Mingwei Xu}
\author[3]{Haiyi Zhang}
\author[2]{Siyu Zou}
\author[2,3]{Zhangyi Hu}
\author[6]{Zhuoxuan Li}
\author[7]{Yuguang Yang}
\author[2,\textdagger]{Rongjunchen Zhang}
\author[6]{Cheng Chi}
\author[3,\sgcorr]{Hui Xiong}
\author[1,\sgcorr]{Yan Wang}
\sglogos{arxiv_style/logos/hithink}{arxiv_style/logos/tsinghua_air,arxiv_style/logos/hkust_gz}
\sglogotrim{arxiv_style/logos/tsinghua_air}{1 2 2 2}   %
\sglogoscale{arxiv_style/logos/hithink}{1.15}          %
\affiliation[1][arxiv_style/logos/mark_tsinghua_air]{Institute for AI Industry Research (AIR), Tsinghua University}
\affiliation[2][arxiv_style/logos/mark_hithink]{HiThink Research}
\affiliation[3][arxiv_style/logos/mark_hkust_gz]{The Hong Kong University of Science and Technology (Guangzhou)}
\affiliation[4][arxiv_style/logos/mark_astar]{Agency for Science, Technology and Research (A*STAR), Singapore}
\affiliation[5][arxiv_style/logos/mark_uw]{University of Washington}
\affiliation[6][arxiv_style/logos/mark_ruc]{Renmin University of China}
\affiliation[7][arxiv_style/logos/mark_beihang]{Beihang University}
\sgnotes{\textsuperscript{\textasteriskcentered}Equal contribution\qquad\textsuperscript{\textdagger}Project leader\qquad\textsuperscript{\sgcorr}Corresponding authors}
\metadata[Models]{\hflink{\sgModelURL}\qquad\hflink{\sgStageOneURL}\qquad\hflink{\sgStageTwoURL}}
\metadata*[Data]{\hflink{\sgDataURL}}
\metadata[Code]{\iconlink[sggh]{\faGithub}{\sgCodeURL}{GitHub}}
\metadata*[Project Page]{\iconlink{\faGlobe}{\sgProjectURL}{Project Website}}
\hypersetup{pdftitle={\sgtitle},pdfauthor={\sgauthornames}}

\begin{document}
\begin{abstract}
Autonomous driving is turning to vision-language-action models (VLAs).
Most answer and plan by decoding, one step at a time.
Each step is another forward pass, so a plan can take over a second of GPU time.
Embodied AI often pairs such a slow, reasoning \mbox{System}~2 with a fast \mbox{System}~1 that acts.
Human drivers plan at a glance, with \mbox{System}~1.
Faster designs usually put \mbox{System}~1 outside the vision-language model (VLM), in a planner without its abilities.
Yet before a VLM writes its first token, it has already read the scene into one hidden state.
We show that, once trained, this hidden state is enough to drive and to answer questions about what the model sees.
\method reads it to turn a decoder-only VLM into an encoder-style driving policy.
Its linear head scores a fixed set of \val{r.k.rl} candidate trajectories.
With the set fixed, rewards are computed offline, so supervised targets can be reward-weighted and reinforcement learning maximizes the exact expected reward.
Before and during supervised policy training, multiple-choice training keeps the VLM's visual abilities in one pass, so the policy still answers questions about driving scenes.
Training questions are built against shortcuts.
Each must be answered right in two option orders.
Against drift, a KL term keeps each stage near its starting model.
One pass offers VLAs a more plausible path to \mbox{System}~1.
\method answers visual multiple-choice questions as well as its base VLM does by decoding.
It plans in \val{r.l.ours}~ms of GPU time, \val{r.l.xqdw} times faster than Qwen-Drive-1.0, and at least as well on NAVSIM, PAI-AV and WOD-E2E.
In AlpaSim closed loop, km per at-fault incident holds for both planners up to a 300-ms delay and drops beyond.
With GPU time charged as delay, \method's lead over Qwen-Drive-1.0 grows from \val{r.dly.xqd0a} to \val{r.as.x} times.
Delayed by 1.2~s, our model falls below Qwen-Drive-1.0 at its own \val{r.l.qd}~ms.
\end{abstract}
\vspace{-12pt}   %

\begin{strip}
\centering
\includegraphics[width=\linewidth,trim=0 3.3bp 0 4bp,clip]{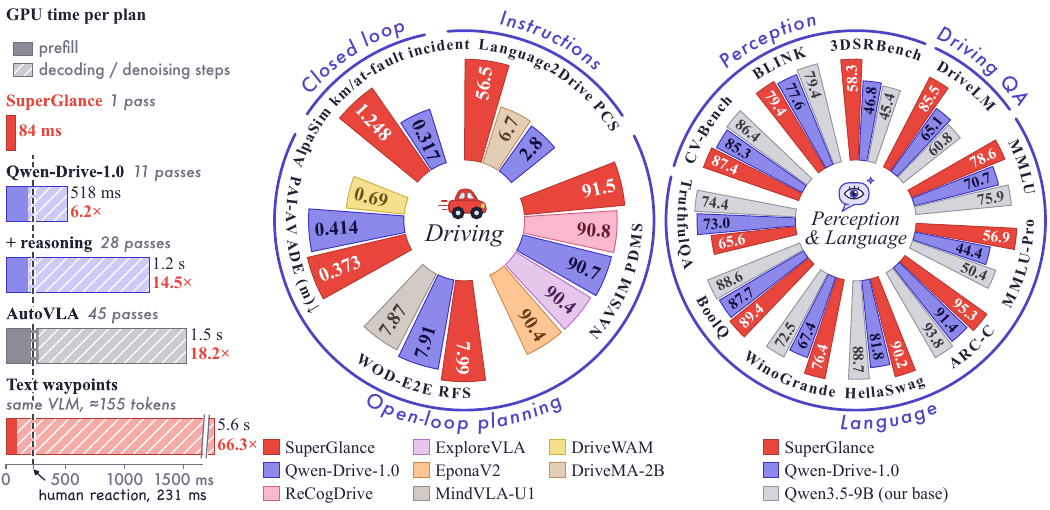}
\captionsetup{hypcap=false}
\captionof{figure}{\textbf{\method and benchmark results.}
A driving \mbox{System}~1 needs world knowledge and fast reactions.   %
\method plans in one pass, \val{r.l.xqd} times faster than Qwen-Drive-1.0, and is at or above it on ten of eleven QA and all driving benchmarks.
Left: GPU time per plan, charged as delay in AlpaSim.
Middle: ours, Qwen-Drive-1.0 at its best per benchmark~\cite{qwendrive} and others; Language2Drive as in \cref{tab:l2d}.
Right: ours, Qwen-Drive-1.0 (4B) and our base.
Ring bars start above zero; longer ADE bars mean lower error.   %
}
\captionsetup{hypcap=true}
\label{fig:teaser}
\vspace{6pt}   %
\end{strip}

\maketitle

\section{Introduction}
\begin{figure*}[t]
\centering
\includegraphics[width=\linewidth,trim=0 4bp 0 0bp,clip]{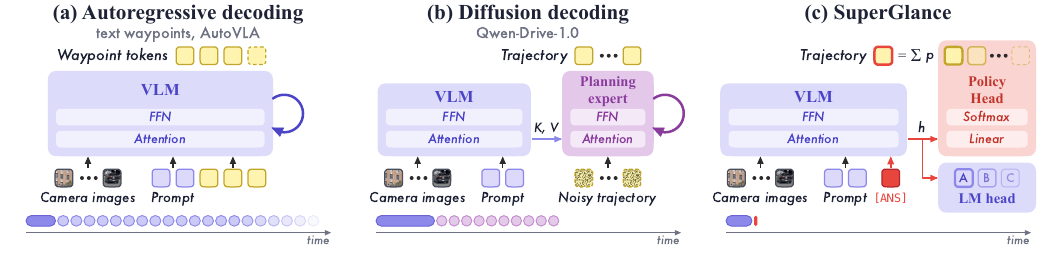}
\caption{\textbf{Three ways a vision-language model produces a plan.}
Only (c) has no loop.
(a)~Autoregressive decoding writes a trajectory one token per forward pass (text waypoints, AutoVLA~\cite{autovla}).
(b)~Diffusion decoding runs the VLM once, then a planning expert refines a noisy trajectory over several steps (Qwen-Drive-1.0~\cite{qwendrive}).
(c)~\method reads the last hidden state $h$ once, at \texttt{[ANS]}, the prompt's last position.
The Policy Head gives a probability (outline width and opacity) to each of \val{r.k.rl} fixed candidates, and for questions, the LM head gives one to each option letter.
Each bottom strip shows one plan's measured GPU time to scale; in (a), our VLM writing text runs off its axis.}
\label{fig:arch}
\vspace{-6pt}   %
\end{figure*}
\label{sec:intro}

Autonomous driving is turning to vision-language-action models (VLAs), vision-language models (VLMs) trained end to end as driving policies~\cite{drivelm,drivevlm,autovla,recogdrive,qwendrive}.
They bring broad visual knowledge.
Such a policy reads camera frames and the ego state and outputs a trajectory for the next few seconds.
A VLM answers by decoding, and most driving VLAs produce this trajectory step by step as well (\cref{fig:arch}a,b): they write waypoints as text or action tokens, one per forward pass~\cite{adathinkdrive,curiousvla,driveteach,autovla}, or denoise it over several steps with a planning expert that reads the VLM's features~\cite{recogdrive,qwendrive}.

Every step is another forward pass, and the passes add up.
On one H200 GPU at batch size one, Qwen-Drive-1.0~\cite{qwendrive} spends \val{r.l.qd}~ms of GPU time on a plan over \val{r.l.qdpass} passes, and \val{r.l.qdr}~ms when also writing short reasoning (\cref{fig:teaser}, left).
The car keeps moving while the policy thinks, so the plan arrives late (\cref{sec:main}).
A human driver plans at a glance, with the fast, automatic thinking known as System~1~\cite{kahneman}.
Faster designs mostly move planning out of the VLM into a separate planner or smaller student (\cref{sec:related}), leaving the VLM's abilities behind.
So far, a fast driving VLA that answers questions either plans outside the VLM or regresses one trajectory from query tokens~\cite{simlingo}.

Yet before a VLM writes its first token, it has already read the scene into one hidden state.
We show that, once trained, this hidden state is enough to drive and to answer questions about what the model sees.
VLMs became decoder-only to write free text; a plan is not free text.
BERT~\cite{bert} and GPT~\cite{gpt1} classify from one hidden state with a small head, as did early vision-language transformers.
We bring this classification approach to driving VLAs, with candidate trajectories as the classes, so a plan takes one forward pass.

We present \method, which turns a decoder-only VLM into an encoder-style driving policy that plans in one forward pass (\cref{fig:teaser}, left; \cref{fig:arch}c).
A linear Policy Head on the last hidden state at the answer position, \texttt{[ANS]}, scores a fixed set of \val{r.k.rl} candidate trajectories, most of them cluster centers of human trajectories (\cref{sec:vocab}).
As the set is fixed, every candidate's reward in every training scene is computed once, offline.
Training has three stages (\cref{fig:data}).
Stages~1 and~2 train on multiple-choice questions to keep the VLM's visual abilities in one forward pass and carry them to questions about driving scenes.
Stage~3 trains the policy, first toward a supervised target that can weight candidates by reward.
The supervised policy can still leave much probability on unsafe candidates (\cref{fig:choose}b), so reinforcement learning (RL) raises the expected reward, computed exactly over all candidates without sampling, and lifts NAVSIM's no-collision and time-to-collision scores.
The language-model head is kept and question data stays in training, so the final model still answers questions at almost no cost to planning.
With no decoding, denoising or separate planner, the VLM itself plans at a glance, as a human driver does.

We test our final RL checkpoint on NAVSIM, WOD-E2E, PAI-AV and Language2Drive, and in AlpaSim closed loop~\cite{alpasim}.
With each plan's GPU time charged as delay, it drives \val{r.as.x} times as far per at-fault incident as Qwen-Drive-1.0 (\cref{fig:teaser}).
Without delay, it still leads, as does every one-pass variant in our ablations (\cref{sec:main}).
\begin{itemize}
\item \method plans from one hidden state of a VLM by scoring fixed candidates, so targets can weight them by reward and RL maximizes their exact expected reward.
\item Training in two option orders keeps the VLM's visual abilities in one pass and carries them to questions about driving scenes, at almost no cost to planning.
\item One checkpoint plans in \val{r.l.ours}~ms, \val{r.l.xqd} times faster than Qwen-Drive-1.0, and at least as well on NAVSIM, WOD-E2E, PAI-AV, AlpaSim and Language2Drive; its visual multiple-choice answers rival its base VLM's decoded ones.
\end{itemize}

\section{Related Work}
\label{sec:related}
\begin{table}[t]
\centering
\scriptsize
\setlength{\tabcolsep}{2.75pt}   %
\begin{tabular}{llcccc}
\toprule
& Planning & Sequential & Planner & Keeps & RL \\
Method & paradigm & steps & in VLM & language & objective \\
\midrule
\grouprow\multicolumn{6}{l}{\textit{Decoding or denoising VLAs}} \\
AutoVLA~\cite{autovla} & Autoregressive & 10 + text & \cmark & \cmark & GRPO \\
Qwen-Drive-1.0~\cite{qwendrive} & Flow matching & 11 + text & \xmark & \cmark & GRPO \\
\grouprow\multicolumn{6}{l}{\textit{One-pass or fast VLM planners}} \\
SimLingo~\cite{simlingo} & Regression & 1 + text & \cmark & \cmark & \xmark \\
AnchorVLA~\cite{anchorvla} & Anchor + flow & 3 + text & \xmark & \cmark & \xmark \\
\grouprow\multicolumn{6}{l}{\textit{Vocabulary scorers without a VLM}} \\
ZTRS~\cite{ztrs} & Scoring & 1 & \xmark & \xmark & Exact \\   %
\midrule
\ours \method & Scoring & 1 & \cmark & \cmark & Exact \\
\bottomrule
\end{tabular}
\caption{\textbf{Related driving policies.} Text adds a step per token.}
\label{tab:related}
\vspace{-6pt}   %
\end{table}

\subsection{Vision-language models as driving policies}
Recent driving VLAs differ mainly in how they emit a trajectory (\cref{tab:related}).
AdaThinkDrive~\cite{adathinkdrive}, Curious-VLA~\cite{curiousvla} and DriveTeach-VLA~\cite{driveteach} write waypoints as text, AutoVLA~\cite{autovla} decodes action tokens, DriveFine~\cite{drivefine} decodes bin tokens by masked diffusion, and WAM-Flow~\cite{wamflow} refines numeric tokens by discrete flow matching.
ReCogDrive~\cite{recogdrive} and Qwen-Drive-1.0~\cite{qwendrive} attach a denoising planner or action expert, and world-model policies~\cite{epona,eponav2,drivewam,simwam} add future-frame prediction.
Most plan over sequential decoding or denoising steps; \method reads its plan from one hidden state in a single forward pass (\cref{fig:arch}).

Fast designs usually move planning out of the VLM: dual systems pair it with a conventional or smaller planner~\cite{drivevlm,leapad,adadrive} or lightweight action expert~\cite{drivevlaw0}, and distillation trains a student that skips reasoning~\cite{critiquedrivevlm}.
CarLLaVA~\cite{carllava}, Reasoning-VLA~\cite{reasoningvla} and, for robot actions, OpenVLA-OFT~\cite{openvlaoft} regress from query tokens in one pass; SimLingo~\cite{simlingo} does so in a language-capable VLM.
AnchorVLA~\cite{anchorvla} refines a decoded trajectory-anchor token with a flow model, ORION~\cite{orion} a planning token with a generative planner, and FiS-VLA~\cite{fisvla} uses a robot VLA's last blocks as a fast policy that still denoises.
Unlike dual systems and students, \method plans with the full VLM.
Unlike regression and anchor-token policies, it plans by scoring candidates from one hidden state, so its target weights each by reward (\cref{sec:closed}) and RL uses all of their rewards.

Many driving VLAs add GRPO-style RL~\cite{deepseekmath} over sampled trajectories~\cite{autovla,recogdrive,qwendrive}.
Perception-based planners score a fixed trajectory vocabulary~\cite{vadv2,hydramdp,drivesuprim,sparsedrivev2}, per-scene candidates~\cite{drivor,world4scorer} or both~\cite{gtrs}.
ZTRS~\cite{ztrs} takes an exact policy gradient over such a vocabulary, and DriveDPO~\cite{drivedpo} distills imitation and safety scores into a policy aligned by preference optimization.
\method brings the vocabulary and exact, sampling-free RL into a VLM.

\subsection{Decision making in vision-language models}
Early VLMs such as ViLBERT~\cite{vilbert}, LXMERT~\cite{lxmert} and UNITER~\cite{uniter} followed BERT~\cite{bert}, casting visual question answering (VQA) as classification over frequent answers with a small head per task on a pre-trained encoder.
Flamingo~\cite{flamingo}, BLIP-2~\cite{blip2} and LLaVA~\cite{llava} instead write their answers token by token, as do Qwen3-VL~\cite{qwen3vl} and Qwen3.5~\cite{qwen35}.
Scoring a fixed answer set takes one pass for any answer length; so does reading option-letter logits for multiple choice~\cite{mcp}, though with letter and position biases~\cite{pride}.
\method brings the classification head back into a generative VLM: its Policy Head scores a fixed set of trajectories as a VQA head scores answers.

\section{Method}
\label{sec:method}

\begin{figure*}[t]
\centering
\includegraphics[width=\linewidth,trim=0 6.9bp 0 0bp,clip]{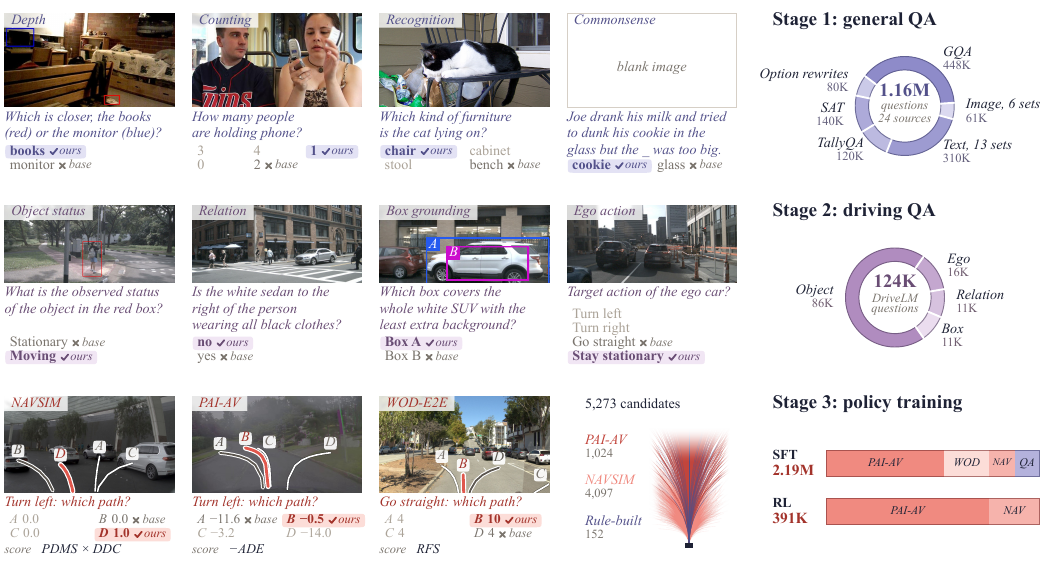}
\caption{\textbf{Training data of each stage.}
Every stage trains an answer read in one forward pass: a distribution over option letters, and in stage~3 also over candidates.
Left: held-out samples where \nineb is wrong and the model after that stage is right.
A tick marks our pick, a cross \nineb's, and questions are lightly abridged.
In the stage-3 row, \nineb chooses among four paths: our supervised model's pick among its \val{r.k.total} candidates and three low-reward candidates, each with its score.
Right: each stage's sources with shares (rings, bars) and counts, and the \val{r.k.rl} candidates seen from above, colored by origin (\cref{sec:vocab}).}
\label{fig:data}
\vspace{-6pt}   %
\end{figure*}

\method plans and answers questions from one hidden state. It reads that state through two heads, each in one forward pass: the language-model head answers multiple-choice questions and a linear Policy Head scores a fixed set of candidate trajectories (\cref{fig:arch}).
Two training stages on multiple-choice questions keep the VLM's visual abilities in one forward pass, and a third trains the policy.

\subsection{Architecture}
\label{sec:select}
\paragraph{Inputs.}
Our VLM, \nineb~\cite{qwen35}, receives the current front-camera frame at 1024$\times$576 and, if available, the \val{r.in.nhist} frames taken \val{r.in.past}, 1.0 and \val{r.in.gap}~s earlier at \val{r.in.histw}$\times$\val{r.in.histh}: 576 visual tokens for the current frame and 170 per past frame.
The prompt gives the ego speed, acceleration, past \val{r.in.past}~s of positions, navigation command and any instruction, and ends at the answer position, \texttt{[ANS]}.
No detector or map output enters the prompt; the scene comes from the frames alone.

\paragraph{Candidates.}
\label{sec:vocab}
Each candidate is a \val{r.k.horizon}-s trajectory of \val{r.k.poses} poses (position and heading) at \val{r.k.hz}~Hz in the ego frame.
Of the $K$\,=\,\val{r.k.rl} candidates, \val{r.k.km} are k-means centers of human trajectories in the \val{r.k.fit} NAVSIM training scenes, and one is stationary.
NAVSIM rarely reaches \val{r.k.fastv}~m/s, so \val{r.k.fast} more are k-means centers of PAI-AV driving at that speed or faster.
Clustering misses stops and curves that are rare in human driving, so \val{r.ab.v2.n} more are built by rule for RL (\suppword{}).
The best clustered candidate per scene reaches a mean PDM score (PDMS) of \val{r.k.bestho} on \val{r.k.nho} held-out NAVSIM training scenes and a mean rater feedback score (RFS) of \val{r.k.bestwod} on the rated WOD-E2E validation frames.

\paragraph{Policy Head.}
The Policy Head is a linear layer with weights $W\in\mathbb{R}^{K\times d}$ and bias $b\in\mathbb{R}^{K}$, one row per candidate, on the last hidden state $h\in\mathbb{R}^{d}$ at \texttt{[ANS]}, $d=4{,}096$.
For scene inputs $x$, it gives the policy and the plan
\begin{equation}
\begin{gathered}
p_\theta(k\mid x)=\operatorname{softmax}(W h+b)_k,\\
\bar a(x)=\textstyle\sum_{k\in T(x)} w_k\,a_k,\quad w_k\propto p_\theta(k\mid x)^{1/\tau},
\end{gathered}
\label{eq:policy}
\end{equation}
where $a_k$ is candidate $k$'s trajectory, $T(x)$ holds the \val{r.pai.topk} most probable candidates and the temperature is $\tau=\val{r.pai.tau}$.
This top-\val{r.pai.topk} mean adds \val{r.t.meansft} PDMS on navtest over the most probable candidate, and \val{r.t.meanrl} after RL.
In AlpaSim, a low-level controller executes every method's plan as in NAVSIM scoring: NAVSIM's PDM simulator~\cite{navsim,pdm}, an LQR tracker on a kinematic bicycle model.
It runs the plan from the ego's current speed and acceleration and the steering angle reached while running the previous plan, and passes 40 poses, 0.1~s apart, to AlpaSim's controller.
The language-model head is unchanged; for multiple choice, its answer distribution $p$ is the softmax over the option-letter logits at \texttt{[ANS]}.

\subsection{Data}
\label{sec:data}
\paragraph{QA data.}
Stage~1 uses \val{r.d.s1}M questions from text and visual~\cite{gqa,sat,tallyqa,ai2d,chartqa,textvqa} datasets and stage~2 uses \val{r.d.s2} driving questions from DriveLM-nuScenes~\cite{drivelm}, all recast as multiple choice with one gold option letter (\cref{fig:data}; \suppword{}).

\paragraph{Rewards.}
\label{sec:reward}
Because the candidates are fixed, every candidate's reward $r(x,k)$ in every training scene $x$ is computed once before training.
On NAVSIM~\cite{navsim}, it is the PDMS times the driving-direction compliance (DDC), both in $[0,1]$.
A candidate that collides, leaves the drivable area or violates the time-to-collision margin counts as unsafe, so it loses a further~\val{r.t.pen} and ranks below every safe candidate.
On PAI-AV~\cite{paiav}, it is $-0.2$ times the average displacement error (ADE, in meters) to the logged future, and on the rated WOD-E2E frames~\cite{wode2e} it is the RFS divided by 10.

\paragraph{Rollout rows.}
In closed loop, a policy drifts into states that logged driving never shows.
To learn to recover, stage~3 also trains on \val{r.d.dag} frames, \val{r.d.dagrow}\% of its rows (examples), rendered along AlpaSim rollouts of earlier models on \val{r.d.dagsc} scenes outside the evaluation scenarios.
Their target trajectory rejoins the human path in about 3~s, never ahead of the human, and slows behind a lead vehicle.

\subsection{Training}
\label{sec:stages}
Each stage trains rank-\val{r.t.rank} LoRA adapters~\cite{lora} on the language model, merged before the next, and a KL term keeps each stage near its starting model, so that later stages keep what earlier ones learned.

\paragraph{Stage 1: general QA.}
Read in one pass, a VLM's option letter, like an LLM's~\cite{pride}, is biased toward some letters and positions (\cref{fig:shortcut}c), so stage~1 trains it to follow the content.
Each question with gold letter $y$ has a twin with shifted options and gold $y'$.
One letter is drawn from each order's distribution, $p$ and $p'$, and the pair hits only if both are gold, so an order-dependent answer earns no credit (\cref{fig:shortcut}).
The reward is one minus the Brier score~\cite{brier1950} of the pair distribution $p\otimes p'$ against the gold pair, and \val{train.draws} drawn pairs estimate its gradient without bias (\suppword{}).

\paragraph{Stage 2: driving QA.}
Stage~2 repeats stage~1 on DriveLM, carrying the VLM's abilities to questions about driving scenes: objects, their relations and the ego vehicle.
Stage-1 questions stay in its batches to keep the general abilities.

\paragraph{Stage 3: policy training.}
Stage~3 trains the policy, first toward a supervised target $q(k\mid x)\propto\exp\big(s(x,k)/t\big)$ that spreads probability over the candidates by a score $s(x,k)$:
\begin{equation}
\mathcal{L}_{\mathrm{SFT}}=-\textstyle\sum_{k} q(k\mid x)\log p_\theta(k\mid x).
\label{eq:target}
\end{equation}
On NAVSIM, $s$ is PDMS$\times$DDC minus \val{r.t.adew} times the ADE in meters, with $t=\val{r.t.tnav}$, so the target favors safe candidates and, among them, those near the human trajectory.
On PAI-AV and WOD-E2E, $s$ is the negative ADE, with $t=\val{r.t.tpai}$~m and \val{r.t.twod}~m; a last phase uses WOD-E2E frames around the rated ones, with an RFS-based score (\suppword{}).
Stage-1 and stage-2 questions stay in the batches so that the policy keeps answering questions (\cref{sec:vqa}).

\begin{figure}[t]
\centering
\includegraphics[width=\linewidth,trim=0 4.8bp 0 0bp,clip]{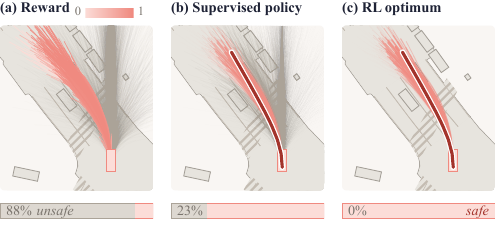}
\caption{\textbf{From rewards to a policy on one navtest scene.}
(a)~Each candidate is colored by its reward; unsafe ones are gray (\cref{sec:reward}).
(b)~In the supervised policy, width and opacity grow with probability, and unsafe candidates hold nearly a quarter of it.
(c)~The RL optimum, computed from (a) and (b) by \cref{eq:opt} without the RL model, gives them none.
Bars give the unsafe share of candidates (a) and of probability (b, c).}
\label{fig:choose}
\vspace{-6pt}   %
\end{figure}
Before RL, the Policy Head grows to all \val{r.k.rl} candidates: its old rows stay, and new rows for the \val{r.ab.v2.n} rule-built ones are fit on the supervised model's features (\suppword{}).
The resulting policy, $p_{\mathrm{SFT}}$, still leaves probability on unsafe candidates (\cref{fig:choose}b), so RL maximizes the expected reward with a KL term that keeps the policy close to $p_{\mathrm{SFT}}$:
\begin{equation}
\mathbb{E}_{x}\big[\textstyle\sum_{k} p_\theta(k\mid x)\,r(x,k)-\beta\,\mathrm{KL}(p_\theta\,\|\,p_{\mathrm{SFT}})\big].
\label{eq:rl}
\end{equation}
The sum runs over all $K$ candidates, so the expectation and its gradient are exact; no trajectory is sampled or simulated (\suppword{}).
As in KL-regularized RL~\cite{korbak2022kl,dpo}, the policy that maximizes \cref{eq:rl} is, in each scene,
\begin{equation}
p^\star(k\mid x)\propto p_{\mathrm{SFT}}(k\mid x)\,\exp\!\big(r(x,k)/\beta\big).
\label{eq:opt}
\end{equation}
It reweights $p_{\mathrm{SFT}}$ by reward, and $\beta$ sets how far it moves from $p_{\mathrm{SFT}}$ (\cref{fig:choose}c); we use $\beta=\val{r.t.beta}$.

\section{Experiments}
\label{sec:exp}

\subsection{Setup}
\label{sec:setup}
\begin{figure*}[t]
\centering
\includegraphics[width=\linewidth]{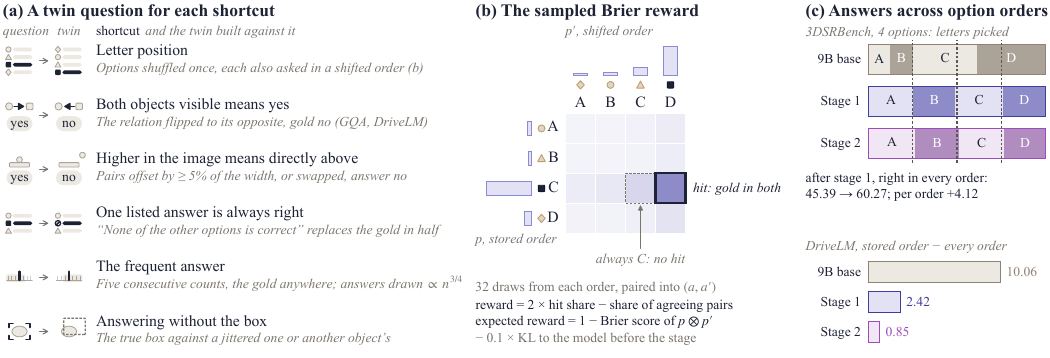}
\caption{\textbf{Questions built against shortcuts in stages~1 and~2.}
A one-pass letter can be right by position alone.
(a)~A shortcut answers a question right (left) and its twin wrong (right); gold is dark.
(b)~Rows use the stored order (gold at C) and columns the shifted one (gold at D); bars show illustrative letter probabilities, shades pair probabilities.
(c)~Letter shares on 3DSRBench's 4-option questions, with gold spread evenly (dashed: 25\% each); \nineb picks C or D for 75\%, by position, and after stage~2 each letter takes 23--27\%.}
\label{fig:shortcut}
\end{figure*}
\begin{table}[t]
\centering
\scriptsize
\setlength{\tabcolsep}{2.5pt}
\setlength{\aboverulesep}{0pt}\setlength{\belowrulesep}{0pt}\setlength{\extrarowheight}{1.6pt}   %
\begin{tabular}{lc>{\color{black!55}}cc*{3}{w{c}{18.97pt}}}
\toprule
& \multicolumn{2}{c}{\nineb} & & \multicolumn{3}{c}{\method after stage} \\
\cmidrule(lr){2-3}\cmidrule(l){5-7}
Benchmark & letter & decoded & Qwen-Drive-1.0 & 1 & 2 & 3 \\
\midrule
\grouprow\textit{Vision} & & & & & & \\
CV-Bench~\cite{cambrian} & \heatv{35}{\val{r.q.cv.base}} & \val{r.q.cv.dec} & \heatv{0}{\val{r.q.cv.qd}} & \heatv{86}{\val{r.q.cv.s1}} & \heatv{100}{\val{r.q.cv.s2}} & \heatv{70}{\val{r.q.cv.rl}} \\
BLINK~\cite{blink} & \heatv{29}{\val{r.q.bl.base}} & \val{r.q.bl.dec} & \heatv{0}{\val{r.q.bl.qd}} & \heatv{100}{\val{r.q.bl.s1}} & \heatv{60}{\val{r.q.bl.s2}} & \heatv{29}{\val{r.q.bl.rl}} \\
3DSRBench~\cite{threedsr} & \heatv{0}{\val{r.q.sr.base}} & \val{r.q.sr.dec} & \heatv{17}{\val{r.q.sr.qd}} & \heatv{87}{\val{r.q.sr.s1}} & \heatv{100}{\val{r.q.sr.s2}} & \heatv{68}{\val{r.q.sr.rl}} \\
\grouprow\textit{Language} & & & & & & \\
MMLU~\cite{mmlu} & \heatv{41}{\val{r.q.ml.base}} & \val{r.q.ml.dec} & \heatv{0}{\val{r.q.ml.qd}} & \heatv{100}{\val{r.q.ml.s1}} & \heatv{87}{\val{r.q.ml.s2}} & \heatv{69}{\val{r.q.ml.rl}} \\
MMLU-Pro~\cite{mmlupro} & \heatv{33}{\val{r.q.mp.base}} & \val{r.q.mp.dec} & \heatv{0}{\val{r.q.mp.qd}} & \heatv{100}{\val{r.q.mp.s1}} & \heatv{84}{\val{r.q.mp.s2}} & \heatv{68}{\val{r.q.mp.rl}} \\
ARC-C~\cite{arc} & \heatv{42}{\val{r.q.ar.base}} & \val{r.q.ar.dec} & \heatv{0}{\val{r.q.ar.qd}} & \heatv{100}{\val{r.q.ar.s1}} & \heatv{72}{\val{r.q.ar.s2}} & \heatv{87}{\val{r.q.ar.rl}} \\
HellaSwag~\cite{hellaswag} & \heatv{50}{\val{r.q.hs.base}} & \val{r.q.hs.dec} & \heatv{0}{\val{r.q.hs.qd}} & \heatv{87}{\val{r.q.hs.s1}} & \heatv{100}{\val{r.q.hs.s2}} & \heatv{70}{\val{r.q.hs.rl}} \\
WinoGrande~\cite{winogrande} & \heatv{42}{\val{r.q.wg.base}} & \val{r.q.wg.dec} & \heatv{0}{\val{r.q.wg.qd}} & \heatv{78}{\val{r.q.wg.s1}} & \heatv{100}{\val{r.q.wg.s2}} & \heatv{78}{\val{r.q.wg.rl}} \\
BoolQ~\cite{boolq} & \heatv{36}{\val{r.q.bq.base}} & \val{r.q.bq.dec} & \heatv{0}{\val{r.q.bq.qd}} & \heatv{100}{\val{r.q.bq.s1}} & \heatv{82}{\val{r.q.bq.s2}} & \heatv{68}{\val{r.q.bq.rl}} \\
TruthfulQA~\cite{truthfulqa} & \heatv{100}{\val{r.q.tq.base}} & \val{r.q.tq.dec} & \heatv{79}{\val{r.q.tq.qd}} & \heatv{33}{\val{r.q.tq.s1}} & \heatv{16}{\val{r.q.tq.s2}} & \heatv{0}{\val{r.q.tq.rl}} \\
\grouprow\textit{Driving} & & & & & & \\
DriveLM~\cite{drivelm} & \heatv{0}{\val{r.q.dl.base}} & \val{r.q.dl.dec} & \heatv{21}{\val{r.q.dl.qd}} & \heatv{51}{\val{r.q.dl.s1}} & \heatv{85}{\val{r.q.dl.s2}} & \heatv{100}{\val{r.q.dl.rl}} \\
\bottomrule
\end{tabular}
\caption{\textbf{Question answering.} Accuracy (\%) after each training stage of \method; all models are scored by us with the same prompts and option-letter readout.
Shading deepens from the worst to the best score, by row here and by column below.
The decoded column reads \nineb's written replies (thinking off).
Stage~3 is our final RL model.
For BLINK we use counting, object localization, relative depth and spatial relation.
DriveLM (our scene-split test questions) and 3DSRBench count an answer only if it holds in every cyclic option order.
Only DriveLM is in our training data.}
\label{tab:vqa}
\vspace{-6pt}   %
\end{table}
\paragraph{Benchmarks.}
We evaluate driving broadly, on five benchmarks: open-loop planning on NAVSIM, WOD-E2E and PAI-AV, closed-loop driving in AlpaSim and instruction following on Language2Drive (\cref{tab:l2d}).
\textbf{NAVSIM}~\cite{navsim} v1.1 navtest has \val{r.nav.n} scenes scored by PDMS, which multiplies no-collision (NC) and drivable-area compliance (DAC) by a weighted sum of ego progress (EP), time-to-collision (TTC) and comfort (Comf.).
\textbf{WOD-E2E}~\cite{wode2e} covers long-tail scenes; we report RFS and 3- and 5-s ADE on its test server.
\textbf{PAI-AV}~\cite{paiav} open loop is scored by 3- and 5-s ADE on 700 frames from 140 held-out clips.
\textbf{AlpaSim}~\cite{alpasim} runs closed loop on the 916 scenarios of PAI-AV-NuRec~\cite{painurec,threedgut} v26.01, and the policy plans anew every 100~ms.
We report collision and off-road rates, progress, share of scenarios passed and distance driven per incident (collision or off-road event).
At-fault rates leave out hits on the ego's rear.
A scenario passes if it has no at-fault collision, off-road event or lateral-corridor exit.
Its score is progress along the human path divided by 0.8, capped at 1, or zero if it fails.

\paragraph{Baselines.}
We run Qwen-Drive-1.0's supervised and RL planners on PAI-AV and in AlpaSim, executed like ours.

\subsection{Efficiency and question answering}
\label{sec:eff}
\label{sec:vqa}
Generation dominates a generated plan's GPU time on one H200 (\cref{fig:teaser}, left, and \cref{fig:latency}, top; protocol in the \suppword{}): a flow step of Qwen-Drive-1.0 or a decoded token of our VLM takes \val{r.l.step}~ms, so Qwen-Drive-1.0's ten flow steps alone take four times our whole \val{r.l.ours}~ms pass, and writing the plan as text takes our VLM \val{r.l.xtext} times as long.
Reading alone takes Qwen-Drive-1.0 \val{r.l.qdread}~ms, over twice our whole pass, as it reads nearly four times as many image tokens.

Multiple-choice training keeps the VLM's visual abilities in one pass (\cref{tab:vqa}).
On 3DSRBench, the base model's letter trails its own decoded replies by \val{r.q.dec.srgap} points, mostly through position bias rather than missing knowledge.
Stage~1 closes the gap by answering consistently across option orders (\cref{fig:shortcut}c).
After all three stages, our letter is level with or above \nineb's decoded replies (thinking off) on every vision benchmark and \val{r.q.dec.dl} points above them on DriveLM, which we train on.
It is at or above \nineb's letter on every benchmark but TruthfulQA, and policy training in stage~3 moves no score by more than \val{r.q.maxdrop} points.
On language, \nineb's written replies lead our letter by up to \val{r.q.dec.mp} points when reasoning first; when answering first, they stay within \val{r.q.dec.once} points of its letter (\suppword{}).

\paragraph{Planning and answering.}
Question data keeps the VLM's world knowledge and costs planning almost nothing (\suppword{} \cref{tab:supp_closed}): without it, the model is level on navtest and AlpaSim and \val{r.q.f6.pai}~m lower in PAI-AV's 3-s ADE, but falls below \nineb's letter on all \val{r.q.n} benchmarks.
Skipping only stages~1 and~2 leaves planning level and costs DriveLM and 3DSRBench \val{r.q.f1.dl} and \val{r.q.f1.sr} points, a tenth and a third of the loss without question data.

\subsection{Driving results}
\label{sec:main}
\begin{figure*}[t]
\centering
\includegraphics[width=\linewidth]{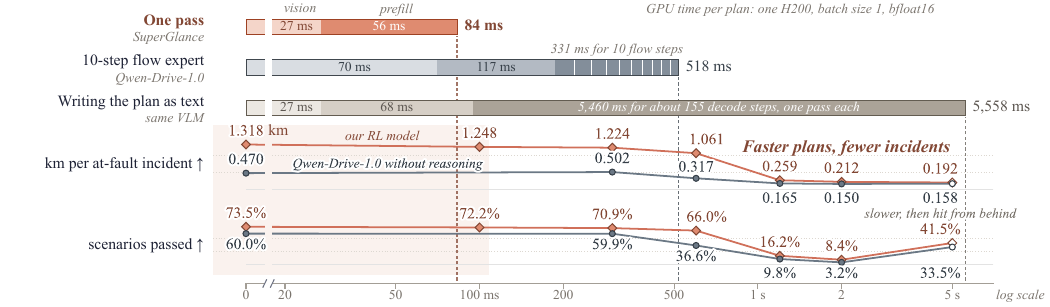}
\caption{\textbf{Time per plan and distance in closed loop.}
Top: a plan takes \val{r.l.ours}~ms of GPU time in our single pass, \val{r.l.qd}~ms in Qwen-Drive-1.0 with 10 flow steps and \val{r.l.text}~ms when our VLM writes it as text.
Bottom: in AlpaSim, each time (dashed) becomes the next 100-ms lag, and both planners keep their distance per at-fault incident within \val{r.dly.w300}\% up to 300~ms.
At 600~ms, Qwen-Drive-1.0 falls to \val{r.dly.qdpkm}~km and ours to \val{r.dly.kmqd}~km; at 1.2~s ours falls to \val{r.dly.km}~km, below Qwen-Drive-1.0 at its own \val{r.l.qd}~ms.
At 5~s every plan arrives after its last pose (hollow): both planners drive more slowly and vehicles behind hit them, which is not at fault and ends the scenario as a pass.}   %
\label{fig:latency}
\end{figure*}

\begin{table}[t]
\centering
\scriptsize
\setlength{\tabcolsep}{1.65pt}   %
\begin{tabular}{lcccccc}
\toprule
Method & RL & Coll.$\downarrow$ & Off-road$\downarrow$ & Prog.$\uparrow$ & km / incident$\uparrow$ & Pass$\uparrow$ \\
\midrule
\grouprow\multicolumn{7}{l}{\textit{Each plan's GPU time as delay}} \\   %
Qwen-Drive-1.0, \val{r.l.qd}~ms & & \heatin{0}{\val{r.dly.qdpcol}}/\heatin{35}{\val{r.dly.qdpcolf}} & \heat{55}{\val{r.dly.qdpoff}} & \heat{47}{\val{r.dly.qdpprog}} & \heatin{30}{\val{r.dly.qdpkma}}/\heatin{32}{\val{r.dly.qdpkm}} & \heat{47}{\val{r.dly.qdppass}} \\
\quad + reasoning, 1.2~s & & \heatin{39}{\val{r.dly.qdcol}}/\heatin{0}{\val{r.dly.qdcolf}} & \heat{0}{\val{r.dly.qdoff}} & \heat{0}{\val{r.dly.qdprog}} & \heatin{0}{\val{r.dly.qdkma}}/\heatin{0}{\val{r.dly.qdkm}} & \heat{0}{\val{r.dly.qdpass}} \\
\ours \method, \val{r.l.ours}~ms & \checkmark & \heatin{100}{\val{r.dly.col}}/\heatin{100}{\val{r.dly.colf}} & \heat{100}{\val{r.dly.off}} & \heat{100}{\val{r.dly.prog}} & \heatin{100}{\val{r.dly.kma}}/\heatin{100}{\val{r.dly.kmown}} & \heat{100}{\val{r.dly.passown}} \\
\midrule
\grouprow\multicolumn{7}{l}{\textit{No delay (simulator default)}} \\
Qwen-Drive-1.0~\cite{qwendrive} & & \heatin{33}{\val{r.as.qdsft.col}}/\heatin{23}{\val{r.as.qdsft.colf}} & \heat{0}{\val{r.as.qdsft.off}} & \heat{39}{\val{r.as.qdsft.prog}} & \heatin{22}{\val{r.as.qdsft.km}}/\heatin{21}{\val{r.as.qdsft.kmf}} & \heat{29}{\val{r.as.qdsft.pass}} \\
Qwen-Drive-1.0 & \checkmark & \heatin{0}{\val{r.as.qdrl.col}}/\heatin{0}{\val{r.as.qdrl.colf}} & \heat{31}{\val{r.as.qdrl.off}} & \heat{0}{\val{r.as.qdrl.prog}} & \heatin{0}{\val{r.as.qdrl.km}}/\heatin{0}{\val{r.as.qdrl.kmf}} & \heat{0}{\val{r.as.qdrl.pass}} \\
\ours \method & & \heatin{82}{\val{r.as.sft.col}}/\heatin{83}{\val{r.as.sft.colf}} & \heat{76}{\val{r.as.sft.off}} & \heat{82}{\val{r.as.sft.prog}} & \heatin{78}{\val{r.as.sft.km}}/\heatin{77}{\val{r.as.sft.kmf}} & \heat{80}{\val{r.as.sft.pass}} \\
\ours \method & \checkmark & \heatin{100}{\val{r.as.rl.col}}/\heatin{100}{\val{r.as.rl.colf}} & \heat{100}{\val{r.as.rl.off}} & \heat{100}{\val{r.as.rl.prog}} & \heatin{100}{\val{r.as.rl.km}}/\heatin{100}{\val{r.as.rl.kmf}} & \heat{100}{\val{r.as.rl.pass}} \\
\bottomrule
\end{tabular}
\caption{\textbf{AlpaSim closed loop.}
Top: each plan's GPU time acts as its delay, a lag of 600~ms, 1.2~s or 100~ms (\suppword{} \cref{sec:supp_protocols}); bottom: no delay.
Qwen-Drive-1.0 plans with reasoning except in its \val{r.l.qd}~ms row.
Coll.\ and km\,/\,incident: all\,/\,at-fault; Coll., off-road, progress and pass in \%; shading ranks each half.}
\label{tab:alpasim}
\vspace{-6pt}   %
\end{table}
\paragraph{AlpaSim.}
Time is distance in closed loop (\cref{tab:alpasim,fig:latency}): with each plan's GPU time as its delay, our RL model drives \val{r.as.x} times as far per at-fault incident as Qwen-Drive-1.0 (95\% interval \val{r.as.x.ci}) and passes \val{r.dly.passown}\% of scenarios, against at most \val{r.dly.qdppass}\%.
Delay costs distance past 300~ms, where successive plans stop agreeing (\suppword{}): up to that lag, both planners keep their km per at-fault incident within \val{r.dly.w300}\%; at Qwen-Drive-1.0's own \val{r.l.qd}~ms, it loses a factor of \val{r.dly.qdlagx} and ours \val{r.dly.lagx}, and at 1.2~s ours loses \val{r.dly.lagx12}.
Without delay (\suppword{} \cref{fig:closedloop}), our supervised model drives \val{r.as.x0} and \val{r.as.x0rl} times as far per at-fault incident as Qwen-Drive-1.0's two planners, and every one-pass variant in \suppword{} \cref{tab:supp_closed} at least \val{r.cl.xqd} times.
RL keeps the mean scenario score level (\val{r.as.rlsft.d}, 95\% interval \val{r.as.rlsft.ci}).

\begin{table}[t]
\centering
\scriptsize
\setlength{\tabcolsep}{5.1pt}
\begin{tabular}{lcccccc}
\toprule
Method & NC$\uparrow$ & DAC$\uparrow$ & EP$\uparrow$ & TTC$\uparrow$ & Comf.$\uparrow$ & PDMS$\uparrow$ \\
\midrule
\grouprow\multicolumn{7}{l}{\textit{Supervised}} \\
TransFuser~\cite{transfuser} & \heat{28}{97.7} & \heat{9}{92.8} & \heat{27}{79.2} & \heat{38}{92.8} & 100.0 & \heat{23}{84.0} \\
DRAMA~\cite{drama} & \heat{49}{98.0} & \heat{17}{93.1} & \heat{37}{80.1} & \heat{88}{94.8} & 100.0 & \heat{36}{85.5} \\
Hydra-MDP~\cite{hydramdp} & \heat{76}{98.3} & \heat{59}{96.0} & \heat{19}{78.7} & \heat{75}{94.6} & 100.0 & \heat{53}{86.5} \\
DiffusionDrive~\cite{diffusiondrive} & \heat{67}{98.2} & \heat{66}{96.2} & \heat{70}{82.2} & \heat{82}{94.7} & 100.0 & \heat{73}{88.1} \\
Epona~\cite{epona} & \heat{40}{97.9} & \heat{45}{95.1} & \heat{44}{80.4} & \heat{51}{93.8} & 99.9 & \heat{45}{86.2} \\
ReCogDrive~\cite{recogdrive} & \heat{58}{98.1} & \heat{36}{94.7} & \heat{52}{80.9} & \heat{59}{94.2} & 100.0 & \heat{53}{86.5} \\
AutoVLA~\cite{autovla} & \heat{0}{96.9} & \heat{0}{92.4} & \heat{0}{75.8} & \heat{0}{88.1} & 99.9 & \heat{0}{80.5} \\
Qwen-Drive-1.0~\cite{qwendrive} & \heat{67}{98.2} & \heat{73}{96.4} & \heat{63}{82.0} & \heat{67}{94.4} & 100.0 & \heat{65}{87.8} \\
\quad + reasoning & \heat{85}{98.4} & \heat{81}{96.6} & \heat{76}{82.4} & \heat{82}{94.7} & 100.0 & \heat{80}{88.2} \\
\ours \method & \heat{100}{\vone{r.nav.sft.nc}} & \heat{100}{\vone{r.nav.sft.dac}} & \heat{100}{\vone{r.nav.sft.ep}} & \heat{100}{\vone{r.nav.sft.ttc}} & \vone{r.nav.sft.c} & \heat{100}{\vone{r.nav.sft}} \\
\midrule
\grouprow\multicolumn{7}{l}{\textit{RL}} \\
ReCogDrive & \heat{0}{97.9} & \heat{37}{97.3} & \heat{100}{87.3} & \heat{0}{94.9} & 100.0 & \heat{73}{90.8} \\
AutoVLA & \heat{30}{98.4} & \heat{0}{95.6} & \heat{0}{81.9} & \heat{100}{98.0} & 99.9 & \heat{0}{89.1} \\
ExploreVLA~\cite{explorevla} & \heat{70}{98.8} & \heat{85}{98.4} & \heat{27}{83.5} & \heat{56}{96.5} & 99.9 & \heat{40}{90.4} \\
EponaV2~\cite{eponav2} & \heat{50}{98.6} & \heat{57}{97.9} & \heat{64}{84.8} & \heat{23}{95.7} & 100.0 & \heat{40}{90.4} \\
Qwen-Drive-1.0 & \heat{50}{98.6} & \heat{72}{98.2} & \heat{64}{84.8} & \heat{36}{95.9} & 100.0 & \heat{58}{90.7} \\
\ours \method & \heat{100}{\vone{r.nav.rl.nc}} & \heat{100}{\vone{r.nav.rl.dac}} & \heat{44}{\vone{r.nav.rl.ep}} & \heat{84}{\vone{r.nav.rl.ttc}} & \vone{r.nav.rl.c} & \heat{100}{\vone{r.nav.rl}} \\
\bottomrule
\end{tabular}
\caption{\textbf{NAVSIM v1.1 navtest.}
One plan per scene; shading ranks the methods within each group, and Comf.\ is unshaded.}
\label{tab:navsim}
\vspace{-6pt}   %
\end{table}

\paragraph{NAVSIM.}
Supervised training alone puts \method above every other method in \cref{tab:navsim}, owing to its reward target and the top-\val{r.pai.topk} mean (\cref{sec:select,sec:closed}).
RL then trades progress for safety: the exact expectation (\cref{eq:rl}) raises NC and TTC (\suppword{} \cref{tab:abl_rl}), and the RL model reaches \vone{r.nav.rl} PDMS, \vone{r.nav.rl.dqd} above Qwen-Drive-1.0 after RL.

\sgtable{t}{b}
\centering
\scriptsize
\setlength{\tabcolsep}{6.03pt}
\begin{tabular}{lcccc}
\toprule
Method & RL & 3-s ADE$\downarrow$ & 5-s ADE$\downarrow$ & RFS$\uparrow$ \\
\midrule
MindVLA-U1~\cite{mindvlau1} & & \heat{70}{1.16} & \heat{48}{2.67} & \heat{34}{7.77} \\
AutoVLA~\cite{autovla} & \checkmark & \heat{0}{1.35} & \heat{0}{2.96} & \heat{0}{7.56} \\
MindVLA-U1 & \checkmark & \heat{100}{1.09} & \heat{62}{2.66} & \heat{66}{7.87} \\
\midrule
Qwen-Drive-1.0 + reasoning~\cite{qwendrive} & & \heat{47}{1.19} & \heat{75}{2.65} & \heat{46}{7.78} \\
Qwen-Drive-1.0 + reasoning & \checkmark & \heat{47}{1.19} & \heat{48}{2.67} & \heat{81}{7.91} \\
\midrule
\ours \method & \checkmark & \heat{47}{\vtwo{r.wod.test.adet}} & \heat{100}{\vtwo{r.wod.test.adef}} & \heat{100}{\vtwo{r.wod.test}} \\
\bottomrule
\end{tabular}
\caption{\textbf{WOD-E2E test split.}
ADE in meters; our 4-s plans are extended to 5~s at constant velocity before scoring.}
\label{tab:wod}
\vspace{-6pt}   %
\end{table}   %

\begin{table}[t]
\centering
\scriptsize
\setlength{\tabcolsep}{5.94pt}
\begin{tabular}{lccccc}
\toprule
& & \multicolumn{2}{c}{One plan} & \multicolumn{2}{c}{Best of six} \\
\cmidrule(lr){3-4}\cmidrule(l){5-6}
Method & RL & 3\,s$\downarrow$ & 5\,s$\downarrow$ & 3\,s$\downarrow$ & 5\,s$\downarrow$ \\
\midrule
DriveWAM~\cite{drivewam}$^\dagger$ & & \heat{0}{0.69\phantom{0}} & \na & \heat{37}{0.38\phantom{0}} & \na \\
\midrule
Qwen-Drive-1.0 + reasoning~\cite{qwendrive} & & \heat{69}{\val{r.pai.qdsft.t}} & \heat{25}{\val{r.pai.qdsft.f}} & \heat{23}{\val{r.pai.qdsft.mt}} & \heat{18}{\val{r.pai.qdsft.mf}} \\
Qwen-Drive-1.0 + reasoning & \checkmark & \heat{50}{\val{r.pai.qdrl.t}} & \heat{0}{\val{r.pai.qdrl.f}} & \heat{0}{\val{r.pai.qdrl.mt}} & \heat{0}{\val{r.pai.qdrl.mf}} \\
\midrule
\ours \method & & \heat{84}{\val{r.pai.sft.t}} & \heat{73}{\val{r.pai.sft.f}} & \heat{83}{\val{r.pai.sft.mt}} & \heat{77}{\val{r.pai.sft.mf}} \\
\ours \method & \checkmark & \heat{100}{\val{r.pai.rl.t}} & \heat{100}{\val{r.pai.rl.f}} & \heat{100}{\val{r.pai.rl.mt}} & \heat{100}{\val{r.pai.rl.mf}} \\
\bottomrule
\end{tabular}
\caption{\textbf{PAI-AV open loop.} ADE (m)$\downarrow$ on 700 held-out frames.
One plan: our top-\val{r.pai.topk} mean and Qwen-Drive-1.0's first sample; best of six: our six most probable candidates and its six samples.
We extend our 4-s plans to 5~s at constant velocity; $^\dagger$: reproduced on these frames in the Qwen-Drive-1.0 paper; \na: not evaluated.}
\label{tab:pai}
\vspace{-6pt}   %
\end{table}
\paragraph{WOD-E2E and PAI-AV.}
No WOD-E2E trajectory is among the candidates, yet our plan scores \vtwo{r.wod.test} RFS on its long-tail test scenes, \val{r.wod.dtest} above Qwen-Drive-1.0 after RL (\cref{tab:wod}).
On PAI-AV, our single plan lands closer to the logged future than either Qwen-Drive-1.0 planner at both horizons, even when that planner keeps the best of six samples, and the best of our six most probable candidates is closer still (\cref{tab:pai}).
Most of RL's PAI-AV gain over our supervised model comes from the \val{r.ab.v2.n} candidates built by rule for braking and medium-speed curves (\suppword{}).

\begin{figure*}[t]
\centering
\includegraphics[width=\linewidth]{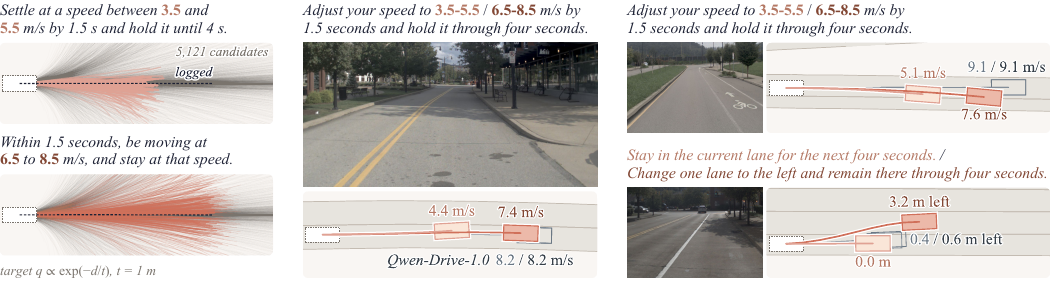}
\caption{\textbf{Language steers the plan.}
Each instruction selects its own plan from the same candidates.
Left: two instructions on a training scene, each with the candidates that follow it (coral); line weight is the target $q\propto\exp(-d/t)$, $t=1$~m, with $d$ the distance to the logged future in what the instruction leaves free (across the path for a speed instruction).
Right: plans for both instructions (light, dark) in three Language2Drive scenes, boxes at 4~s; ours (coral) change with the instruction, Qwen-Drive-1.0's (slate) barely.}
\label{fig:language}
\end{figure*}
\paragraph{Manoeuvre and choice.}
One pass settles the manoeuvre and leaves the choice within it. On navtest, our RL model's most probable candidate takes the driver's manoeuvre in \val{r.tk.rl.top1}\% of scenes (the driver goes straight in \val{r.tk.straight}\%; \suppword{} \cref{fig:qual}a), and its six most probable candidates end \val{r.tk.rl.spread}~m apart at 4~s (median of mean pairwise distances), against \val{r.tk.rand.spread}~m for six random ones.
Within the manoeuvre, the best of the six still scores \val{r.tk.rl.k6} PDMS against \val{r.tk.rl.k1} for the most probable one (\suppword{} \cref{fig:qual}b).

\begin{table}[t]
\centering
\scriptsize
\setlength{\tabcolsep}{4pt}
\begin{tabular}{lcccc}
\toprule
Method & SemSISR$\uparrow$ & SemCFPS$\uparrow$ & SCS$\uparrow$ & PCS$\uparrow$ \\
\midrule
Qwen-Drive-1.0 (4B)~\cite{qwendrive} & \heat{24}{15.8} & \heat{0}{0.0} & \heat{10}{25.3} & \heat{0}{2.8} \\
AutoVLA~\cite{autovla} & \heat{37}{18.8} & \heat{0}{0.0} & \heat{38}{30.2} & \heat{27}{5.4} \\
DriveMA-2B~\cite{drivema} & \heat{48}{20.4} & \heat{0}{0.0} & \heat{52}{30.6} & \heat{45}{6.7} \\
ReCogDrive~\cite{recogdrive} & \heat{11}{13.2} & \heat{0}{0.0} & \heat{0}{22.5} & \heat{0}{2.8} \\
OneVL~\cite{onevl} & \heat{57}{20.7} & \heat{0}{0.0} & \heat{45}{30.5} & \heat{10}{4.6} \\
\midrule
\method, no instruction & \heat{0}{11.8} & \heat{0}{0.0} & \heat{19}{26.2} & \heat{36}{6.2} \\
\method, no instruction rows & \heat{24}{15.8} & \heat{0}{0.0} & \heat{27}{27.2} & \heat{18}{4.7} \\   %
\ours \method & \heat{100}{40.5} & \heat{100}{21.1} & \heat{100}{65.0} & \heat{100}{56.5} \\
\bottomrule
\end{tabular}
\caption{\textbf{Language2Drive, NAVSIM track.} \val{r.l2d.n} instruction pairs in four wordings each, scored with its current scorer; no instruction rows: supervised, none in training.
SemSISR, SemCFPS: single instructions and pairs followed (\%); SCS, PCS: graded scores.}
\label{tab:l2d}
\vspace{-6pt}   %
\end{table}   %
\paragraph{Instructions.}
Under Language2Drive's current scorer~\cite{l2d}, our policy follows instruction pairs its five VLAs miss.
The benchmark gives two instructions per scene, such as two speed bands, and checks whether each plan follows its own.
Our RL model follows both in \val{r.l2d.cfps}\% of pairs, against \val{r.l2d.cfpsvla}\% for all five and for ours without the instruction (\cref{tab:l2d,fig:language}).
As in language models, each ability comes with its own rows and costs the others little (\cref{sec:vqa}): DriveLM questions lift DriveLM (\cref{tab:vqa}), and instruction rows in this style, on training scenes only, bring instruction following; without them, our supervised model follows none.

\paragraph{History and rollout rows.}
Closed loop shows what navtest does not.
For our supervised model before its last phase, dropping the history frames doubles the AlpaSim scenarios with an at-fault collision, from \val{r.cl.af.base} to \val{r.cl.af.nh}, and dropping the rollout rows raises them to \val{r.cl.af.nr}, while navtest moves by under half a point (\suppword{} \cref{tab:supp_closed}).
Without history, front hits on stopped vehicles rise from \val{r.cl.st.base} to \val{r.cl.st.nh}.
In behavior cloning, by contrast, history is reported to help offline yet hurt closed-loop driving by copying past actions~\cite{dehaan2019,wen2020copycat}.

\subsection{Ablations on output and target}
\label{sec:closed}
\label{sec:ablation}
\begin{table}[t]
\centering
\scriptsize
\setlength{\tabcolsep}{5pt}
\begin{tabular}{llccc}
\toprule
& & Open loop & \multicolumn{2}{c}{Closed loop (AlpaSim)} \\
\cmidrule(lr){3-3}\cmidrule(l){4-5}
Output & Target & PDMS$\uparrow$ & km / incident$\uparrow$ & Pass (\%)$\uparrow$ \\
\midrule
Text, decoded & Imitation & \heatv{40}{\vone{r.cl.txt.nav}} & \na & \na \\
Linear regression & Imitation & \heatv{56}{\vone{r.cl.reg.nav}} & \heatv{77}{\val{r.cl.reg.km}} & \heatv{56}{\val{r.cl.reg.pass}} \\
Policy Head & Imitation & \heatv{0}{\vone{r.ab.ih.nav}} & \heatv{53}{\val{r.cl.im.km}} & \heatv{75}{\val{r.cl.im.pass}} \\
\ours Policy Head & \ours Reward & \heatv{100}{\vone{r.ab.rh.nav}} & \heatv{100}{\val{r.cl.base.km}} & \heatv{100}{\val{r.cl.base.pass}} \\
\midrule
\multicolumn{2}{l}{Qwen-Drive-1.0 + reasoning~\cite{qwendrive}} & \heatv{71}{88.2} & \heatv{0}{\val{r.as.qdsft.kmf}} & \heatv{0}{\val{r.as.qdsft.pass}} \\
\bottomrule
\end{tabular}
\caption{\textbf{Output and target.} Variants of our supervised model before its last phase, with Qwen-Drive-1.0 for reference. Imitation: the target follows the human trajectory; \na: not evaluated (writing the plan as text takes \val{r.l.text}~ms, longer than the 4~s it covers). PDMS is on navtest; AlpaSim without delay, km per at-fault incident.}
\label{tab:closed}
\vspace{-6pt}   %
\end{table}

The hidden state before the first token is enough to plan: given the same data, inputs and imitation target, linear regression on it plans at least as well as the same VLM writing the plan as text (\cref{tab:closed}; level on PAI-AV and WOD-E2E, \suppword{}).
The regression also has the lowest single-plan PAI-AV ADE of any model we ran, \val{r.cl.reg.ade}~m at 3~s (\suppword{} \cref{tab:supp_closed}).
A fixed set pays off through its target.
Under imitation it trails the regression on navtest, partly because even the candidate nearest the human trajectory reaches only \vone{r.k.nearnav} PDMS on held-out training scenes.
The reward target lifts the Policy Head to \vone{r.ab.rh.nav} and raises AlpaSim's mean scenario score (\suppword{} \cref{tab:supp_closed}).

\section{Conclusion}
\label{sec:conclusion}
We find that before a VLM writes its first token, it has already read the scene into one hidden state, which, once trained, is enough to drive and to answer questions about what the model sees.
\method reads it to turn a decoder-only VLM into an encoder-style driving policy whose fixed candidates let targets weight plans by reward, RL maximize the exact expected reward and instructions pick among them.
Questions built against shortcuts, each answered in both option orders, keep the VLM's visual abilities in one pass.
One pass also reacts faster, and time is distance: with each plan's GPU time as delay, our lead in km per at-fault incident over Qwen-Drive-1.0 grows from \val{r.dly.xqd0a} to \val{r.as.x} times.
One model answers questions on \val{r.q.n} benchmarks and plans in \val{r.l.ours}~ms of GPU time: a path to System~1 in the VLM.

\ifdefempty{\sgstatement}{}{\section*{\sgstatementhead}{\widowpenalties 3 10000 10000 0 \sgstatement\par}}   %

\phantomsection\addcontentsline{toc}{section}{References}   %
\bibliographystyle{ieeenat_fullname_unsrt}
\bibliography{refs,refs_extra}

\begin{thebibliography}{104}
\providecommand{\natexlab}[1]{#1}
\providecommand{\url}[1]{\texttt{#1}}
\expandafter\ifx\csname urlstyle\endcsname\relax
  \providecommand{\doi}[1]{doi: #1}\else
  \providecommand{\doi}{doi: \begingroup \urlstyle{rm}\Url}\fi

\bibitem[Zhou et~al.(2026)Zhou, Zhao, Yang, et~al.]{qwendrive}
Xin Zhou, Zongchuang Zhao, Zhibo Yang, et~al.
\newblock {Qwen-Drive-1.0: An Initial Step towards a Vision-Language Foundation
  Model for Autonomous Driving}.
\newblock \emph{arXiv preprint arXiv:2609.00111}, 2026.

\bibitem[Zhou et~al.(2025)Zhou, Cai, Zhao, et~al.]{autovla}
Zewei Zhou, Tianhui Cai, Seth~Z. Zhao, et~al.
\newblock {AutoVLA: A Vision-Language-Action Model for End-to-End Autonomous
  Driving with Adaptive Reasoning and Reinforcement Fine-Tuning}.
\newblock In \emph{NeurIPS}, 2025.

\bibitem[Sima et~al.(2024)Sima, Renz, Chitta, et~al.]{drivelm}
Chonghao Sima, Katrin Renz, Kashyap Chitta, et~al.
\newblock {DriveLM: Driving with Graph Visual Question Answering}.
\newblock In \emph{ECCV}, 2024.

\bibitem[Tian et~al.(2024)Tian, Gu, Li, et~al.]{drivevlm}
Xiaoyu Tian, Junru Gu, Bailin Li, et~al.
\newblock {DriveVLM: The Convergence of Autonomous Driving and Large
  Vision-Language Models}.
\newblock In \emph{CoRL}, 2024.

\bibitem[Li et~al.(2026)Li, Xiong, Guo, et~al.]{recogdrive}
Yongkang Li, Kaixin Xiong, Xiangyu Guo, et~al.
\newblock {ReCogDrive: A Reinforced Cognitive Framework for End-to-End
  Autonomous Driving}.
\newblock In \emph{ICLR}, 2026.

\bibitem[Luo et~al.(2026)Luo, Li, Xu, et~al.]{adathinkdrive}
Yuechen Luo, Fang Li, Shaoqing Xu, et~al.
\newblock {AdaThinkDrive: Adaptive Thinking via Reinforcement Learning for
  Autonomous Driving}.
\newblock In \emph{ICRA}, 2026.

\bibitem[Chen et~al.(2026)Chen, Yang, Tan, et~al.]{curiousvla}
Canyu Chen, Yuguang Yang, Zhewen Tan, et~al.
\newblock {Devil is in Narrow Policy: Unleashing Exploration in Driving VLA
  Models}.
\newblock In \emph{Findings of CVPR}, 2026.

\bibitem[Yang et~al.(2026)Yang, Chen, Tan, et~al.]{driveteach}
Yuguang Yang, Canyu Chen, Zhewen Tan, et~al.
\newblock {Teaching Vision-Language-Action Models What to See and Where to
  Look}.
\newblock In \emph{ECCV}, 2026.

\bibitem[Kahneman(2011)]{kahneman}
Daniel Kahneman.
\newblock \emph{Thinking, Fast and Slow}.
\newblock Farrar, Straus and Giroux, 2011.

\bibitem[Renz et~al.(2025)Renz, Chen, Arani, and Sinavski]{simlingo}
Katrin Renz, Long Chen, Elahe Arani, and Oleg Sinavski.
\newblock {SimLingo}: Vision-only closed-loop autonomous driving with
  language-action alignment.
\newblock In \emph{CVPR}, 2025.

\bibitem[Devlin et~al.(2019)Devlin, Chang, Lee, and Toutanova]{bert}
Jacob Devlin, Ming-Wei Chang, Kenton Lee, and Kristina Toutanova.
\newblock {BERT}: Pre-training of deep bidirectional transformers for language
  understanding.
\newblock In \emph{NAACL-HLT}, 2019.

\bibitem[Radford et~al.(2018)Radford, Narasimhan, Salimans, and
  Sutskever]{gpt1}
Alec Radford, Karthik Narasimhan, Tim Salimans, and Ilya Sutskever.
\newblock Improving language understanding by generative pre-training.
\newblock Technical report, OpenAI, 2018.

\bibitem[{NVIDIA} et~al.(2025){NVIDIA}, Cao, de~Lutio, et~al.]{alpasim}
{NVIDIA}, Yulong Cao, Riccardo de Lutio, et~al.
\newblock {AlpaSim}: A modular, lightweight, and data-driven research simulator
  for autonomous driving.
\newblock \url{https://github.com/NVlabs/alpasim}, 2025.

\bibitem[Liu et~al.(2026)Liu, Li, Wang, Zhang, and He]{anchorvla}
Qi Liu, Yabei Li, Hongsong Wang, Heng Zhang, and Lei He.
\newblock {AnchorVLA: Bridging Discrete Decisions and Continuous Trajectories
  for Vision-Language-Action Planning}.
\newblock \emph{arXiv preprint arXiv:2607.03182}, 2026.

\bibitem[Li et~al.(2026)Li, Chang, Yao, et~al.]{ztrs}
Zhenxin Li, Nadine Chang, Wenhao Yao, et~al.
\newblock {Zero-Human Demonstration End-to-end Autonomous Driving with
  Trajectory Scorer}.
\newblock In \emph{ECCV}, 2026.

\bibitem[Dang et~al.(2026)Dang, Ang, Li, et~al.]{drivefine}
Chenxu Dang, Sining Ang, Yongkang Li, et~al.
\newblock {DriveFine: Refining-Augmented Masked Diffusion VLA for Efficient and
  Robust Driving}.
\newblock In \emph{ECCV}, 2026.

\bibitem[Xu et~al.(2026)Xu, Cui, Zhu, et~al.]{wamflow}
Yifang Xu, Jiahao Cui, Zhihao Zhu, et~al.
\newblock {WAM-Flow: Parallel Coarse-to-Fine Motion Planning via Discrete Flow
  Matching for Autonomous Driving}.
\newblock In \emph{CVPR}, 2026.

\bibitem[Zhang et~al.(2025)Zhang, Tang, Hu, et~al.]{epona}
Kaiwen Zhang, Zhenyu Tang, Xiaotao Hu, et~al.
\newblock {Epona}: Autoregressive diffusion world model for autonomous driving.
\newblock In \emph{ICCV}, 2025.

\bibitem[Xu et~al.(2026)Xu, Zhong, Shu, et~al.]{eponav2}
Jiawei Xu, Zhizhou Zhong, Zhijian Shu, et~al.
\newblock {EponaV2}: Driving world model with comprehensive future reasoning.
\newblock \emph{arXiv preprint arXiv:2605.14696}, 2026.

\bibitem[Shi et~al.(2026)Shi, Xu, Shi, Sheng, Zhang, and Jiang]{drivewam}
Chen Shi, Jinrui Xu, Shaoshuai Shi, Kehua Sheng, Bo Zhang, and Li Jiang.
\newblock {DriveWAM}: Video generative priors enable scalable world-action
  modeling for autonomous driving.
\newblock \emph{arXiv preprint arXiv:2605.28544}, 2026.

\bibitem[Zhao et~al.(2026)Zhao, Zhou, Xu, et~al.]{simwam}
Zongchuang Zhao, Xin Zhou, Tianyang Xu, et~al.
\newblock {SimWAM}: A simple world action model for end-to-end autonomous
  driving.
\newblock \emph{arXiv preprint arXiv:2608.07468}, 2026.

\bibitem[Mei et~al.(2024)Mei, Ma, Yang, et~al.]{leapad}
Jianbiao Mei, Yukai Ma, Xuemeng Yang, et~al.
\newblock Continuously learning, adapting, and improving: A dual-process
  approach to autonomous driving.
\newblock In \emph{NeurIPS}, 2024.

\bibitem[Zhang et~al.(2025)Zhang, Xie, Zhang, et~al.]{adadrive}
Ruifei Zhang, Junlin Xie, Wei Zhang, et~al.
\newblock {AdaDrive}: Self-adaptive slow-fast system for language-grounded
  autonomous driving.
\newblock In \emph{ICCV}, 2025.

\bibitem[Li et~al.(2026)Li, Shang, Liu, et~al.]{drivevlaw0}
Yingyan Li, Shuyao Shang, Weisong Liu, et~al.
\newblock {DriveVLA-W0: World Models Amplify Data Scaling Law in Autonomous
  Driving}.
\newblock In \emph{ICLR}, 2026.

\bibitem[Liu et~al.(2026)Liu, Ye, Zhang, and Qi]{critiquedrivevlm}
Zhaohong Liu, Hao Ye, Xianlin Zhang, and Mengshi Qi.
\newblock {CritiqueDriveVLM}: From verifier-guided reinforcement learning to
  latent thought distillation for autonomous driving.
\newblock In \emph{ECCV}, 2026.

\bibitem[Renz et~al.(2024)Renz, Chen, Marcu, et~al.]{carllava}
Katrin Renz, Long Chen, Ana-Maria Marcu, et~al.
\newblock {CarLLaVA}: Vision language models for camera-only closed-loop
  driving.
\newblock \emph{arXiv preprint arXiv:2406.10165}, 2024.

\bibitem[Zhang et~al.(2026)Zhang, Yuan, Chen, et~al.]{reasoningvla}
Dapeng Zhang, Zhenlong Yuan, Zhangquan Chen, et~al.
\newblock {Reasoning-VLA}: An efficient and spatial-guided general
  vision-language-action reasoning model for autonomous driving.
\newblock In \emph{ICML}, 2026.

\bibitem[Kim et~al.(2025)Kim, Finn, and Liang]{openvlaoft}
Moo~Jin Kim, Chelsea Finn, and Percy Liang.
\newblock Fine-tuning vision-language-action models: Optimizing speed and
  success.
\newblock In \emph{RSS}, 2025.

\bibitem[Fu et~al.(2025)Fu, Zhang, Zhao, et~al.]{orion}
Haoyu Fu, Diankun Zhang, Zongchuang Zhao, et~al.
\newblock {ORION}: A holistic end-to-end autonomous driving framework by
  vision-language instructed action generation.
\newblock In \emph{ICCV}, 2025.

\bibitem[Chen et~al.(2025)Chen, Liu, Gu, et~al.]{fisvla}
Hao Chen, Jiaming Liu, Chenyang Gu, et~al.
\newblock Fast-in-slow: A dual-system {VLA} model unifying fast manipulation
  within slow reasoning.
\newblock In \emph{NeurIPS}, 2025.

\bibitem[Shao et~al.(2024)Shao, Wang, Zhu, et~al.]{deepseekmath}
Zhihong Shao, Peiyi Wang, Qihao Zhu, et~al.
\newblock {DeepSeekMath}: Pushing the limits of mathematical reasoning in open
  language models.
\newblock \emph{arXiv preprint arXiv:2402.03300}, 2024.

\bibitem[Jiang et~al.(2026)Jiang, Chen, Gao, et~al.]{vadv2}
Bo Jiang, Shaoyu Chen, Hao Gao, et~al.
\newblock {VADv2: End-to-End Vectorized Autonomous Driving via Probabilistic
  Planning}.
\newblock In \emph{ICLR}, 2026.

\bibitem[Li et~al.(2024)Li, Li, Wang, et~al.]{hydramdp}
Zhenxin Li, Kailin Li, Shihao Wang, et~al.
\newblock {Hydra-MDP: End-to-end Multimodal Planning with Multi-target
  Hydra-Distillation}.
\newblock \emph{arXiv preprint arXiv:2406.06978}, 2024.

\bibitem[Yao et~al.(2026)Yao, Li, Lan, et~al.]{drivesuprim}
Wenhao Yao, Zhenxin Li, Shiyi Lan, et~al.
\newblock {DriveSuprim: Towards Precise Trajectory Selection for End-to-End
  Planning}.
\newblock In \emph{AAAI}, 2026.

\bibitem[Sun et~al.(2026)Sun, Lin, Chen, et~al.]{sparsedrivev2}
Wenchao Sun, Xuewu Lin, Keyu Chen, et~al.
\newblock {SparseDriveV2: Scoring is All You Need for End-to-End Autonomous
  Driving}.
\newblock In \emph{ECCV}, 2026.

\bibitem[Kirby et~al.(2026)Kirby, Boulch, Xu, et~al.]{drivor}
Ellington Kirby, Alexandre Boulch, Yihong Xu, et~al.
\newblock {Driving on Registers}.
\newblock In \emph{CVPR}, 2026.

\bibitem[Pei et~al.(2026)Pei, Lu, Ang, et~al.]{world4scorer}
Jieyuan Pei, Meiyi Lu, Sining Ang, et~al.
\newblock {World4Scorer: Outcome-Grounded World Modeling for Autonomous
  Driving}.
\newblock \emph{arXiv preprint arXiv:2609.36438}, 2026.

\bibitem[Li et~al.(2025)Li, Yao, Wang, et~al.]{gtrs}
Zhenxin Li, Wenhao Yao, Zi Wang, et~al.
\newblock {Generalized Trajectory Scoring for End-to-end Multimodal Planning}.
\newblock \emph{arXiv preprint arXiv:2506.06664}, 2025.

\bibitem[Shang et~al.(2025)Shang, Chen, Wang, Li, and Zhang]{drivedpo}
Shuyao Shang, Yuntao Chen, Yuqi Wang, Yingyan Li, and Zhaoxiang Zhang.
\newblock {DriveDPO}: Policy learning via safety {DPO} for end-to-end
  autonomous driving.
\newblock In \emph{NeurIPS}, 2025.

\bibitem[Lu et~al.(2019)Lu, Batra, Parikh, and Lee]{vilbert}
Jiasen Lu, Dhruv Batra, Devi Parikh, and Stefan Lee.
\newblock {ViLBERT}: Pretraining task-agnostic visiolinguistic representations
  for vision-and-language tasks.
\newblock In \emph{NeurIPS}, 2019.

\bibitem[Tan and Bansal(2019)]{lxmert}
Hao Tan and Mohit Bansal.
\newblock {LXMERT}: Learning cross-modality encoder representations from
  transformers.
\newblock In \emph{EMNLP-IJCNLP}, 2019.

\bibitem[Chen et~al.(2020)Chen, Li, Yu, et~al.]{uniter}
Yen-Chun Chen, Linjie Li, Licheng Yu, et~al.
\newblock {UNITER}: {UNiversal} image-{TExt} representation learning.
\newblock In \emph{ECCV}, 2020.

\bibitem[Alayrac et~al.(2022)Alayrac, Donahue, Luc, et~al.]{flamingo}
Jean-Baptiste Alayrac, Jeff Donahue, Pauline Luc, et~al.
\newblock Flamingo: a visual language model for few-shot learning.
\newblock In \emph{NeurIPS}, 2022.

\bibitem[Li et~al.(2023)Li, Li, Savarese, and Hoi]{blip2}
Junnan Li, Dongxu Li, Silvio Savarese, and Steven Hoi.
\newblock {BLIP-2}: Bootstrapping language-image pre-training with frozen image
  encoders and large language models.
\newblock In \emph{ICML}, 2023.

\bibitem[Liu et~al.(2023)Liu, Li, Wu, and Lee]{llava}
Haotian Liu, Chunyuan Li, Qingyang Wu, and Yong~Jae Lee.
\newblock Visual instruction tuning.
\newblock In \emph{NeurIPS}, 2023.

\bibitem[Bai et~al.(2025)Bai, Cai, Chen, et~al.]{qwen3vl}
Shuai Bai, Yuxuan Cai, Ruizhe Chen, et~al.
\newblock {Qwen3-VL Technical Report}.
\newblock \emph{arXiv preprint arXiv:2511.21631}, 2025.

\bibitem[{Qwen Team}(2026)]{qwen35}
{Qwen Team}.
\newblock {Qwen3.5}: Towards native multimodal agents.
\newblock \url{https://qwen.ai/blog?id=qwen3.5}, 2026.

\bibitem[Robinson et~al.(2023)Robinson, Rytting, and Wingate]{mcp}
Joshua Robinson, Christopher~Michael Rytting, and David Wingate.
\newblock {Leveraging Large Language Models for Multiple Choice Question
  Answering}.
\newblock In \emph{ICLR}, 2023.

\bibitem[Zheng et~al.(2024)Zheng, Zhou, Meng, Zhou, and Huang]{pride}
Chujie Zheng, Hao Zhou, Fandong Meng, Jie Zhou, and Minlie Huang.
\newblock {Large Language Models Are Not Robust Multiple Choice Selectors}.
\newblock In \emph{ICLR}, 2024.

\bibitem[Dauner et~al.(2024)Dauner, Hallgarten, Li, et~al.]{navsim}
Daniel Dauner, Marcel Hallgarten, Tianyu Li, et~al.
\newblock {NAVSIM: Data-Driven Non-Reactive Autonomous Vehicle Simulation and
  Benchmarking}.
\newblock In \emph{NeurIPS}, 2024.

\bibitem[Dauner et~al.(2023)Dauner, Hallgarten, Geiger, and Chitta]{pdm}
Daniel Dauner, Marcel Hallgarten, Andreas Geiger, and Kashyap Chitta.
\newblock Parting with misconceptions about learning-based vehicle motion
  planning.
\newblock In \emph{CoRL}, 2023.

\bibitem[Hudson and Manning(2019)]{gqa}
Drew~A. Hudson and Christopher~D. Manning.
\newblock {GQA: A New Dataset for Real-World Visual Reasoning and Compositional
  Question Answering}.
\newblock In \emph{CVPR}, 2019.

\bibitem[Ray et~al.(2025)Ray, Duan, Brown, et~al.]{sat}
Arijit Ray, Jiafei Duan, Ellis Brown, et~al.
\newblock {SAT: Dynamic Spatial Aptitude Training for Multimodal Language
  Models}.
\newblock In \emph{COLM}, 2025.

\bibitem[Acharya et~al.(2019)Acharya, Kafle, and Kanan]{tallyqa}
Manoj Acharya, Kushal Kafle, and Christopher Kanan.
\newblock {TallyQA: Answering Complex Counting Questions}.
\newblock In \emph{AAAI}, 2019.

\bibitem[Kembhavi et~al.(2016)Kembhavi, Salvato, Kolve, Seo, Hajishirzi, and
  Farhadi]{ai2d}
Aniruddha Kembhavi, Mike Salvato, Eric Kolve, Minjoon Seo, Hannaneh Hajishirzi,
  and Ali Farhadi.
\newblock {A Diagram Is Worth A Dozen Images}.
\newblock In \emph{ECCV}, 2016.

\bibitem[Masry et~al.(2022)Masry, Long, Tan, Joty, and Hoque]{chartqa}
Ahmed Masry, Do~Xuan Long, Jia~Qing Tan, Shafiq Joty, and Enamul Hoque.
\newblock {ChartQA: A Benchmark for Question Answering about Charts with Visual
  and Logical Reasoning}.
\newblock In \emph{Findings of ACL}, 2022.

\bibitem[Singh et~al.(2019)Singh, Natarajan, Shah, et~al.]{textvqa}
Amanpreet Singh, Vivek Natarajan, Meet Shah, et~al.
\newblock {Towards VQA Models That Can Read}.
\newblock In \emph{CVPR}, 2019.

\bibitem[{NVIDIA}(2025)]{paiav}
{NVIDIA}.
\newblock Physical {AI} autonomous vehicles dataset.
\newblock
  \url{https://huggingface.co/datasets/nvidia/PhysicalAI-Autonomous-Vehicles},
  2025.

\bibitem[Xu et~al.(2026)Xu, Lin, Jeon, et~al.]{wode2e}
Runsheng Xu, Hubert Lin, Wonseok Jeon, et~al.
\newblock {WOD-E2E}: {W}aymo open dataset for end-to-end driving in challenging
  long-tail scenarios.
\newblock In \emph{CVPR}, 2026.

\bibitem[Hu et~al.(2022)Hu, Shen, Wallis, et~al.]{lora}
Edward~J. Hu, Yelong Shen, Phillip Wallis, et~al.
\newblock {LoRA: Low-Rank Adaptation of Large Language Models}.
\newblock In \emph{ICLR}, 2022.

\bibitem[Brier(1950)]{brier1950}
Glenn~W. Brier.
\newblock {Verification of Forecasts Expressed in Terms of Probability}.
\newblock \emph{Monthly Weather Review}, 78\penalty0 (1):\penalty0 1--3, 1950.

\bibitem[Korbak et~al.(2022)Korbak, Perez, and Buckley]{korbak2022kl}
Tomasz Korbak, Ethan Perez, and Christopher~L. Buckley.
\newblock {RL} with {KL} penalties is better viewed as {B}ayesian inference.
\newblock In \emph{Findings of EMNLP}, 2022.

\bibitem[Rafailov et~al.(2023)Rafailov, Sharma, Mitchell, Ermon, Manning, and
  Finn]{dpo}
Rafael Rafailov, Archit Sharma, Eric Mitchell, Stefano Ermon, Christopher~D.
  Manning, and Chelsea Finn.
\newblock Direct preference optimization: Your language model is secretly a
  reward model.
\newblock In \emph{NeurIPS}, 2023.

\bibitem[Tong et~al.(2024)Tong, Brown, Wu, et~al.]{cambrian}
Shengbang Tong, Ellis Brown, Penghao Wu, et~al.
\newblock {Cambrian-1: A Fully Open, Vision-Centric Exploration of Multimodal
  LLMs}.
\newblock In \emph{NeurIPS}, 2024.

\bibitem[Fu et~al.(2024)Fu, Hu, Li, et~al.]{blink}
Xingyu Fu, Yushi Hu, Bangzheng Li, et~al.
\newblock {BLINK: Multimodal Large Language Models Can See but Not Perceive}.
\newblock In \emph{ECCV}, 2024.

\bibitem[Ma et~al.(2025)Ma, Chen, Zhang, et~al.]{threedsr}
Wufei Ma, Haoyu Chen, Guofeng Zhang, et~al.
\newblock {3DSRBench: A Comprehensive 3D Spatial Reasoning Benchmark}.
\newblock In \emph{ICCV}, 2025.

\bibitem[Hendrycks et~al.(2021)Hendrycks, Burns, Basart, et~al.]{mmlu}
Dan Hendrycks, Collin Burns, Steven Basart, et~al.
\newblock {Measuring Massive Multitask Language Understanding}.
\newblock In \emph{ICLR}, 2021.

\bibitem[Wang et~al.(2024)Wang, Ma, Zhang, et~al.]{mmlupro}
Yubo Wang, Xueguang Ma, Ge Zhang, et~al.
\newblock {MMLU-Pro: A More Robust and Challenging Multi-Task Language
  Understanding Benchmark}.
\newblock In \emph{NeurIPS}, 2024.

\bibitem[Clark et~al.(2018)Clark, Cowhey, Etzioni, et~al.]{arc}
Peter Clark, Isaac Cowhey, Oren Etzioni, et~al.
\newblock {Think you have Solved Question Answering? Try ARC, the AI2 Reasoning
  Challenge}.
\newblock \emph{arXiv preprint arXiv:1803.05457}, 2018.

\bibitem[Zellers et~al.(2019)Zellers, Holtzman, Bisk, Farhadi, and
  Choi]{hellaswag}
Rowan Zellers, Ari Holtzman, Yonatan Bisk, Ali Farhadi, and Yejin Choi.
\newblock {HellaSwag: Can a Machine Really Finish Your Sentence?}
\newblock In \emph{ACL}, 2019.

\bibitem[Sakaguchi et~al.(2020)Sakaguchi, Bras, Bhagavatula, and
  Choi]{winogrande}
Keisuke Sakaguchi, Ronan~Le Bras, Chandra Bhagavatula, and Yejin Choi.
\newblock {WinoGrande: An Adversarial Winograd Schema Challenge at Scale}.
\newblock In \emph{AAAI}, 2020.

\bibitem[Clark et~al.(2019)Clark, Lee, Chang, Kwiatkowski, Collins, and
  Toutanova]{boolq}
Christopher Clark, Kenton Lee, Ming-Wei Chang, Tom Kwiatkowski, Michael
  Collins, and Kristina Toutanova.
\newblock {BoolQ: Exploring the Surprising Difficulty of Natural Yes/No
  Questions}.
\newblock In \emph{NAACL-HLT}, 2019.

\bibitem[Lin et~al.(2022)Lin, Hilton, and Evans]{truthfulqa}
Stephanie Lin, Jacob Hilton, and Owain Evans.
\newblock {TruthfulQA: Measuring How Models Mimic Human Falsehoods}.
\newblock In \emph{ACL}, 2022.

\bibitem[{NVIDIA}(2025)]{painurec}
{NVIDIA}.
\newblock Physical {AI} autonomous vehicles {NuRec} dataset.
\newblock
  \url{https://huggingface.co/datasets/nvidia/PhysicalAI-Autonomous-Vehicles-NuRec},
  2025.

\bibitem[Wu et~al.(2025)Wu, Esturo, Mirzaei, Moenne-Loccoz, and
  Gojcic]{threedgut}
Qi Wu, Janick~Martinez Esturo, Ashkan Mirzaei, Nicolas Moenne-Loccoz, and Zan
  Gojcic.
\newblock {3DGUT}: Enabling distorted cameras and secondary rays in {G}aussian
  splatting.
\newblock In \emph{CVPR}, 2025.

\bibitem[Chitta et~al.(2023)Chitta, Prakash, Jaeger, Yu, Renz, and
  Geiger]{transfuser}
Kashyap Chitta, Aditya Prakash, Bernhard Jaeger, Zehao Yu, Katrin Renz, and
  Andreas Geiger.
\newblock {TransFuser: Imitation with Transformer-Based Sensor Fusion for
  Autonomous Driving}.
\newblock \emph{IEEE TPAMI}, 45\penalty0 (11):\penalty0 12878--12895, 2023.

\bibitem[Yuan et~al.(2024)Yuan, Zhang, Sun, et~al.]{drama}
Chengran Yuan, Zhanqi Zhang, Jiawei Sun, et~al.
\newblock {DRAMA}: An efficient end-to-end motion planner for autonomous
  driving with {M}amba.
\newblock \emph{arXiv preprint arXiv:2408.03601}, 2024.

\bibitem[Liao et~al.(2025)Liao, Chen, Yin, et~al.]{diffusiondrive}
Bencheng Liao, Shaoyu Chen, Haoran Yin, et~al.
\newblock {DiffusionDrive}: Truncated diffusion model for end-to-end autonomous
  driving.
\newblock In \emph{CVPR}, 2025.

\bibitem[Sheng et~al.(2026)Sheng, Ye, Luo, Chen, and Ren]{explorevla}
Zihao Sheng, Xin Ye, Jingru Luo, Sikai Chen, and Liu Ren.
\newblock {ExploreVLA}: Dense world modeling and exploration for end-to-end
  autonomous driving.
\newblock In \emph{ECCV}, 2026.

\bibitem[Huang et~al.(2026)Huang, Zhu, Lu, et~al.]{mindvlau1}
Yuzhou Huang, Benjin Zhu, Hengtong Lu, et~al.
\newblock {MindVLA-U1}: {VLA} beats {VA} with unified streaming architecture
  for autonomous driving.
\newblock \emph{arXiv preprint arXiv:2605.12624}, 2026.

\bibitem[Zheng et~al.(2026)Zheng, Huang, Sun, Li, and Zhao]{drivema}
Weicheng Zheng, Yixin Huang, Qiao Sun, Derun Li, and Hang Zhao.
\newblock {DriveMA}: Driving vision-language-action models with verifiable
  meta-actions.
\newblock In \emph{CoRL}, 2026.

\bibitem[Lu et~al.(2026)Lu, Guan, Huang, et~al.]{onevl}
Jinghui Lu, Jiayi Guan, Zhijian Huang, et~al.
\newblock Xiaomi {OneVL}: One-step latent reasoning and planning with
  vision-language explanation.
\newblock \emph{arXiv preprint arXiv:2604.18486}, 2026.

\bibitem[You et~al.(2026)You, Fan, Lyu, et~al.]{l2d}
Zihan You, Zehua Fan, Wenqi Lyu, et~al.
\newblock {Language2Drive}: Do vision-language-action models really follow the
  instructions in autonomous driving?
\newblock \emph{arXiv preprint arXiv:2610.16634}, 2026.

\bibitem[de~Haan et~al.(2019)de~Haan, Jayaraman, and Levine]{dehaan2019}
Pim de Haan, Dinesh Jayaraman, and Sergey Levine.
\newblock Causal confusion in imitation learning.
\newblock In \emph{NeurIPS}, 2019.

\bibitem[Wen et~al.(2020)Wen, Lin, Darrell, Jayaraman, and Gao]{wen2020copycat}
Chuan Wen, Jierui Lin, Trevor Darrell, Dinesh Jayaraman, and Yang Gao.
\newblock Fighting copycat agents in behavioral cloning from observation
  histories.
\newblock In \emph{NeurIPS}, 2020.

\bibitem[Kuznetsova et~al.(2020)Kuznetsova, Rom, Alldrin, et~al.]{openimages}
Alina Kuznetsova, Hassan Rom, Neil Alldrin, et~al.
\newblock {The Open Images Dataset V4: Unified image classification, object
  detection, and visual relationship detection at scale}.
\newblock \emph{IJCV}, 128\penalty0 (7):\penalty0 1956--1981, 2020.

\bibitem[Yang et~al.(2019)Yang, Russakovsky, and Deng]{spatialsense}
Kaiyu Yang, Olga Russakovsky, and Jia Deng.
\newblock {SpatialSense: An Adversarially Crowdsourced Benchmark for Spatial
  Relation Recognition}.
\newblock In \emph{ICCV}, 2019.

\bibitem[Caesar et~al.(2020)Caesar, Bankiti, Lang, et~al.]{nuscenes}
Holger Caesar, Varun Bankiti, Alex~H. Lang, et~al.
\newblock {nuScenes: A multimodal dataset for autonomous driving}.
\newblock In \emph{CVPR}, 2020.

\bibitem[Talmor et~al.(2019)Talmor, Herzig, Lourie, and Berant]{csqa}
Alon Talmor, Jonathan Herzig, Nicholas Lourie, and Jonathan Berant.
\newblock {CommonsenseQA: A Question Answering Challenge Targeting Commonsense
  Knowledge}.
\newblock In \emph{NAACL-HLT}, 2019.

\bibitem[Sap et~al.(2019)Sap, Rashkin, Chen, {Le Bras}, and Choi]{siqa}
Maarten Sap, Hannah Rashkin, Derek Chen, Ronan {Le Bras}, and Yejin Choi.
\newblock {Social IQa: Commonsense Reasoning about Social Interactions}.
\newblock In \emph{EMNLP-IJCNLP}, 2019.

\bibitem[Huang et~al.(2019)Huang, {Le Bras}, Bhagavatula, and Choi]{cosmosqa}
Lifu Huang, Ronan {Le Bras}, Chandra Bhagavatula, and Yejin Choi.
\newblock {Cosmos QA: Machine Reading Comprehension with Contextual Commonsense
  Reasoning}.
\newblock In \emph{EMNLP-IJCNLP}, 2019.

\bibitem[Bisk et~al.(2020)Bisk, Zellers, Bras, Gao, and Choi]{piqa}
Yonatan Bisk, Rowan Zellers, Ronan~Le Bras, Jianfeng Gao, and Yejin Choi.
\newblock {PIQA: Reasoning about Physical Commonsense in Natural Language}.
\newblock In \emph{AAAI}, 2020.

\bibitem[Pal et~al.(2022)Pal, Umapathi, and Sankarasubbu]{medmcqa}
Ankit Pal, Logesh~Kumar Umapathi, and Malaikannan Sankarasubbu.
\newblock {MedMCQA: A Large-scale Multi-Subject Multi-Choice Dataset for
  Medical domain Question Answering}.
\newblock In \emph{CHIL}, 2022.

\bibitem[Jin et~al.(2021)Jin, Pan, Oufattole, Weng, Fang, and Szolovits]{medqa}
Di Jin, Eileen Pan, Nassim Oufattole, Wei-Hung Weng, Hanyi Fang, and Peter
  Szolovits.
\newblock {What Disease Does This Patient Have? A Large-Scale Open Domain
  Question Answering Dataset from Medical Exams}.
\newblock \emph{Applied Sciences}, 11\penalty0 (14):\penalty0 6421, 2021.

\bibitem[Khot et~al.(2020)Khot, Clark, Guerquin, Jansen, and Sabharwal]{qasc}
Tushar Khot, Peter Clark, Michal Guerquin, Peter Jansen, and Ashish Sabharwal.
\newblock {QASC: A Dataset for Question Answering via Sentence Composition}.
\newblock In \emph{AAAI}, 2020.

\bibitem[Mihaylov et~al.(2018)Mihaylov, Clark, Khot, and Sabharwal]{openbookqa}
Todor Mihaylov, Peter Clark, Tushar Khot, and Ashish Sabharwal.
\newblock {Can a Suit of Armor Conduct Electricity? A New Dataset for Open Book
  Question Answering}.
\newblock In \emph{EMNLP}, 2018.

\bibitem[Ling et~al.(2017)Ling, Yogatama, Dyer, and Blunsom]{aqua}
Wang Ling, Dani Yogatama, Chris Dyer, and Phil Blunsom.
\newblock {Program Induction by Rationale Generation: Learning to Solve and
  Explain Algebraic Word Problems}.
\newblock In \emph{ACL}, 2017.

\bibitem[Liu et~al.(2022)Liu, Swayamdipta, Smith, and Choi]{wanli}
Alisa Liu, Swabha Swayamdipta, Noah~A. Smith, and Yejin Choi.
\newblock {WANLI: Worker and AI Collaboration for Natural Language Inference
  Dataset Creation}.
\newblock In \emph{Findings of EMNLP}, 2022.

\bibitem[Bhagavatula et~al.(2020)Bhagavatula, Bras, Malaviya, et~al.]{anli}
Chandra Bhagavatula, Ronan~Le Bras, Chaitanya Malaviya, et~al.
\newblock {Abductive Commonsense Reasoning}.
\newblock In \emph{ICLR}, 2020.

\bibitem[Geva et~al.(2021)Geva, Khashabi, Segal, Khot, Roth, and
  Berant]{strategyqa}
Mor Geva, Daniel Khashabi, Elad Segal, Tushar Khot, Dan Roth, and Jonathan
  Berant.
\newblock {Did Aristotle Use a Laptop? A Question Answering Benchmark with
  Implicit Reasoning Strategies}.
\newblock \emph{TACL}, 9:\penalty0 346--361, 2021.

\bibitem[Clark et~al.(2020)Clark, Tafjord, and Richardson]{ruletaker}
Peter Clark, Oyvind Tafjord, and Kyle Richardson.
\newblock {Transformers as Soft Reasoners over Language}.
\newblock In \emph{IJCAI}, 2020.

\bibitem[Caesar et~al.(2021)Caesar, Kabzan, Tan, et~al.]{nuplan}
Holger Caesar, Juraj Kabzan, Kok~Seang Tan, et~al.
\newblock {NuPlan: A closed-loop ML-based planning benchmark for autonomous
  vehicles}.
\newblock \emph{arXiv preprint arXiv:2106.11810}, 2021.

\bibitem[Li et~al.(2020)Li, Wang, and Ramanan]{streaming}
Mengtian Li, Yu-Xiong Wang, and Deva Ramanan.
\newblock Towards streaming perception.
\newblock In \emph{ECCV}, 2020.

\bibitem[Woods et~al.(2015)Woods, Wyma, Yund, Herron, and Reed]{woods2015srt}
David~L. Woods, John~M. Wyma, E.~William Yund, Timothy~J. Herron, and Bruce
  Reed.
\newblock Factors influencing the latency of simple reaction time.
\newblock \emph{Frontiers in Human Neuroscience}, 9:\penalty0 131, 2015.

\end{thebibliography}

\clearpage
\edef\sgsupppage{\the\value{page}}
\clearpage
\appendix
\raggedbottom   %
\clubpenalties 3 10000 10000 0 \widowpenalties 3 10000 10000 0   %
\setcounter{page}{1}
\maketitlesupplementary

\renewcommand{\dblfloatpagefraction}{0.8}   %
\renewcommand{\dbltopfraction}{0.85}\setcounter{dbltopnumber}{2}   %
\makeatletter\setlength{\@dblfptop}{0pt}\setlength{\@dblfpsep}{18pt}\setlength{\@dblfpbot}{0pt plus 1fil}\makeatother   %
\section{Training data}
\label{sec:supp_data}
\label{sec:supp_stages}
This section lists the training data of the three stages: the multiple-choice questions of stages~1 and~2 (\cref{tab:supp_qa}) and the rows of stage~3 (\cref{tab:supp_pools,tab:supp_prompt,tab:supp_inputs}), with how each source is built and what is kept out of training.

\begin{figure*}[t]
\centering
\includegraphics[width=\linewidth]{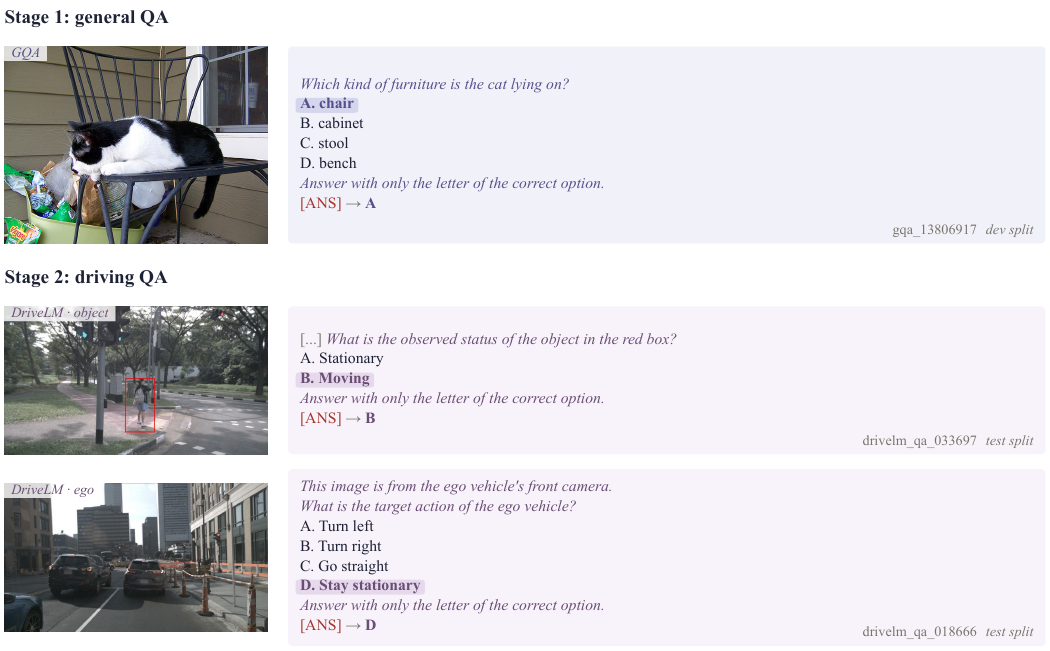}
\caption{\textbf{Held-out question samples of stages~1 and~2.}
Each prompt is shown beside its image as the model reads it, verbatim apart from the cut marked \texttt{[...]}, with the gold option highlighted and its letter as the target at \texttt{[ANS]}; the grey tag gives each question's ID and split.}
\label{fig:supp_samples}
\end{figure*}

\paragraph{Answer readout.}
Every question of the first two stages is a multiple-choice question with options labelled A, B and so on.
The prompt lists the options and ends with \tq{Answer with only the letter of the correct option.}
The answer distribution is the softmax over the logits of the option letters predicted at \texttt{[ANS]}, the last prompt position, so one forward pass gives the whole distribution; yes/no questions are two-option questions.
\Cref{fig:supp_samples} shows questions of both stages as the model reads them, and \cref{fig:data} a stage-3 row.

\begin{table*}[tp]
\centering
\footnotesize
\renewcommand{\arraystretch}{1.15}
\setlength{\tabcolsep}{4pt}
\begin{tabular}{@{}>{\raggedright\arraybackslash}p{78pt}>{\raggedright\arraybackslash}p{168pt}>{\raggedright\arraybackslash}p{140pt}rr@{}}
\toprule
Source & Questions about & Options & Questions & Drawn (\%) \\
\midrule
\grouprow\multicolumn{5}{@{}l}{\textit{Stage~1: \val{s1.pool.g} questions, \val{s1.draws} drawn}} \\
GQA~\cite{gqa} & Relations, attributes and objects (yes/no); colour, material, category & Yes/no, or the gold and 3 others of its kind & \val{s1.n.gqa} & 19.91 \\ %
\quad flipped & Spatial relations, turned into their opposite & Yes/no; the gold is \tq{no} & \val{s1.flips} & \\ %
SAT~\cite{sat} & Spatial relations, counts, action consequences, goals, perspective & SAT's own: 2 for relations, 5 for counts & \val{s1.n.sat} & 19.91 \\ %
TallyQA~\cite{tallyqa} & Counts & 5 consecutive integers around the gold & \val{s1.n.tally} & 16.59 \\ %
ChartQA~\cite{chartqa} & Chart values; yes/no & 5 values or 5 years around the gold & \val{s1.n.chart} & 5.31 \\ %
AI2D~\cite{ai2d} & Science diagrams & The dataset's & \val{s1.n.aitwod} & 3.98 \\ %
TextVQA~\cite{textvqa} & Scene text: yes/no and counts & Yes/no; counts as for TallyQA & \val{s1.n.textvqa} & 0.66 \\ %
OpenImages~\cite{openimages} & Which of two drawn boxes shows a named object & The true box and a jittered one or a box on another object &\val{s1.box} & 3.65 \\ %
\quad directly above & Whether one object is directly above another in the real world &Yes for stacked pairs, such as a cup on a saucer; no for offset or swapped pairs & \val{s1.above} & 1.99 \\ %
SpatialSense~\cite{spatialsense} & \tq{Is the \textit{subject} \textit{predicate} the \textit{object}?} & Yes/no & \val{s1.ss} & 1.99 \\ %
Text, 13 datasets & Commonsense, science, medicine, mathematics, inference and rules, with a blank image & The dataset's: 2 to 8, or yes/no & 309,553 & 19.68 \\ %
\quad answer checks & \tq{Proposed answer: \ldots{} Is the proposed answer correct?} & Yes/no; the proposed answer is the gold in half & 39,999 & 4.32 \\ %
\quad extra option & \tq{None of the other options is correct} & Replaces the gold (half) or joins the options (half) & \val{s1.hedge} & 1.99 \\ %
\midrule
\grouprow\multicolumn{5}{@{}l}{\textit{Stage~2: \val{s2.pool} questions, 192,000 drawn}} \\
DriveLM~\cite{drivelm,nuscenes} objects & An object in a red box on the camera that shows it: motion, future state, appearance, relevance to the ego vehicle, type & 2 to 5, one answer set per kind; appearance: the gold and 3 descriptions that share no word with it &86,373 & \val{s2.mix.obj}\phantom{.00} \\ %
\quad ego vehicle & Front camera: target action, steering, speed, signals, traffic elements & 2 to 5, one answer set per kind & 15,680 & \val{s2.mix.ego}\phantom{.00} \\ %
\quad relations & Two key objects: left or right, closer or farther & Yes/no, each with a word-flipped twin & 11,352 & \val{s2.mix.rel}\phantom{.00} \\ %
\quad box choice & Which of two drawn boxes shows a key object & The true box and a jittered one &10,973 & \val{s2.mix.box}\phantom{.00} \\ %
Stage-1 questions & Stage~1 minus directly-above and extra-option rows & As in stage~1 & \val{s1.pool.f} & \val{s2.mix.sone}\phantom{.00} \\ %
\bottomrule
\end{tabular}
\caption{\textbf{Question sources of stages~1 and~2.}
Questions: training questions in the pool; drawn: share of the draws in training.
Indented rows are built from the source above, and GQA's flipped questions count toward its share.
Text: CSQA~\cite{csqa}, SIQA~\cite{siqa}, CosmosQA~\cite{cosmosqa}, PIQA~\cite{piqa}, MedMCQA~\cite{medmcqa}, MedQA~\cite{medqa}, QASC~\cite{qasc}, OpenBookQA~\cite{openbookqa}, AQuA~\cite{aqua}, WANLI~\cite{wanli}, $\alpha$NLI~\cite{anli}, StrategyQA~\cite{strategyqa} and RuleTaker~\cite{ruletaker}, at most 40,000 questions each.}
\label{tab:supp_qa}
\end{table*}

\paragraph{Question pools.}
\Cref{tab:supp_qa} lists the sources of stages~1 and~2; stage~2 converts only DriveLM's closed-form questions.
The sources carry answer cues: SAT lists the gold first in all \val{s1.satfirst} of its questions, and \val{s1.satcnt1} of its \val{s1.satcnt} counting questions (\val{s1.satcnt1pct}\%) are answered 1; the option shuffling and the tempered answer sampling below counter both.
Count options are five consecutive integers that start zero to four below the gold, never below zero, so the gold is not always the middle value.
Chart values are spaced by the larger of the gold's precision and \val{s1.chartgap}\% of the gold, and a jittered box overlaps the true one at IoU \val{s1.boxiou}.
The extra option, an answer that rejects every listed option, goes to text questions with three or more options, except WANLI's, whose three labels cover every case.
Only single-image questions are kept.
Options are shuffled once with a fixed seed, so questions whose options name other options, such as \tq{all of the above}, are dropped.
Within a source, a draw picks a question type with probability proportional to $n^{\val{s1.taskexp}}$ and then an answer with probability proportional to $n^{\val{s1.ansexp}}$, where $n$ is the number of questions of that type or with that answer.
Images with identical bytes or the same Visual Genome or COCO image never cross the training and development splits.
The development splits hold \val{s1.dev} questions of SAT, GQA, TallyQA, AI2D, ChartQA and TextVQA and \val{s2.devper} DriveLM questions of each kind; DriveLM is split by scene into \val{s2.split.tr} training, \val{s2.split.dev} development and \val{s2.split.te} test scenes.

\paragraph{Decontamination.}
We drop \val{s1.rmdup} duplicate questions, \val{s1.rmmulti} questions with more than one correct option, and \val{s1.rmext} questions on an image of SAT's evaluation splits~\cite{sat}, BLINK~\cite{blink}, CV-Bench~\cite{cambrian} or 3DSRBench~\cite{threedsr}, matched by identical bytes or the same Visual Genome or COCO image.
Images added for OpenImages, SpatialSense and DriveLM questions are also dropped within Hamming distance \val{s1.phash} of an evaluation image under a 63-bit DCT hash, among them \val{s2.ph.qa} DriveLM images of the question kinds and \val{s2.ph.box} of the box and relation kinds.
Text questions that share at least \val{s1.ngramfrac}\% of their \val{s1.ngram}-word shingles (runs of \val{s1.ngram} consecutive words) with the questions of the language benchmarks of \cref{tab:vqa} or ARC-Easy~\cite{arc} are dropped as well.

\begin{table*}[tp]
\centering
\footnotesize
\renewcommand{\arraystretch}{1.15}
\setlength{\tabcolsep}{4pt}
\begin{tabular}{@{}>{\raggedright\arraybackslash}p{74pt}rr>{\raggedright\arraybackslash}p{160pt}>{\raggedright\arraybackslash}p{140pt}@{}}
\toprule
Source & Rows & Drawn (\%) & Target $q$ or reward $r$ & Rows from \\
\midrule
\grouprow\multicolumn{5}{@{}l}{\textit{Supervised: \val{r.d.sft} rows, \val{r.t.sftsteps} steps of \val{r.t.batch} rows, \val{r.k.total} candidates}} \\
NAVSIM scenes & \val{r.k.fit} & 12.17 & $q\propto\exp\big((\mathrm{PDMS}\cdot\mathrm{DDC}-\val{r.t.adew}\,\mathrm{ADE})/\val{r.t.tnav}\big)$ & One per navtrain scene of 1,083 logs \\ %
NAVSIM frames & 44,243 & 5.83 & $q\propto\exp(-\mathrm{ADE}/1\,\mathrm{m})$ & Other frames of 1,076 of these logs that pass the navtrain filter \\ %
PAI-AV logged & 1,044,771 & 36\phantom{.00} & $q\propto\exp(-\mathrm{ADE}/\val{r.t.tpai}\,\mathrm{m})$ & 149,253 training clips, every 2~s \\ %
PAI-AV rollouts & \val{r.d.dag} & \val{r.d.dagw}\phantom{0} & As above, with the ADE to a recovery trajectory & AlpaSim rollouts of earlier models in \val{r.d.dagsc} scenes \\ %
WOD-E2E & 406,205 & 18\phantom{.00} & $q\propto\exp(-\mathrm{ADE}/\val{r.t.twod}\,\mathrm{m})$ & Every 10-Hz frame of 1,991 training segments \\ %
Instructions & 301,158 & 10\phantom{.00} & $q\propto\exp(-d/t)$, $t$ of the source, over the candidates that follow the instruction (see text) &100,386 scenes of the three sources, three rows each \\ %
DriveLM & \val{s2.rows} & 9.45 & Stage-2 objective, KL \val{r.t.kl} to the stage-2 model & All training questions of stage~2 \\ %
Other questions & 132,012 & 4.05 & As for DriveLM & 22 stage-1 sources, among them SAT, GQA and TallyQA \\ %
\midrule
\grouprow\multicolumn{5}{@{}l}{\textit{Last phase: 1,846,944 rows, \val{r.t.ftsteps} steps of \val{r.t.batch} rows, \val{r.k.total} candidates}} \\
WOD-E2E, rated range & \val{r.d.wodft} & \val{r.t.ftwod}\phantom{.00} & $q\propto\exp\big((\tilde R-\mathrm{ADE}/2)/\val{r.t.tft}\big)$ & Frames 135--165, around the rated frames 147--150, of 1,987 training segments, in place of the WOD-E2E rows \\ %
Other rows & 1,785,484 & 60\phantom{.00} & Their supervised targets & The other supervised rows, each weight $\times\,0.7317$ \\ %
\midrule
\grouprow\multicolumn{5}{@{}l}{\textit{RL: \val{r.d.rl} rows, \val{r.t.rlsteps} steps of \val{r.t.rlbatch} rows, \val{r.k.rl} candidates}} \\
NAVSIM scenes & \val{r.d.rlnav} & 50\phantom{.00} & $r=\mathrm{PDMS}\cdot\mathrm{DDC}$, minus \val{r.t.pen} if NC, DAC or TTC is below 1 & The navtrain scenes above \\ %
PAI-AV logged & \val{r.d.rlpai} & 45\phantom{.00} & $r=-0.2\,\mathrm{ADE}$ & Up to two per training clip \\ %
WOD-E2E rated & \val{r.d.rlwod} & 5\phantom{.00} & $r=\mathrm{RFS}/10$ & The rated validation frames \\ %
\bottomrule
\end{tabular}
\caption{\textbf{Training pools of stage~3}, by phase.
Drawn: share of the rows drawn in the phase; ADE: metres to the logged future over the eight poses at 0.5--4~s; $\tilde R$: the RFS with the logged future as the only rated trajectory.
WOD-E2E rates one moment of each segment, around frames 147--150, so the last phase trains on WOD-E2E frames around that moment, scored with $\tilde R$.
None of the phases trains on the 10,949 scenes of NAVSIM's 109 held-out training logs.}
\label{tab:supp_pools}
\end{table*}

\begin{table*}[tp]
\centering
\footnotesize
\setlength{\fboxsep}{7pt}
\colorbox{jdmain!5}{\parbox{0.955\linewidth}{\raggedright\frenchspacing
\texttt{You are driving the ego vehicle; the image is its front camera. Speed \textit{v} m/s, longitudinal acceleration \textit{a} {\fontencoding{T1}\selectfont m/s\textasciicircum 2}, navigation command: \textit{c}. Past positions over the last 1.5 s relative to now (x~ahead, y~left, metres): (\textit{x}, \textit{y}), (\textit{x}, \textit{y}), (\textit{x}, \textit{y}). Choose the trajectory for the next 4 seconds.}\\[5pt]
\textit{In one NAVSIM training row:} \textit{v} = \texttt{10.5}, \textit{a} = \texttt{+0.3}, \textit{c} = \texttt{go straight}, and the past positions \texttt{(-14.6,~-0.1), (-10.0,~-0.1), (-5.1,~-0.0)}.\\[3pt]
\textit{With past frames:} \texttt{the image is its front camera} becomes \texttt{the images are its front camera 1.5 s, 1.0 s and 0.5 s ago (frames 0-2) and now (frame 3)}, and \texttt{frame: \textit{k}} precedes each image, oldest first.\\[3pt]
\textit{Rows with an instruction:} the command is \texttt{go straight}, and the prompt ends with \texttt{Driving instruction:} and the instruction, such as \texttt{Bring the car to a stop and stay stopped.}}}
\caption{\textbf{Stage-3 prompt}, the same for every driving source.
The chat template runs with thinking off, and \texttt{[ANS]} is the last position, after its empty thinking block.}
\label{tab:supp_prompt}
\end{table*}

\begin{table*}[tp]
\centering
\footnotesize
\renewcommand{\arraystretch}{1.15}
\setlength{\tabcolsep}{4pt}
\def\sgnone{\textcolor{black!45}{None}}   %
\begin{tabular}{@{}>{\raggedright\arraybackslash}p{64pt}>{\raggedright\arraybackslash}p{134pt}>{\raggedright\arraybackslash}p{134pt}>{\raggedright\arraybackslash}p{134pt}@{}}
\toprule
& NAVSIM & PAI-AV & WOD-E2E \\
\midrule
\grouprow\multicolumn{4}{@{}l@{}}{\textit{Images}} \\
Camera & nuPlan's~\cite{nuplan} front camera, CAM\_F0, as logged & The 120$^\circ$ f-theta front-wide camera, re-projected into CAM\_F0 & FRONT, FRONT\_LEFT and FRONT\_RIGHT, re-projected into CAM\_F0 by rotation, each pixel from the camera with the nearest optical axis \\ %
Current frame & \multicolumn{3}{>{\raggedright\arraybackslash}p{418pt}@{}}{1920$\times$1080 with CAM\_F0's intrinsics and distortion; re-projected frames are saved at JPEG quality 92. The processor resizes the frame to 1024$\times$576, its limit of \val{r.in.maxpix} pixels, which gives 576 visual tokens.} \\ %
Past frames & 1.5, 1.0 and 0.5~s earlier, from the 2-Hz log & The same times, nearest frame within 0.1~s; clips with an even hash, about half & \sgnone \\ %
\quad\textcolor{black!55}{\textit{all sources}} & \multicolumn{3}{>{\raggedright\arraybackslash}p{418pt}@{}}{556$\times$313 (at most 174,080 pixels), 544$\times$320 in the processor, 170 visual tokens each. Rows without all three, rollout rows and question rows stay single-frame; 237,246 of the \val{r.d.rl} RL rows have past frames.} \\ %
\grouprow\multicolumn{4}{@{}l@{}}{\textit{Ego state and command}} \\
Speed and acceleration & Current ego status & Egomotion at about 10~Hz, central differences over $\pm$0.2~s & Last past state, at 4~Hz \\ %
Past positions & Logged poses at 2~Hz & Egomotion at $-1.5$, $-1.0$ and $-0.5$~s & Past states at 4~Hz \\ %
Input dropout & \sgnone & Training rows with an odd token hash: acceleration 0 and past positions at constant speed along the path, so the plan cannot just repeat the past speed changes; evaluation keeps the real values & \sgnone \\ %
Command & The devkit's: left, straight or right; frames with \tq{unknown} dropped & From the heading 4~s ahead: left above 20$^\circ$, right below $-20^\circ$, else straight & The frame's intent: straight, left, right or unknown \\ %
\grouprow\multicolumn{4}{@{}l@{}}{\textit{Targets}} \\
Logged future & Poses at 2~Hz up to 4~s & Egomotion at 0.5, 1.0, \ldots, 4.0~s & Every second 4-Hz state up to 4~s \\ %
Candidate scores & NAVSIM v1.1 PDM scorer: NC, DAC, EP, TTC, Comf., DDC and PDMS & ADE to the logged future, or to the recovery trajectory in rollout rows & ADE; $\tilde R$ in the last phase; for RL, RFS on the 479 rated frames \\ %
\grouprow\multicolumn{4}{@{}l@{}}{\textit{Rows}} \\
Sampling & One row per navtrain scene (4 past and 10 future frames, a route), and other frames of the same logs that pass this filter & Every 2~s from 1.9~s after a clip's start to 4.4~s before its end; RL: up to 2 per clip &Every 10-Hz frame; 46 segments held out by hash \\ %
Excluded & navtest; the 109 held-out logs & Logged and instruction rows: clips outside the training split, the 916 AlpaSim scenes, Language2Drive's AlpaSim scenes and our evaluation clips; rollout rows: only test-split and evaluation clips &Validation and test segments, except the 479 rated frames for RL \\ %
\bottomrule
\end{tabular}
\caption{\textbf{Driving inputs by source.}
All sources share the prompt of \cref{tab:supp_prompt} and the camera model of CAM\_F0; positions are in the current ego frame, speed is $|v|$ and acceleration longitudinal.}
\label{tab:supp_inputs}
\end{table*}

\paragraph{Driving pools.}
\Cref{tab:supp_pools} lists the rows of each phase of stage~3, \cref{tab:supp_prompt} their prompt and \cref{tab:supp_inputs} how each source becomes a row.
No training pool holds a frame or clip of any evaluation set; the last row of \cref{tab:supp_inputs} lists what each source leaves out.
Rollout rows teach recovery: their frames follow AlpaSim rollouts of earlier models in \val{r.d.dagsc} NuRec scenes, 327 from PAI-AV's training split and 165 from its validation split, none among the 916 AlpaSim scenarios.
The recovery trajectory of a rollout row returns to the human path in about 3~s, never ahead of the human, and slows behind a lead vehicle.
Each instruction scene gives three rows on the same frame: one without an instruction, one with an instruction that the logged future follows and one with an instruction that it does not follow.
Each row's target is $q\propto\exp(-d/t)$, with $t=1$~m on NAVSIM and WOD-E2E and 0.5~m on PAI-AV.
Without an instruction, it spreads over all candidates with $d$ the ADE; with an instruction that the logged future follows, over the candidates that follow it; with one that it does not follow, over the feasible candidates that follow it (on NAVSIM, no collision, inside the drivable area and lane changes only into a lane of the same direction; on PAI-AV, no overlap with an obstacle box from 0.5 to 4~s; on every source, acceleration at most 3.5~m/s$^2$, also from the current speed, and yaw rate at most 0.8~rad/s), with $d$ the distance to the logged future in the direction the instruction leaves free: along the path for lane and turn instructions, across it for all other kinds.
Instructions come from 8 templates for each of 12 kinds, and the last two of each kind are used only in development.
To keep Language2Drive's wordings out of training, we wrote the templates ourselves: none comes within a string similarity of 0.65 of any of Language2Drive's 904 instruction strings, and no instruction row comes from a Language2Drive log.
A development set of 1,000 rows, 200 each from held-out NAVSIM logs, rollout rows, WOD-E2E, held-out PAI-AV training clips and instruction rows, tracks the loss every 1,000 steps; we keep the last checkpoint.
Of its rows, 65 come from 40 of the 916 AlpaSim scenarios.

\section{Training details}
\label{sec:supp_train}
\label{sec:supp_stage3}
\Cref{tab:supp_hparams} lists the settings of every stage; the paragraphs below give the objective of stages~1 and~2 and how the Policy Head grows before RL.

\begin{table*}[tp]
\centering
\footnotesize
\renewcommand{\arraystretch}{1.15}
\setlength{\tabcolsep}{4pt}
\begin{tabular}{@{}>{\raggedright\arraybackslash}p{60pt}*{5}{>{\raggedright\arraybackslash}p{78pt}}@{}}
\toprule
& Stage~1 & Stage~2 & Stage~3, supervised & Last phase & RL \\
\midrule
Starts from & \nineb, new LoRA & Stage-1 model, merged; new LoRA & Stage-2 model, merged; new LoRA; Policy Head at zero & The supervised model & The last-phase model; Policy Head refit to \val{r.k.rl} rows \\ %
Trainable & \val{train.pnine} & \val{train.pnine} & \val{train.pnine} + 20,980,737 & \val{train.pnine} + 20,980,737 & \val{train.pnine} + 21,603,481 \\ %
LoRA & \multicolumn{5}{>{\raggedright\arraybackslash}p{422pt}@{}}{Rank \val{train.rank}, $\alpha$ \val{train.alpha}, dropout 0, on 200 projections: q, k, v and o of the 8 full-attention layers; in\_proj\_qkv, in\_proj\_z and out\_proj of the 24 Gated DeltaNet layers; gate, up and down of all 32 MLPs. The vision tower, the embeddings and the language-model head stay frozen.} \\ %
Objective & Paired Brier, \val{train.draws} draws per option order & As stage~1 & Cross-entropy to $q$; questions as in stage~2 & As stage~3 & Exact expected reward \\ %
KL term & \val{train.kl} to \nineb & \val{train.kl} to the stage-1 model & \val{r.t.kl} to the stage-2 model, on question rows & As stage~3 & $\beta=\val{r.t.beta}$ to $p_{\mathrm{SFT}}$ \\ %
Batch & \val{train.batch} & \val{train.batch} & \val{r.t.batch} & \val{r.t.batch} & \val{r.t.rlbatch} \\ %
GPUs $\times$ micro $\times$ accum. & 8 $\times$ 4 $\times$ 2 & 8 $\times$ 4 $\times$ 2 & 8 $\times$ 4 $\times$ 4 & 8 $\times$ 4 $\times$ 4 & 8 $\times$ 4 $\times$ 2 \\ %
Steps & \val{s1.steps} & \val{s2.steps} & \val{r.t.sftsteps} & \val{r.t.ftsteps} & \val{r.t.rlsteps} \\ %
Rows drawn & \val{s1.draws} & 192,000 & 2,675,200 & 320,000 & 128,000 \\ %
Learning rate & \val{train.lr} & \val{train.lr} & \val{r.t.lr} & \val{r.t.ftlr} & \val{r.t.rllr} \\ %
Warm-up & \val{train.warm}\% & \val{train.warm}\% & \val{r.t.warm}\% & 5\% & 5\% \\ %
Weight decay & \val{train.wd} & \val{train.wd} & \val{r.t.wd} & 0.01 & 0.01 \\ %
Optimizer & \multicolumn{5}{>{\raggedright\arraybackslash}p{422pt}@{}}{AdamW (0.9, 0.999, $\epsilon=10^{-8}$) with cosine decay after a linear warm-up, gradient norm clipped at 1.0, bfloat16} \\ %
Image pixels & 65,536--262,144 & 65,536--262,144 & \multicolumn{3}{>{\raggedright\arraybackslash}p{250pt}@{}}{Current frame 65,536--\val{r.in.maxpix}; past frames at most 174,080} \\ %
Seed & 20260918 & 20260926 & 20261022 & 20261022 & 20261023 \\ %
Kept checkpoint & Best on development (the last step) & Best on development, scored every 1,000 steps & Last & Last & Last \\ %
GPU hours & 31.6 (H100) & 11.6 (H100) & \val{r.t.gpuh} (H200) & 16.7 (H200) & 7.7 (H200) \\ %
\bottomrule
\end{tabular}
\caption{\textbf{Training settings.} Every stage trains a LoRA of the same shape on the language model, and stage~3 adds the Policy Head.
The last phase continues the supervised model of stage~3, and RL the last-phase model, without merging.
Batches and draws count questions in stages~1 and~2 and rows in stage~3.
Candidate rewards take about 1,600 single-thread CPU process-hours, and rendering the frames of one AlpaSim rollout for the rollout rows takes about 1.1 H200 GPU minutes.}
\label{tab:supp_hparams}
\end{table*}

\paragraph{Objective.}
Stages~1 and~2 train on each question in its stored option order and a shifted one, the gold letter moving with its content.
The objective is a paired Brier estimator over draws of the answer distribution, which counts a pair of draws from the two orders as correct only if both are gold, plus a KL term to the model before the stage (\cref{sec:stages}).
With $p$, $p'$, $y$ and $y'$ as in \cref{sec:stages}, the reward is twice the share of pairs that hit minus how often two pairs agree.
Its expectation, $2p_yp'_{y'}-\sum_c p_c^2\sum_{c'}p'^2_{c'}$, is one minus the Brier score of $p\otimes p'$~\cite{brier1950}.
The gradient from \val{train.draws} pairs is unbiased: over two to nine options, the mean of 48 such estimates has a cosine of at least 0.997 with the exact gradient, and one estimate a median cosine of 0.99 at two options, 0.86 at nine.   %

\paragraph{Runs before the correction.}
The ablation runs, in the main text and here, use this objective as implemented before we corrected it, except our supervised model without instruction rows (\cref{tab:l2d}); the models of the main result tables, the \method rows of \cref{tab:supp_alpasim} and its last row were retrained from stage~1 after the correction.

\paragraph{Head refit.}
Before RL, the Policy Head grows from \val{r.k.total} to \val{r.k.rl} rows, adding one for each candidate built by rule (\cref{sec:supp_vocab}): the old rows are kept, and each new row starts as the row of its nearest old candidate with its bias lowered by 4, which leaves the old logits and the most probable candidate unchanged.
Only the new rows are then fit, by L-BFGS on frozen features of NAVSIM and PAI-AV training rows, with an L2 pull of weight $10^{-3}$ toward their start.
Computing $p_{\mathrm{SFT}}$ over the RL pool takes another 11.4 H200 GPU hours, and extracting the features for the refit 6.6.

\section{Candidate trajectories}
\label{sec:supp_vocab}
The candidate set combines k-means centres of NAVSIM's human trajectories (\cref{tab:supp_cover}), one stationary trajectory, k-means centres of PAI-AV's high-speed driving and, for RL, \val{r.ab.v2.n} candidates built by rule for braking and curves.

\begin{table}[h]
\centering
\scriptsize
\setlength{\tabcolsep}{2.2pt}
\begin{tabular}{lcccc}
\toprule
Candidate set & $K$ & Spacing (m) & Best PDMS$\uparrow$ & Nearest PDMS$\uparrow$ \\
\midrule
k-means 256 + stationary & 257 & 0.98 & \heatv{0}{99.21} & \heatv{0}{77.88} \\
k-means 1,024 + stationary & 1,025 & 0.61 & \heatv{61}{99.66} & \heatv{56}{82.76} \\
\ours k-means 4,096 + stationary & 4,097 & 0.40 & \heatv{100}{\val{r.k.bestho}} & \heatv{100}{\val{r.k.nearnav}} \\
\bottomrule
\end{tabular}
\caption{\textbf{Coverage of NAVSIM scenes by k-means sets of human trajectories.} On \val{r.k.nho} of the 10,949 held-out training scenes, the columns give the mean distance between neighbouring candidates and the mean PDMS of the best candidate and of the candidate nearest to the human trajectory. More centres mainly lift the nearest candidate; the best one is near the PDMS ceiling at every size. Shading deepens from worst to best in each PDMS column.}
\label{tab:supp_cover}
\end{table}

\paragraph{Construction.}
The clustered part runs k-means on the human 4-s trajectories of the \val{r.k.fit} NAVSIM training scenes, each a vector of \val{r.k.poses} positions and headings, and adds one stationary trajectory.
\Cref{tab:supp_cover} shows how the coverage grows with $K$; we use \val{r.k.km} centres.
NAVSIM's training logs contain few scenes above \val{r.k.fastv}~m/s, so k-means on NAVSIM alone leaves few candidates for the highway speeds of PAI-AV.
The \val{r.k.fast} high-speed candidates are the centres of k-means over the \val{r.k.fastrows} PAI-AV training trajectories driven at \val{r.k.fastv}~m/s or faster; their 4-s end points lie 52.5 to 190.5~m ahead.
Every candidate is scored once in every training scene (\cref{sec:reward}).

\paragraph{Braking and curve extension.}
The set has no plan that brakes to a stop from 9~m/s or faster: at 8~m/s, its shortest stop is 11.2~m.
The extension appends \val{r.ab.v2.n} candidates: 84 that brake from every speed between 5 and 14~m/s and from 16, 18, 20 and 22~m/s, 52 of them to a stop within 4~s, and 68 medium-speed curves and ramps, which the clustered set covers poorly on PAI-AV.
Added to the reward-target model with the current frame only (\cref{tab:supp_alpasim}) by fitting only the new rows of its Policy Head, as before RL (\cref{sec:supp_train}), the extension keeps navtest PDMS at \val{r.ab.v2.nav}, lowers the PAI-AV ADE at 3~s from \val{r.ab.v1.pai} to \val{r.ab.v2.pai}~m, and raises the AlpaSim distance per at-fault incident from \val{r.ab.ro.km} to \val{r.ab.v2.km}~km.
The RL rows of the main tables use the extended set, \val{r.k.rl} candidates.
With RL, the extension again lowers the PAI-AV ADE and leaves navtest and the mean scenario score level: in two RL runs from the supervised model before its last phase, one per set, it moves navtest PDMS from \val{r.ab.v1rl.nav} to \val{r.ab.v2rl.nav}, lowers the PAI-AV ADE at 3~s from \val{r.ab.v1rl.pai} to \val{r.ab.v2rl.pai}~m and raises the AlpaSim distance per at-fault incident from \val{r.ab.v1rl.kmf} to \val{r.ab.v2rl.kmf}~km; the mean scenario score changes by \val{r.ab.v2v1.d} (95\% interval \val{r.ab.v2v1.ci}).

\section{Benchmark protocols}
\label{sec:supp_protocols}
\Cref{tab:supp_bench} summarizes the five benchmarks; the paragraphs below give how we read one plan from our model and how each benchmark runs and scores it.
On every benchmark, the front image follows the camera model of nuPlan's CAM\_F0 (\cref{tab:supp_inputs}), re-projected where needed.
Unless noted otherwise, other methods' results are as reported in their papers, and TransFuser's by NAVSIM~\cite{navsim}.

\begin{table*}[tp]
\centering
\footnotesize
\renewcommand{\arraystretch}{1.15}
\setlength{\tabcolsep}{4pt}
\begin{tabular}{@{}>{\raggedright\arraybackslash}p{68pt}>{\raggedright\arraybackslash}p{96pt}l>{\raggedright\arraybackslash}p{110pt}>{\raggedright\arraybackslash}p{110pt}>{\raggedright\arraybackslash}p{48pt}@{}}
\toprule
Benchmark & What it is & Loop & Split and size & Metrics & Horizon \\
\midrule
NAVSIM~\cite{navsim} v1.1 & Scenes of nuPlan~\cite{nuplan} logs; the PDM scorer simulates each plan & Open & navtest: \val{r.nav.n} scenes from 136 logs & PDMS: NC $\times$ DAC $\times$ a weighted sum of EP, TTC and Comf. & 4~s at 2~Hz \\
WOD-E2E~\cite{wode2e} & Long-tail scenes; RFS scores a plan against rated trajectories & Open & Test split, scored by the official server; the \val{r.d.rlwod} rated validation frames provide RL rewards & RFS; 3- and 5-s ADE & 5~s at 4~Hz \\
PAI-AV~\cite{paiav} & Driving clips with released egomotion & Open & Qwen-Drive-1.0's leakage-free subset: 700 frames, five in each of 140 held-out test clips & 3- and 5-s ADE of one plan and of the best of six & 5~s at 10~Hz \\
AlpaSim~\cite{alpasim} & Simulator that renders the camera along the car's own path in PAI-AV-NuRec~\cite{painurec,threedgut} v26.01 scenes & Closed & 916 scenarios; one whose route check fails counts as failed, with score zero & Collision and off-road rates, progress, km per at-fault incident, pass rate and mean scenario score & Whole scenario, a call every 100~ms \\
Language2Drive~\cite{l2d} & Two instructions per scene; checks whether each plan follows its own (NAVSIM track) & Open & \val{r.l2d.n} instruction pairs in four wordings each & SemSISR, SemCFPS: single instructions and pairs followed; SCS, PCS: graded scores & 4~s \\
\bottomrule
\end{tabular}
\caption{\textbf{The five benchmarks.} Four score each plan once (open loop); in AlpaSim, the car drives its plans (closed loop).
Where a benchmark scores 5~s, our 4-s plans are extended at constant speed.}
\label{tab:supp_bench}
\end{table*}

\paragraph{Plan readout.}
On every benchmark, our plan is the mean of our \val{r.pai.topk} most probable candidates, weighted by their probabilities at temperature \val{r.pai.tau}; the number was fixed in advance and the temperature set on the PAI-AV validation split.
On that split, any number from 32 to all candidates keeps the 3-s and 5-s ADE of our supervised and final models within \val{r.pai.kd3} and \val{r.pai.kd5}~m of those at \val{r.pai.topk}, while the most probable candidate alone raises the 3-s ADE by \val{r.pai.k1s} and \val{r.pai.k1r}~m.
A temperature of \val{r.pai.tau} gave the lowest 3-s ADE among 1, 0.5, 0.2, 0.1, 0.05 and 0.02 for each of three earlier checkpoints.

\paragraph{NAVSIM.}
We score navtest with the NAVSIM v1.1 PDM scorer; 12,131 of its \val{r.nav.n} scenes have all past frames, and the others are read from the current frame only.

\paragraph{WOD-E2E.}
We report the test split, scored by the official server, since the \val{r.d.rlwod} rated validation frames, the only ones with ratings for every trajectory, provide the RL rewards.
Where we compute RFS ourselves, on the rated validation frames and for $\tilde R$, we follow the official rater-feedback code.
The input is the current frame only, and our 4-s plans are resampled to the benchmark's 4-Hz, 5-s grid, the last second extended at constant velocity.

\paragraph{PAI-AV.}
The 700 frames are Qwen-Drive-1.0's leakage-free subset: the five timestamps of each of the 140 held-out test clips that it selected.
The logged future comes from our processing of the released egomotion and is the reference for every entry we ran in \cref{tab:pai}.
Our plan's eight poses over 4~s are interpolated to the 10~Hz of the log by a cubic spline that starts at the logged velocity, and extended to 5~s at constant velocity.
One plan is our top-\val{r.pai.topk} mean and Qwen-Drive-1.0's first sample; the best of six is the minimum ADE over our six most probable candidates and over its six samples.
On these 700 frames, Qwen-Drive-1.0's paper reports a mean ADE over six samples of 0.42 and 0.47~m at 3~s for its supervised and RL planners; our runs give \val{r.pai.qdsft.t} and \val{r.pai.qdrl.t}~m for their first sample, whose ADE has the same expectation.

\paragraph{AlpaSim.}
All runs use the 916 scenarios of PAI-AV-NuRec version 26.01, the same simulator configuration and the same scorer, with the ClipGT maps for the off-road check.
Other agents replay their logs without reacting, and actors present for under 3~s are dropped; a collision is at fault when it touches the front or a side of the ego vehicle, never when it touches the rear.
An incident is a collision or an off-road event; a scenario passes if it has no at-fault collision, off-road event or lateral-corridor exit, and its score is progress along the human path divided by 0.8, capped at 1, or zero if it fails.
Mean scores and pass rates are over all 916 scenarios; those whose route fails AlpaSim's sanity check, as it folds back on itself or strays more than 10~m from the logged trajectory, 12 to 14 in our runs and 8 to 12 in Qwen-Drive-1.0's, count as failed, with score zero.
Collision and off-road rates are over the scenarios whose route check passes.
Scored with the XODR maps instead, Qwen-Drive-1.0's supervised and RL planners reach 0.372 and 0.359~km per at-fault incident, against 0.27 and 0.37~km in its paper, on version 26.02.
With a fixed configuration, AlpaSim is deterministic: reruns on other machines, of up to 441 scenarios, matched every metric in every scenario, so each row is one run.

\paragraph{Low-level controller.}
Every plan passes through our low-level controller, the PDM simulator of \cref{sec:select}, which starts a scenario's first plan from the steering angle estimated from the yaw rate; Qwen-Drive-1.0's enters it as the eight of its 50 poses at 0.5, 1.0, \ldots, 4.0~s, and both models read the ego speed and acceleration from AlpaSim's dynamic state.
Without the PDM simulator (\cref{tab:supp_alpasim}), AlpaSim's controller tracks the eight poses of our plan directly.
Our RL model then drives \val{r.as.nox.kmf} against \val{r.as.rl.kmf}~km per at-fault incident and its mean scenario score is \val{r.as.nox.d} lower (95\% interval \val{r.as.nox.ci}).
It leaves the road in \val{r.as.nox.off}\% of the scenarios against \val{r.as.rl.off}\%, while its collision rates change little, so the drop in km per at-fault incident comes mostly from leaving the road.
With plans tracked directly and speed differenced from poses, Qwen-Drive-1.0's supervised and RL planners drive 0.306 and 0.246 against \val{r.as.qdsft.kmf} and \val{r.as.qdrl.kmf}~km per at-fault incident.

\paragraph{Delay.}
The simulator calls the policy every 100~ms of simulated time and stops its clock while the policy computes; the policy returns its full plan.
Following streaming perception~\cite{streaming}, we charge each model's computation time as delay.
For \cref{tab:alpasim} (top), we charge our \val{r.l.ours}~ms and Qwen-Drive-1.0's \val{r.l.qd}~ms as measured and its \val{r.l.qdr}~ms with reasoning as 1.2~s; the simulator rounds a delay up to the next 100~ms, so these plans reach AlpaSim's controller 100, 600 and 1,200~ms after their call.
A late plan keeps the timestamps of its call: the PDM simulator runs it from the ego pose, speed and acceleration at the call and from the steering angle that its run of the previous plan reaches at that time, and AlpaSim's controller anchors it at the ego pose of its first timestamp and tracks it from the current time on, with no prediction of the ego state over the delay.
Until the first plan arrives, AlpaSim's controller follows the logged path.
Over 2,000 resamples of the scenarios, the 95\% intervals of km per at-fault incident are \val{r.ci.rl84} and \val{r.ci.rl1200} for our RL model at 84~ms and 1.2~s and \val{r.ci.qdp518} and \val{r.ci.qdr1200} for Qwen-Drive-1.0 at 518~ms and with reasoning at 1.2~s; without delay, they are \val{r.ci.s1} and \val{r.ci.rl0} for our supervised and RL models and \val{r.ci.qdr0} and \val{r.ci.qdrl0} for Qwen-Drive-1.0's supervised and RL planners.

\paragraph{Fixed lags.}
In \cref{fig:latency} (bottom), every plan arrives a fixed lag after its call, whatever its computation time, and each planner absorbs the lag up to a limit.
With lags of 0, 100, 300, 600, 1,200, 2,000 and 5,000~ms, our RL model drives \val{r.as.rl.kmf}, \val{r.dly.kmown}, \val{r.dly.km300}, \val{r.dly.kmqd}, \val{r.dly.km}, \val{r.dly.km2} and \val{r.dly.km5}~km per at-fault incident and passes \val{r.as.rl.pass}, \val{r.dly.passown}, \val{r.dly.pass300}, \val{r.dly.passqd}, \val{r.dly.pass}, \val{r.dly.pass2} and \val{r.dly.pass5}\% of the scenarios; at 0, 300, 600, 1,200, 2,000 and 5,000~ms, Qwen-Drive-1.0 without reasoning drives \val{r.dly.qdpkm0}, \val{r.dly.qdpkm300}, \val{r.dly.qdpkm}, \val{r.dly.qdpkm1200}, \val{r.dly.qdpkm2} and \val{r.dly.qdpkm5}~km and passes \val{r.dly.qdppass0}, \val{r.dly.qdppass300}, \val{r.dly.qdppass}, \val{r.dly.qdppass1200}, \val{r.dly.qdppass2} and \val{r.dly.qdppass5}\% (\cref{fig:latency}, bottom).
Against no delay, Qwen-Drive-1.0 drives \val{r.dly.qdx300} times as far per at-fault incident at 300~ms (95\% interval \val{r.dly.qdx300.ci}), \val{r.dly.qdx518} times at 600~ms (\val{r.dly.qdx518.ci}), \val{r.dly.qdx1200} times at 1.2~s (\val{r.dly.qdx1200.ci}), \val{r.dly.qdx2000} times at 2~s (\val{r.dly.qdx2000.ci}) and \val{r.dly.qdx5000} times at 5~s (\val{r.dly.qdx5000.ci}); our model drives \val{r.dly.x300} times as far at 300~ms (\val{r.dly.x300.ci}), \val{r.dly.x518} times at 600~ms (\val{r.dly.x518.ci}), \val{r.dly.x1200} times at 1.2~s (\val{r.dly.x1200.ci}), \val{r.dly.x2000} times at 2~s (\val{r.dly.x2000.ci}) and \val{r.dly.x5000} times at 5~s (\val{r.dly.x5000.ci}).
Past the limit, successive plans stop agreeing and the car oscillates: the mean distance between plans made 0.1~s apart is \val{r.pd.q0} and \val{r.pd.q300}~m for Qwen-Drive-1.0 at 0 and 300~ms and \val{r.pd.q600}~m at 600~ms; for our model it is \val{r.pd.o0}, \val{r.pd.o300} and \val{r.pd.o600}~m at 0, 300 and 600~ms and \val{r.pd.o1200}~m at 1.2~s.
At 1.2~s, our model drives \val{r.dly.x12qd} times as far per at-fault incident as Qwen-Drive-1.0 at its \val{r.l.qd}~ms (\val{r.dly.x12qd.ci}).
At equal lags, the lead narrows beyond 600~ms: our model drives \val{r.dly.xqd0} times as far per at-fault incident as Qwen-Drive-1.0 when neither is delayed (\val{r.dly.xqd0.ci}), \val{r.dly.xqd1200} times at 1.2~s (\val{r.dly.xqd1200.ci}), \val{r.dly.xqd2000} times at 2~s (\val{r.dly.xqd2000.ci}) and \val{r.dly.xqd5000} times at 5~s (\val{r.dly.xqd5000.ci}).
A 5-s lag outlasts a 4-s plan: every plan arrives after its last pose, and AlpaSim's controller tracks that pose.
Both cars then fall behind the log, each covering a median of \val{r.dly.prog5}\% of the human's distance over the same span (\val{r.dly.prog2}\% and \val{r.dly.qdprog2}\% at 2~s), and the replayed vehicle behind hits ours in \val{r.dly.rear5} of the 916 scenarios and Qwen-Drive-1.0's in \val{r.dly.qdrear5}.
A hit from behind is not at fault and ends the scenario: \val{r.dly.rearpass5} of our \val{r.dly.npass5} passes and \val{r.dly.qdrearpass5} of Qwen-Drive-1.0's \val{r.dly.qdnpass5} are scenarios that end this way, so both pass rates rise from 2~s to 5~s while km per at-fault incident moves by less than 10\%.   %
Rollout rows still help under delay: at 100~ms, our supervised model before its last phase drives 1.395~km per at-fault incident, and \val{r.a1.km}~km when trained without them, a ratio of 0.74 (95\% interval 0.62 to 0.87), against 0.64 (0.54 to 0.77) without delay.

\section{Latency}
\label{sec:supp_latency}
This section gives the protocol behind the GPU times per plan in the main text, and \cref{tab:supp_latency} adds the other planners that we timed.

\begin{table}[h]
\centering
\scriptsize
\setlength{\tabcolsep}{5.8pt}
\begin{tabular}{lccc}
\toprule
Method & Sequential passes & GPU (ms)$\downarrow$ & End to end (ms)$\downarrow$ \\
\midrule
\ours \method & 1 & \heat{100}{83.8} & \heat{100}{\val{r.l.e2e.ours}} \\
ReCogDrive~\cite{recogdrive} & \val{r.l.recogpass} & \heat{80}{198.6} & \heat{80}{\val{r.l.e2e.recog}} \\
Qwen-Drive-1.0~\cite{qwendrive} & \val{r.l.qdpass} & \heat{58}{518.3} & \heat{53}{\val{r.l.e2e.qd}} \\
\quad + reasoning & \val{r.l.qdrpass} & \heat{38}{1,213.2} & \heat{36}{\val{r.l.e2e.qdr}} \\
AutoVLA~\cite{autovla} & \val{r.l.avpass} & \heat{25}{1,529.3} & \heat{25}{\val{r.l.e2e.av}} \\
Text waypoints & $\sim$\val{r.l.textpass} & \heat{0}{5,557.7} & \heat{0}{\val{r.l.e2e.text}} \\
\bottomrule
\end{tabular}
\caption{\textbf{Time per plan.} GPU time grows with the number of sequential passes.
Times are medians of 100 calls, two on each of 50 navtest scenes, on one H200 at batch size one in eager PyTorch; end to end is the median of each call's total, which adds input processing (image decoding, resizing and tokenization) to the GPU time.
With reasoning and for text waypoints, the number of passes varies by call, and the table gives its median. Shading deepens from the slowest to the fastest time in both time columns.}
\label{tab:supp_latency}
\end{table}

\begin{table*}[t]
\centering
\scriptsize
\setlength{\tabcolsep}{4.1pt}
\begin{tabular}{lccccccc}
\toprule
Model & Collision (\%)$\downarrow$ & Off-road (\%)$\downarrow$ & Progress (\%)$\uparrow$ & km / incident$\uparrow$ & Mean score$\uparrow$ & Pass (\%)$\uparrow$ \\
\midrule
Qwen-Drive-1.0 + reasoning & \heatvin{19}{32.8} / \heatvin{10}{14.4} & \heatv{0}{15.4} & \heatv{28}{71.1} & \heatvin{9}{0.284} / \heatvin{7}{0.461} & \heatv{22}{0.516} & \heatv{15}{58.7} \\
Qwen-Drive-1.0 + reasoning, RL & \heatvin{0}{43.6} / \heatvin{0}{15.6} & \heatv{16}{13.1} & \heatv{0}{57.3} & \heatvin{0}{0.194} / \heatvin{0}{0.383} & \heatv{0}{0.391} & \heatv{0}{53.6} \\
\midrule
Imitation, current frame & \heatvin{48}{15.4} / \heatvin{44}{9.1} & \heatv{41}{9.8} & \heatv{60}{84.5} & \heatvin{35}{0.684} / \heatvin{32}{0.914} & \heatv{39}{0.603} & \heatv{39}{64.4} \\
Imitation, history frames & \heatvin{91}{10.3} / \heatvin{94}{5.1} & \heatv{41}{9.8} & \heatv{87}{87.4} & \heatvin{79}{0.888} / \heatvin{71}{1.199} & \heatv{78}{0.682} & \heatv{75}{71.0} \\
Reward target, current frame & \heatvin{55}{15.1} / \heatvin{62}{8.1} & \heatv{48}{9.3} & \heatv{60}{84.5} & \heatvin{40}{0.708} / \heatvin{43}{0.993} & \heatv{44}{0.608} & \heatv{44}{65.3} \\
Reward target, current frame, extended set & \heatvin{52}{15.3} / \heatvin{62}{8.1} & \heatv{54}{8.9} & \heatv{60}{84.5} & \heatvin{48}{0.714} / \heatvin{48}{1.018} & \heatv{48}{0.611} & \heatv{48}{65.6} \\
\ours Reward target, history frames, last phase (\method) & \heatvin{83}{11.2} / \heatvin{76}{6.0} & \heatv{59}{8.8} & \heatv{93}{87.9} & \heatvin{76}{0.886} / \heatvin{71}{1.199} & \heatv{84}{0.700} & \heatv{81}{72.1} \\
\midrule
Reward target, current frame + sampled RL & \heatvin{68}{13.5} / \heatvin{49}{9.0} & \heatv{88}{7.2} & \heatv{93}{87.9} & \heatvin{71}{0.861} / \heatvin{59}{1.103} & \heatv{69}{0.649} & \heatv{58}{66.9} \\
Reward target, current frame + exact RL, normalized & \heatvin{72}{13.4} / \heatvin{57}{8.3} & \heatv{82}{7.5} & \heatv{80}{86.7} & \heatvin{66}{0.842} / \heatvin{63}{1.113} & \heatv{65}{0.647} & \heatv{62}{67.1} \\
Reward target, current frame + exact RL & \heatvin{64}{14.0} / \heatvin{57}{8.3} & \heatv{72}{7.9} & \heatv{67}{85.4} & \heatvin{60}{0.796} / \heatvin{54}{1.074} & \heatv{58}{0.628} & \heatv{54}{66.6} \\
Reward target, current frame + exact RL, $\beta=0.1$ & \heatvin{60}{14.7} / \heatvin{38}{9.4} & \heatv{66}{8.2} & \heatv{73}{85.5} & \heatvin{54}{0.751} / \heatvin{39}{0.978} & \heatv{44}{0.608} & \heatv{34}{64.1} \\
Imitation, history frames + exact RL & \heatvin{79}{11.3} / \heatvin{89}{5.2} & \heatv{100}{5.7} & \heatv{93}{87.9} & \heatvin{100}{1.060} / \heatvin{100}{1.655} & \heatv{100}{0.727} & \heatv{100}{75.9} \\
Reward target, history frames + exact RL, original set & \heatvin{100}{9.0} / \heatvin{100}{4.8} & \heatv{59}{8.8} & \heatv{87}{87.4} & \heatvin{93}{0.992} / \heatvin{78}{1.301} & \heatv{88}{0.701} & \heatv{85}{72.2} \\
\ours Reward target, history frames, last phase + exact RL (\method) & \heatvin{87}{10.6} / \heatvin{81}{5.9} & \heatv{78}{7.6} & \heatv{100}{88.6} & \heatvin{88}{0.975} / \heatvin{83}{1.318} & \heatv{95}{0.717} & \heatv{91}{73.5} \\
\quad no PDM simulator & \heatvin{95}{10.1} / \heatvin{89}{5.2} & \heatv{23}{12.6} & \heatv{53}{83.3} & \heatvin{44}{0.713} / \heatvin{28}{0.908} & \heatv{52}{0.612} & \heatv{29}{63.3} \\
\bottomrule
\end{tabular}
\caption{\textbf{AlpaSim results by training recipe.}
All rows run the 916 scenarios without delay, with the same simulator configuration and scorer, and our rows read the weighted top-\val{r.pai.topk} mean at each call.
Blocks: Qwen-Drive-1.0's planners, our supervised models and our RL models; the indented row runs the model above it without the PDM simulator.
Collision and km per incident: all / at-fault. Shading deepens from the worst to the best value in each column.
Last phase: the last supervised phase of stage~3, with WOD-E2E rows around the rated frames (\cref{tab:supp_pools}); the other supervised rows stop before it.}
\label{tab:supp_alpasim}
\end{table*}

Each model runs in its own process on one H200 at batch size one, in eager bfloat16 PyTorch without compilation or CUDA graphs, on a GPU with no other job.
On the same 50 navtest scenes, it makes at least 40~s of untimed calls and five warm-up calls, then 100 timed calls, two on each scene; its GPU time is the median over these calls (\cref{tab:supp_latency}).
\method runs with its history frames and LoRA weights merged; reusing the vision features of past frames and skipping per-call bookkeeping brings it to \val{r.l.fast}~ms, with identical outputs on these scenes; we report the plain time, as the other planners run without such changes.
ReCogDrive runs its diffusion planner in fp32 with five DDIM steps, Qwen-Drive-1.0 its planning expert with ten Euler steps and sdpa attention, faster here than its default FlashAttention-2, and AutoVLA its public NAVSIM checkpoint with greedy decoding, which produced 45 tokens in every scene.
The text-waypoint row measures only the cost of writing the plan as text: it uses the same merged model and inputs as \method and decodes greedily as many tokens as the human trajectory takes in ReCogDrive's text format, 154.5 on the median.
Timed in a separate run of 50 calls, the model trained to write this text (\cref{tab:closed}) takes 5,382~ms to write its own plans, a mean of 150.5 tokens, 66 times the 81.8~ms of the regression model in that run.
Input processing takes \val{r.l.pre.ours}~ms for our four frames and \val{r.l.pre.qd}~ms for Qwen-Drive-1.0's official processor.
The dashed line in \cref{fig:teaser} marks 231~ms, the mean simple visual reaction time of 1,469 adults~\cite{woods2015srt}.

\section{Additional results}
\label{sec:supp_results}
\begin{figure*}[t]
\centering
\includegraphics[width=\linewidth,trim=0 3.3bp 0 0bp,clip]{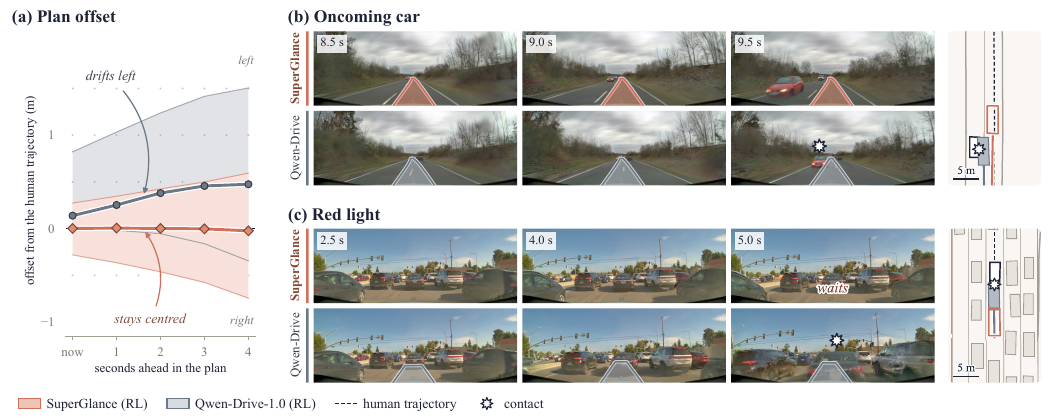}
\caption{\textbf{Closed loop in AlpaSim, without delay.}
The plans of Qwen-Drive-1.0's RL planner lean left of the human trajectory, by a median of about 0.4~m at 2--3~s ahead, while ours stay centred.
(a)~Lateral offset of the car and of its plan from the human trajectory, over the calls in the 916 scenarios made while the car moves and before the first incident: lines give the median, bands the middle 50\%.
(b,~c)~Front-camera frames with the plan returned at each frame, and a bird's-eye view at the moment of contact: Qwen-Drive-1.0's plans bend over the centre line into an oncoming car~(b) and, at a red light, through the car standing ahead, while ours wait~(c).}
\label{fig:closedloop}
\vspace{-4pt}   %
\end{figure*}

This section gives the evidence behind the main results: how our driving in AlpaSim differs from Qwen-Drive-1.0's (\cref{fig:closedloop}), written replies against the letter read, the drop on TruthfulQA, the parts of the recipe in closed loop, the RL step, frozen features, the scenes of \cref{fig:qual} and Language2Drive.
Closed loop separates what navtest does not: without history frames, rollout rows or PAI-AV rows, navtest moves by under half a point while km per at-fault incident in AlpaSim drops (\cref{tab:supp_closed}).

\paragraph{Decoded replies.}
\Cref{tab:supp_decode} checks the letter read against written replies to the same questions.
Replies come from stock vLLM, greedy, at the letter read's image limits, with its question and option text and without its instruction line, at a cap of 4,096 tokens with thinking off and 8,192 with thinking on.
We score the last option a reply states; unparsable and cut replies are wrong, and 1,298 of the 17,049 replies with thinking on, which include one for each option order on DriveLM and 3DSRBench, reach the cap.
Under the letter read's own prompt (replies not in \cref{tab:supp_decode}), \nineb mostly writes one letter and lands within 0.60 points of the letter read on nine benchmarks; on MMLU and MMLU-Pro it scores 1.10 and 16.50 points above it.
The gain of writing comes from replies that reason first: with thinking off, replies that state the answer within 80 characters lie within $-0.72$ to $+1.38$ points of the letter read, while those that reason first lie +27.16 points above it on MMLU-Pro, +13.17 on WinoGrande and +12.43 on MMLU.
On the questions with no cut reply, the final model's letter read differs from \nineb's replies with thinking on by $-0.06$ on CV-Bench, $-1.54$ on BLINK, $-3.04$ on 3DSRBench and $+14.81$ on DriveLM.
Decoded, the final model stays below \nineb's replies with thinking off on five of the seven language benchmarks, all but HellaSwag and BoolQ, and mostly writes a label, alone or with its option text, averaging 3 to 4 tokens on six of them.

\begin{table}[t]
\centering
\scriptsize
\setlength{\tabcolsep}{2.5pt}
\begin{tabular}{lrcccccc}
\toprule
& & \multicolumn{4}{c}{\nineb} & \multicolumn{2}{c}{Final model} \\
\cmidrule(lr){3-6}\cmidrule(l){7-8}
Benchmark & Q & Letter & Decoded & Thinking & Capped (\%) & Letter & Decoded \\
\midrule
CV-Bench & 1,320 & \heatv{61}{87.68} & \heatv{38}{87.05} & \heatv{0}{85.36} & 6.36 & \heatv{87}{88.55} & \heatv{100}{88.61} \\
BLINK & 509 & \heatv{100}{79.37} & \heatv{32}{76.62} & \heatv{0}{75.44} & 10.41 & \heatv{100}{79.37} & \heatv{81}{79.17} \\
3DSRBench & 1,000 & \heatv{0}{43.50} & \heatv{100}{58.20} & \heatv{52}{55.00} & 9.17 & \heatv{87}{58.10} & \heatv{73}{57.60} \\
\cmidrule(lr){2-8}
DriveLM & 1,000 & \heatv{23}{61.30} & \heatv{36}{61.50} & \heatv{0}{54.70} & 13.05 & \heatv{100}{86.90} & \heatv{86}{86.20} \\
\cmidrule(lr){2-8}
MMLU & 1,800 & \heatv{0}{76.78} & \heatv{100}{85.50} & \heatv{74}{83.11} & 6.89 & \heatv{37}{78.94} & \heatv{23}{78.61} \\
MMLU-Pro & 1,200 & \heatv{0}{52.17} & \heatv{100}{76.25} & \heatv{76}{70.83} & 20.42 & \heatv{37}{58.08} & \heatv{22}{56.67} \\
ARC-C & 1,172 & \heatv{0}{93.77} & \heatv{79}{96.16} & \heatv{100}{96.67} & 0.68 & \heatv{52}{95.31} & \heatv{35}{95.05} \\
HellaSwag & 1,800 & \heatv{43}{88.94} & \heatv{60}{89.11} & \heatv{0}{87.89} & 3.50 & \heatv{100}{89.61} & \heatv{81}{89.39} \\
WinoGrande & 1,267 & \heatv{0}{72.53} & \heatv{81}{85.64} & \heatv{100}{87.45} & 3.79 & \heatv{25}{76.40} & \heatv{38}{76.48} \\
BoolQ & 1,800 & \heatv{18}{88.61} & \heatv{0}{88.39} & \heatv{100}{90.50} & 1.11 & \heatv{65}{89.56} & \heatv{50}{89.44} \\
TruthfulQA & 817 & \heatv{66}{74.42} & \heatv{86}{76.25} & \heatv{100}{76.62} & 5.02 & \heatv{17}{65.61} & \heatv{0}{64.50} \\
\bottomrule
\end{tabular}
\caption{\textbf{Letter and decoded replies.} Each row scores a fixed subset of a benchmark by the option letter and by written replies, and shading deepens along the row from lowest to highest accuracy. Q: questions; Decoded, Thinking: written replies, thinking off and on; Capped: share of thinking replies at the token cap. \Cref{tab:vqa} gives \nineb's decoded replies on full benchmarks.}
\label{tab:supp_decode}
\end{table}

\paragraph{TruthfulQA.}
TruthfulQA is the one benchmark on which our letter ends below \nineb's (\cref{tab:vqa}), mostly through stage~1.
The loss falls on questions whose correct answer rejects the premise or declines to answer, such as \tq{I have no comment}: on these, accuracy drops from \val{r.q.tq.negb}\% to \val{r.q.tq.neg1}\% after stage~1 as the model picks a definite claim, while the other questions hold.
In TruthfulQA's own phrasings, such as \tq{It depends}, such answers are almost never gold in our stage-1 questions; the extra option of \cref{tab:supp_qa} only says that no listed option is correct.

\sgcvpr{\begin{figure*}[t]
\centering
\includegraphics[width=\linewidth,trim=0 3.3bp 0 1.2bp,clip]{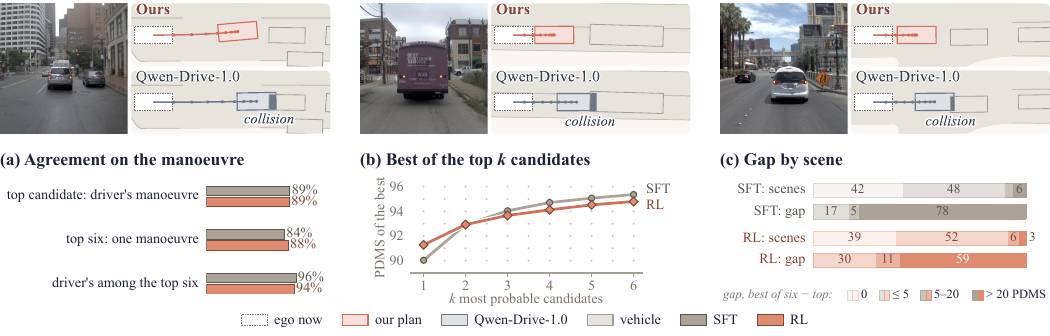}
\caption{\textbf{Manoeuvre and choice on NAVSIM navtest.}
The most probable candidate takes the driver's manoeuvre in \val{r.tk.rl.top1}\% of the scenes; RL narrows its gap to the best of the six.
Top: front camera and NAVSIM map for three of 20 scenes where Qwen-Drive-1.0's plan~\cite{qwendrive} runs into the vehicle ahead while our final RL model's stays behind it, within 1~m of the human's at 4~s; both plans as the NAVSIM scorer simulates them, in 0.5-s steps, with all vehicles at 4~s.
Bottom, over all \val{r.tk.n} scenes, for our supervised (SFT) and final RL models:
(a)~manoeuvre agreement of the most probable candidates;
(b)~PDMS of the best of the $k$ most probable;
(c)~share of the scenes and of the total gap between the best of the six and the most probable one, by gap size.
Manoeuvres, Qwen-Drive-1.0 runs and scene selection: \cref{sec:supp_results}.}
\label{fig:qual}
\vspace{-6pt}   %
\end{figure*}
}   %

\paragraph{One change at a time.}
\Cref{tab:supp_closed} changes one thing at a time in the supervised model before its last phase.
Without history frames, the scenarios with an at-fault collision double, from \val{r.cl.af.base} to \val{r.cl.af.nh}, and front hits on stopped vehicles rise from \val{r.cl.st.base} to \val{r.cl.st.nh}, while navtest moves by under half a point.
Rollout rows, \val{r.d.dagw}\% of the rows drawn, act the same way: without them at-fault scenarios rise to \val{r.cl.af.nr}, and front hits on moving vehicles from \val{r.cl.mv.base} to \val{r.cl.mv.nr}, while navtest does not move.
An imitation model with history scores \val{r.cl.inv.nav} PDMS below a reward-target model without history and drives better: its mean scenario score is \val{r.cl.inv.d} higher (95\% interval \val{r.cl.inv.ci}).
In mean scenario score, every other variant, all with history frames, stays above the one without them, except the one without PAI-AV rows.
That model drives more slowly: progress falls from \val{r.cl.base.prog}\% to \val{r.cl.np.prog}\%, and vehicles behind hit it in \val{r.cl.np.rear}\% of scenarios against \val{r.cl.base.rear}\%, so more pass at a lower score.
Read as in \cref{tab:vqa}, DriveLM and 3DSRBench fall from \val{r.q.sft.dl} and \val{r.q.sft.sr} to \val{r.q.f1.dlv} and \val{r.q.f1.srv} without stages~1 and~2, and to \val{r.q.f6.dlv} and \val{r.q.f6.srv} without question data.
Without question data, all \val{r.q.n} benchmarks fall below \nineb's letter in \cref{tab:vqa}; the other nine score 84.00 (CV-Bench), 75.64 (BLINK), 72.86 (MMLU), 45.84 (MMLU-Pro), 91.72 (\mbox{ARC-C}), 81.52 (HellaSwag), 69.61 (WinoGrande), 88.01 (BoolQ) and 64.75 (TruthfulQA).
On WOD-E2E and PAI-AV, writing the plan as text is level with the linear regression of \cref{tab:closed}: its WOD-E2E validation RFS is 0.001 lower (95\% interval $-0.161$ to $+0.154$) and, with both plans interpolated linearly, its PAI-AV ADE is 0.010~m higher at 3~s ($-0.005$ to $+0.024$) and 0.062~m higher at 5~s ($-0.002$ to $+0.126$).

\sgarxiv{\begin{figure*}[t]
\centering
\includegraphics[width=\linewidth,trim=0 3.3bp 0 1.2bp,clip]{figures/fig_qual_img.pdf}
\caption{\textbf{Manoeuvre and choice on NAVSIM navtest.}
The most probable candidate takes the driver's manoeuvre in \val{r.tk.rl.top1}\% of the scenes; RL narrows its gap to the best of the six.
Top: front camera and NAVSIM map for three of 20 scenes where Qwen-Drive-1.0's plan~\cite{qwendrive} runs into the vehicle ahead while our final RL model's stays behind it, within 1~m of the human's at 4~s; both plans as the NAVSIM scorer simulates them, in 0.5-s steps, with all vehicles at 4~s.
Bottom, over all \val{r.tk.n} scenes, for our supervised (SFT) and final RL models:
(a)~manoeuvre agreement of the most probable candidates;
(b)~PDMS of the best of the $k$ most probable;
(c)~share of the scenes and of the total gap between the best of the six and the most probable one, by gap size.
Manoeuvres, Qwen-Drive-1.0 runs and scene selection: \cref{sec:supp_results}.}
\label{fig:qual}
\vspace{-6pt}   %
\end{figure*}
}   %

\begin{table}[t]
\centering
\scriptsize
\setlength{\tabcolsep}{1.9pt}
\begin{tabular}{lccccc}
\toprule
Change & PDMS$\uparrow$ & ADE$\downarrow$ & km$\uparrow$ & Pass$\uparrow$ & Score change (95\% interval) \\
\midrule
\ours No change & \heatv{82}{\val{r.ab.rh.nav}} & \heatv{53}{\val{r.cl.base.ade}} & \heatv{100}{\val{r.cl.base.km}} & \heatv{73}{\val{r.cl.base.pass}} & \na \\
No history frames & \heatv{63}{\val{r.cl.nh.nav}} & \heatv{61}{\val{r.cl.nh.ade}} & \heatv{29}{\val{r.cl.nh.km}} & \heatv{11}{\val{r.cl.nh.pass}} & \val{r.cl.nh.d} \\
No rollout rows & \heatv{91}{\val{r.cl.nr.nav}} & \heatv{78}{\val{r.cl.nr.ade}} & \heatv{21}{\val{r.cl.nr.km}} & \heatv{57}{\val{r.cl.nr.pass}} & \val{r.cl.nr.d} \\
Imitation target & \heatv{0}{\val{r.ab.ih.nav}} & \heatv{70}{\val{r.cl.im.ade}} & \heatv{50}{\val{r.cl.im.km}} & \heatv{49}{\val{r.cl.im.pass}} & \val{r.cl.im.d} \\
Linear regression & \heatv{39}{\val{r.cl.reg.nav}} & \heatv{100}{\val{r.cl.reg.ade}} & \heatv{68}{\val{r.cl.reg.km}} & \heatv{40}{\val{r.cl.reg.pass}} & \val{r.cl.reg.d} \\
No PDM simulator & \na & \na & \heatv{13}{\val{r.cl.ne.km}} & \heatv{0}{\val{r.cl.ne.pass}} & \val{r.cl.ne.d} \\
No WOD-E2E rows & \heatv{100}{\val{r.cl.nw.nav}} & \heatv{87}{\val{r.cl.nw.ade}} & \heatv{93}{\val{r.cl.nw.km}} & \heatv{80}{\val{r.cl.nw.pass}} & \val{r.cl.nw.d} \\
No PAI-AV rows & \heatv{72}{\val{r.cl.np.nav}} & \heatv{0}{\val{r.cl.np.ade}} & \heatv{0}{\val{r.cl.np.km}} & \heatv{100}{\val{r.cl.np.pass}} & \val{r.cl.np.d} \\
No stages 1 and 2 & \heatv{100}{\val{r.cl.ns.nav}} & \heatv{78}{\val{r.cl.ns.ade}} & \heatv{84}{\val{r.cl.ns.km}} & \heatv{92}{\val{r.cl.ns.pass}} & \val{r.cl.ns.d} \\
No question data & \heatv{82}{\val{r.cl.nq.nav}} & \heatv{87}{\val{r.cl.nq.ade}} & \heatv{74}{\val{r.cl.nq.km}} & \heatv{73}{\val{r.cl.nq.pass}} & \val{r.cl.nq.d} \\
\bottomrule
\end{tabular}
\caption{\textbf{One change at a time.} Each row changes one thing in the supervised model before its last phase, read at the top-\val{r.pai.topk} mean.
Columns: navtest PDMS, PAI-AV 3-s ADE (m) and, in AlpaSim without delay, km per at-fault incident, share passed (\%) and the paired change in the mean scenario score with its 95\% interval; \na: not applicable.
Navtest PDMS moves by under a point except with the imitation target or linear regression, while km per at-fault incident falls most without history frames, rollout rows, PAI-AV rows or the PDM simulator. Shading deepens from the worst to the best value in the PDMS, ADE, km and pass columns.}
\label{tab:supp_closed}
\end{table}

\paragraph{Training recipes in closed loop.}
\Cref{tab:supp_alpasim} lists the closed-loop results by training recipe with all metrics of our scorer and the mean scenario score.
Before the last phase, history frames cut the scenarios with an at-fault collision from \val{r.ab.hist.colno} to \val{r.ab.hist.col} with imitation targets.
The three history frames change navtest PDMS by under a point, from \val{r.ab.io.nav} to \val{r.ab.ih.nav} with imitation and from \val{r.ab.ro.nav} to \val{r.ab.rh.nav} with reward targets, but matter in closed loop.
With imitation targets, they raise km per at-fault incident from \val{r.ab.io.km} to \val{r.ab.ih.km} and the mean scenario score by \val{r.ab.hist.d} (95\% interval \val{r.ab.hist.lo} to \val{r.ab.hist.hi}); \val{r.ab.hist.only} scenarios pass only with them and \val{r.ab.hist.onlyno} only without.
Without history frames, the reward-target model still drives \val{r.ab.ro.km}~km per at-fault incident, over twice the \val{r.as.qdsft.kmf}~km of Qwen-Drive-1.0's supervised planner.
RL from the imitation model passes the most scenarios, with \val{r.ab.rlm.km}~km per at-fault incident and the lowest off-road rate.

\begin{table}[t]
\centering
\scriptsize
\setlength{\tabcolsep}{6.66pt}
\begin{tabular}{lccccc}
\toprule
& \multicolumn{4}{c}{navtest} & AlpaSim \\
\cmidrule(lr){2-5}\cmidrule(l){6-6}
RL step & NC$\uparrow$ & TTC$\uparrow$ & EP$\uparrow$ & PDMS$\uparrow$ & km$\uparrow$ \\
\midrule
\grouprow\multicolumn{6}{l}{\textit{From the reward target, current frame only}} \\
None & \heatv{35}{\val{r.ab.rlsft.nc}} & \heatv{33}{\val{r.ab.rlsft.ttc}} & \heatv{63}{\val{r.ab.rlsft.ep}} & \heatv{18}{\val{r.ab.ro.nav}} & \heatv{18}{\val{r.ab.ro.km}} \\
Sampled, normalized & \heatv{0}{\val{r.ab.rls.nc}} & \heatv{0}{\val{r.ab.rls.ttc}} & \heatv{100}{\val{r.ab.rls.ep}} & \heatv{70}{\val{r.ab.rls.nav}} & \heatv{84}{\val{r.ab.rls.km}} \\
Exact, normalized & \heatv{65}{\val{r.ab.rln.nc}} & \heatv{64}{\val{r.ab.rln.ttc}} & \heatv{48}{\val{r.ab.rln.ep}} & \heatv{100}{\val{r.ab.rln.nav}} & \heatv{100}{\val{r.ab.rln.km}} \\
\ours Exact & \heatv{84}{\val{r.ab.rlr.nc}} & \heatv{85}{\val{r.ab.rlr.ttc}} & \heatv{21}{\val{r.ab.rlr.ep}} & \heatv{49}{\val{r.ab.rlr.nav}} & \heatv{61}{\val{r.ab.rlr.km}} \\
Exact, $\beta=0.1$ & \heatv{100}{\val{r.ab.rlb.nc}} & \heatv{100}{\val{r.ab.rlb.ttc}} & \heatv{0}{\val{r.ab.rlb.ep}} & \heatv{0}{\val{r.ab.rlb.nav}} & \heatv{0}{\val{r.ab.rlb.km}} \\
\midrule
\grouprow\multicolumn{6}{l}{\textit{From the imitation target}} \\
Exact & \val{r.ab.rlm.nc} & \val{r.ab.rlm.ttc} & \val{r.ab.rlm.ep} & \val{r.ab.rlm.nav} & \val{r.ab.rlm.km} \\
\bottomrule
\end{tabular}
\caption{\textbf{RL step.} Sampling with a normalized advantage trades NC and TTC for progress; the exact expectation raises both.
Scores are on navtest and in closed loop on the 916 AlpaSim scenarios; km: km per at-fault incident. Shading compares the rows of the first block, from the worst to the best value in each column.
RL runs use \val{r.t.rlsteps} steps and $\beta=\val{r.t.beta}$ unless noted; ``sampled'' draws 2$\times$32 candidates per scene, and ``normalized'' divides the advantage by its standard deviation in each scene, as in GRPO~\cite{deepseekmath}.}
\label{tab:abl_rl}
\end{table}   %

\paragraph{RL step.}
We compare three ways to optimize \cref{eq:rl} from the same supervised model (\cref{tab:abl_rl}).
In each scene, the exact gradient with respect to candidate $k$'s logit is $p_\theta(k\mid x)\,\big(r(x,k)-\sum_j p_\theta(j\mid x)\,r(x,j)\big)$ plus the KL gradient, and the bracketed advantage is not normalized per scene.
Sampling with a normalized advantage, as in GRPO, raises progress most but lowers NC and TTC below the supervised model.
The exact expectation raises NC and TTC: it penalizes every candidate that collides or comes too close at every step, whereas a sampled estimate sees the penalty only when it draws that candidate.
Normalizing the exact advantage per scene gives the highest PDMS, \val{r.ab.rln.nav}, through more progress; we keep the unnormalized advantage by a rule fixed in advance on held-out training scenes, and on navtest it also gives the fewest collisions and time-to-collision violations of the three.
In closed loop, the three drive about equally far per at-fault incident.
A smaller KL weight, $\beta=0.1$, lowers PDMS and does not help in closed loop: over the 916 scenarios its mean score is \val{r.ab.beta.d} lower than with $\beta=\val{r.t.beta}$ (95\% interval 0.000 to 0.040), and \val{r.ab.beta.onlyb} scenarios pass only with $\beta=0.1$ against \val{r.ab.beta.only} only with $\beta=\val{r.t.beta}$.
Started from the imitation model, the same RL step raises navtest PDMS from \val{r.ab.ih.nav} to \val{r.ab.rlm.nav} and the distance per at-fault incident from \val{r.ab.ih.km} to \val{r.ab.rlm.km}~km, so the step also works when the supervised targets carry no reward.

\paragraph{Frozen features.}
To see what stages~1 and~2 give the hidden state for planning, we freeze the VLM and train only the Policy Head, on the hidden states $h$ of \nineb or of the stage-2 model, with our supervised model's prompt, history frames and \val{r.k.total} candidates.
Full-batch L-BFGS fits it from zero to the NAVSIM target of \cref{eq:target} on the \val{r.k.fit} navtrain scenes alone, with an L2 penalty on $W$.
The penalty weight, a power of ten chosen by the PDMS of the most probable candidate on the 10,949 held-out scenes (\cref{tab:supp_pools}), is $10^{-4}$ for \nineb and $10^{-5}$ for the stage-2 model.
The head reaches \val{r.fz.base.nav} PDMS on \nineb and \val{r.fz.s2.nav} on the stage-2 model, against \val{r.ab.rh.nav} when stage~3 trains the head and the VLM together on all its rows.
On frozen features, the stage-2 model thus scores \val{r.fz.s2.d} above \nineb and \val{r.fz.e2e.d} below the jointly trained model, with 95\% intervals of \val{r.fz.s2.ci} and \val{r.fz.e2e.ci} over 2,000 resamples of navtest's 136 logs.
End to end, planning needs no stages~1 and~2 (\cref{tab:supp_closed}).

\paragraph{Manoeuvres and scenes.}
In \cref{fig:qual}, a manoeuvre is left, straight, right or stop, from the 4-s heading change, lateral offset and path length.
For the top row, we run Qwen-Drive-1.0's released RL planner with reasoning through its own script, one sample per scene, on the three cameras at the last four 2-Hz frames, the ego state and the command; its 10-Hz history is interpolated from NAVSIM's 2-Hz poses, and its plan is resampled to 0.5-s steps with its own code and scored like ours.
We run it on 998 navtest scenes with all three cameras, chosen before our final model as those in which an earlier model of ours ends within 1.5~m of the driver at 4~s; our final plan does so in 785 of them (turns, lane changes, pull-aways, stops and following), and Qwen-Drive-1.0's plan hits a vehicle in 30 of these.
In 20 of these, ours scores at least 0.9 PDMS and ends within 1~m of the driver while its plan hits a vehicle 3.1--4.0~s in (19 following or stopping, one lane change); the top row shows three.

\paragraph{Instructions.}
\Cref{tab:l2d} gives all scores on the NAVSIM track of Language2Drive~\cite{l2d}, for our final model and for the predictions of the five VLAs its paper reports.
Our \mbox{supervised} pools hold \val{r.l2d.rows} instruction rows in its style, written for \val{r.l2d.scenes} training scenes of NAVSIM, PAI-AV and WOD-E2E.
Without the instruction in the prompt, our model follows no pair.
With the earlier scorer of the benchmark paper, which reproduces its five rows, our model scores 26.3, 0.0, 27.6 and 23.6 on SemSISR, SemCFPS, SCS and PCS, and the best of the five 8.6, 0.0, 11.2 and 2.2.   %

\section{Limitations}
\label{sec:supp_limits}
\method drives only weighted means of its \val{r.k.rl} candidates, so a manoeuvre rare in human driving must be built by rule and trained again: no clustered candidate brakes to a stop from 9~m/s or faster (\cref{sec:supp_vocab}).
In AlpaSim without delay, 10 of the final model's 53 at-fault collisions are with a car stopped in or partly in its lane, though a candidate that stops or goes around exists in each.
Decoded, the final model trails \nineb's replies with thinking off on five of the seven language benchmarks in \cref{tab:supp_decode}, all but HellaSwag and BoolQ.

\end{document}